\pdfoutput=1
\documentclass[10pt,twocolumn,letterpaper]{article}
\usepackage[final]{cvpr}              
 \usepackage{times}
\usepackage{bm}

\usepackage[dvipsnames]{xcolor}

\definecolor{cvprblue}{rgb}{0.21,0.49,0.74}
\definecolor{Gray}{gray}{0.9}
\usepackage[pagebackref,breaklinks,colorlinks,citecolor=cvprblue]{hyperref}
\usepackage{color, colortbl}
\usepackage{multirow}
\usepackage{bbding}
\usepackage[ruled,linesnumbered]{algorithm2e}
\usepackage{caption, subcaption}
 
\def\paperID{14485} 
\def\confName{CVPR}
\def\confYear{2024}

\title{Unbiased Faster R-CNN for Single-source Domain Generalized Object Detection}
\author{Yajing Liu$^{1,2,3}$, Shijun Zhou$^{1,2,3}$, Xiyao Liu$^{1,2}$, Chunhui Hao$^{1,2}$, Baojie Fan$^{4}$, Jiandong Tian$^{1,2}$\thanks{Corresponding author}\\
$^{1}$State Key Laboratory of Robotics, Shenyang Institute of Automation, Chinese Academy of Sciences\\
$^{2}$Institutes for Robotics and Intelligent Manufacturing, Chinese Academy of Sciences\\
$^{3}$University of Chinese Academy of Sciences  $^{4}$Nanjing University of Posts and Telecommunications\\
{\tt\small {\{liuyajing,zhoushijun,liuxiyao,haochunhui,tianjd\}@sia.cn, jobfbj@gmail.com}}
}

\begin{document}
\maketitle
\begin{abstract}
Single-source domain generalization (SDG) for object detection is a challenging yet essential task as the distribution bias of the unseen domain degrades the algorithm performance significantly. However, existing methods attempt to extract domain-invariant features, neglecting that the biased data leads the network to learn biased features that are non-causal and poorly generalizable. To this end, we propose an Unbiased Faster R-CNN (UFR) for generalizable feature learning. Specifically, we formulate SDG in object detection from a causal perspective and construct a Structural Causal Model (SCM) to analyze the data bias and feature bias  in the task, which are caused by scene confounders and object attribute confounders. Based on the SCM, we design a Global-Local Transformation module for data augmentation, which effectively simulates domain diversity and mitigates the data bias. Additionally, we introduce a Causal Attention Learning module that incorporates a designed attention invariance loss to learn image-level features that are robust to scene confounders. Moreover, we develop a Causal Prototype Learning module with an explicit instance constraint and an implicit prototype constraint, which further alleviates the negative impact of object attribute confounders. 
Experimental results on five scenes demonstrate the prominent generalization ability of our method,  with an improvement of  3.9\% mAP on the Night-Clear scene. 
\end{abstract}    
\section{Introduction}
\label{sec:intro}
\begin{figure}[t]
\centering
\includegraphics[scale=0.48]{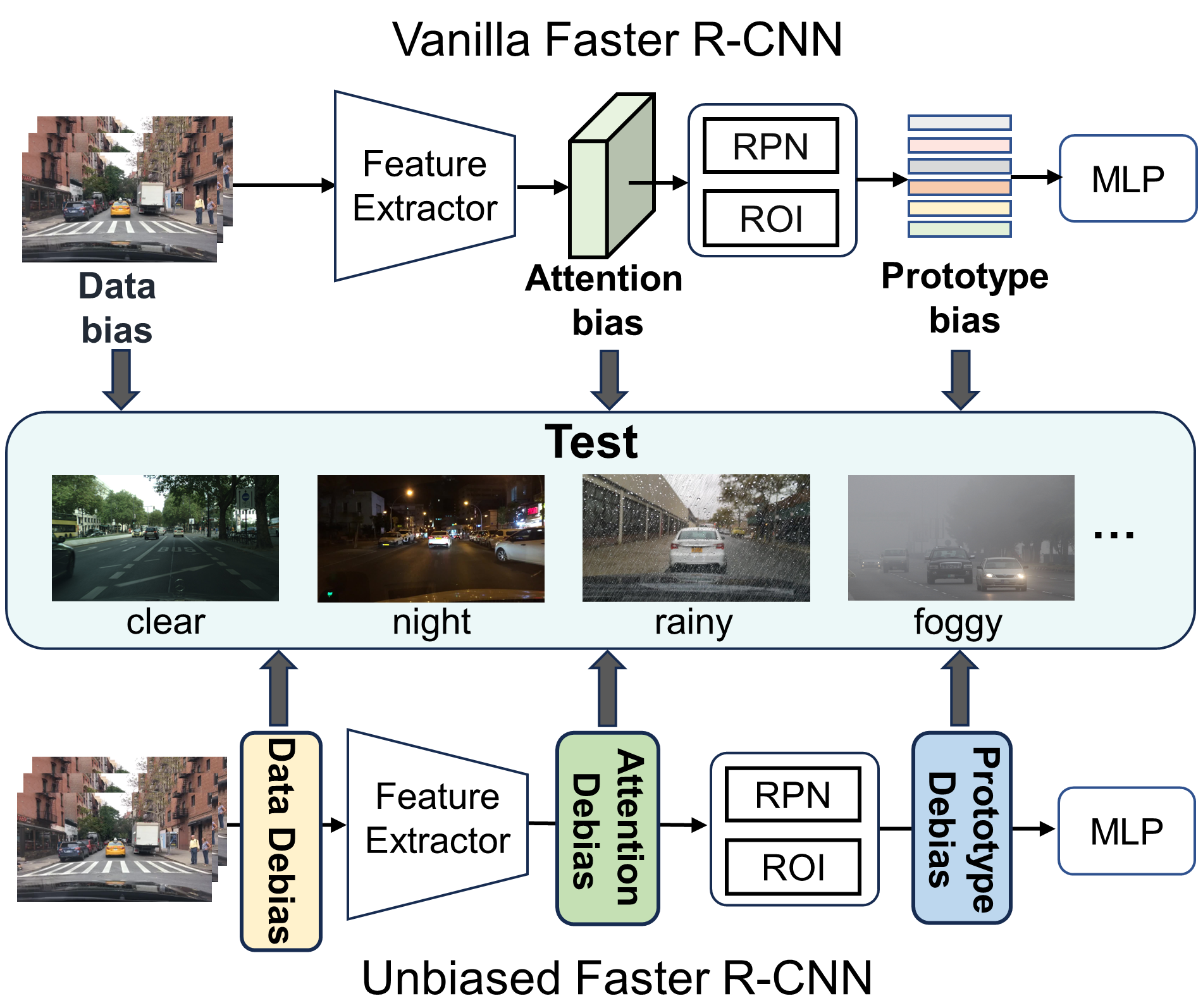}
\captionsetup{singlelinecheck=off}
\caption{Comparison between vanilla Faster R-CNN (FR) \cite{Ren2015Faster} (top) and our proposed Unbiased Faster R-CNN (bottom). For vanilla FR \cite{Ren2015Faster}, the biased distribution of the input data leads the network to learn biased features that favor the seen environment and are poorly generalizable to unseen test environments. The feature bias can be attributed to the image-level attention bias and object-level prototype bias. Our method mitigates the data bias in the input space and further learns unbiased attention and prototypes in the representation space.} 
\label{intro}
\vspace{-1.0em}
\end{figure}
The problem of distribution shift in unseen domains often arises during the deployment of perception systems, leading to a notable decline in the model's performance \cite{Chen2018Domain,Saito2019StrongWeak}. 
Consequently, domain generalization \cite{li2018deep,shao2019multi,du2020learning,zhao2021learning} has emerged as a branch of transfer learning, which aims at generalizing the knowledge from multiple source domains to the unseen target domain. Single-source domain generalization (SDG) is a special case of domain generalization where there is only one source domain \cite{qiao2020learning,wang2021learning,Zhou2022Domain} and it focuses on exploring the robustness of the model under different image corruptions. 

The existing two methods for SDG in object detection adopt different generalization strategies on Faster R-CNN \cite{Ren2015Faster}. The domain-invariant feature learning-based method  \cite{Wu2022SingleDomaina} explicitly decomposes domain-invariant features and domain-specific features by imposing constraints on the network, without relying on data augmentation techniques. And the data augmentation-based method  \cite{vidit2023clip} perturbs the data distribution and increases the diversity of input data to enhance model generalization ability.

\begin{figure}[t]
\centering
\includegraphics[scale=0.59]{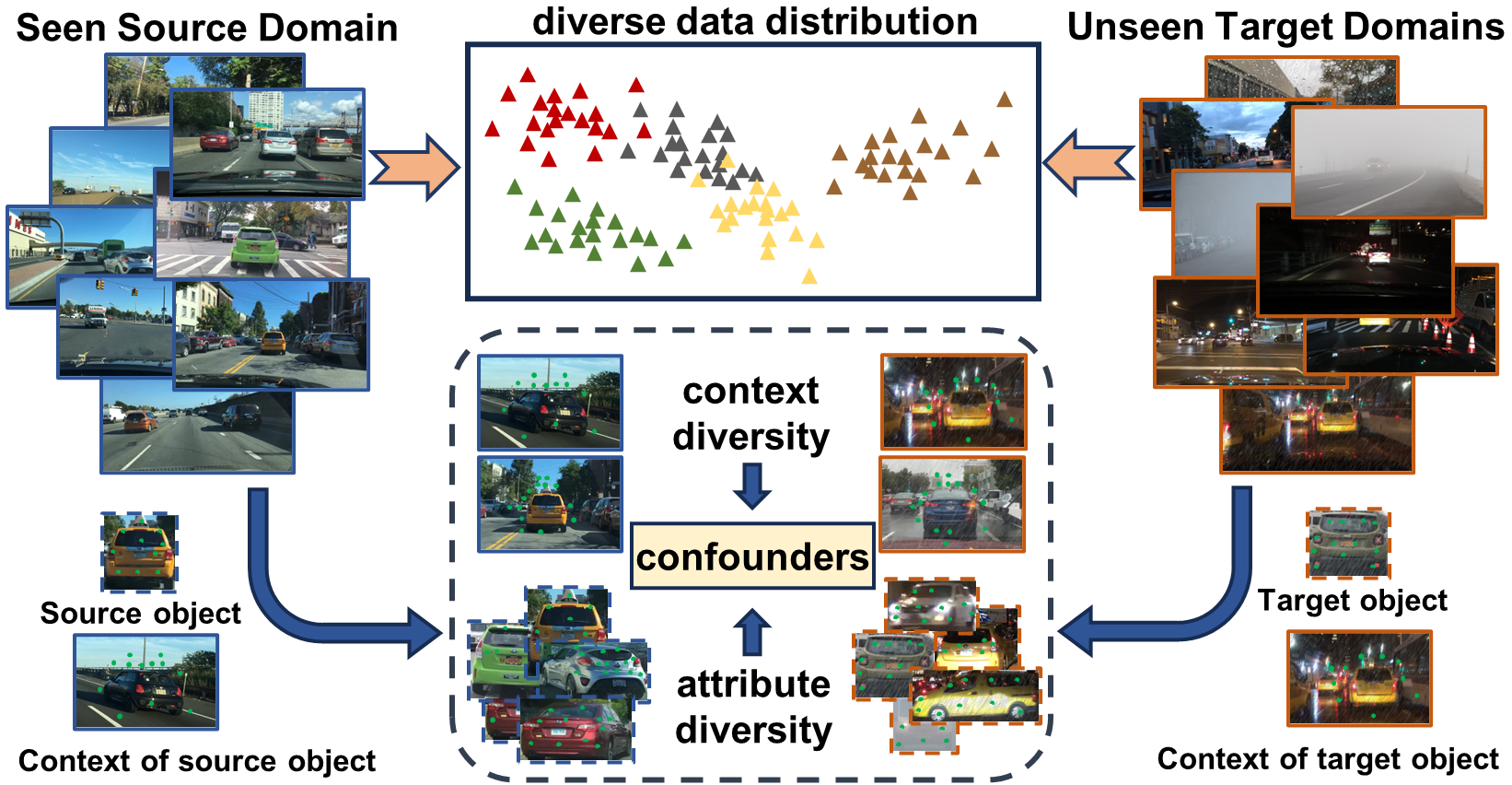}
\captionsetup{singlelinecheck=off}
\caption{Illustration of highly changeable data distribution, diverse context and object attributes in unseen target domains.} 
\label{intro2}
\vspace{-1.0em}
\end{figure}

However, both of these strategies have certain limitations. Firstly, it has been proven in \cite{xu2023multi} that domain-invariant features are inherently domain-dependent and biased, as features that are invariant to current domains may be variant to other domains. This problem can be attributed to the fact that domain-invariant features learned from a biased data distribution are not causal features and cannot adapt well to the unseen target environments. Secondly, the domain generalization methods that use only data augmentation without constraining network features fail to capture causal features from a rich augmented data distribution. This results in the network performing well in the augmented domains but remaining ineffective in the unseen domains, as confirmed in our Ablation Study in Section \ref{secablation}.
We summarize the reasons for the above two limitations as the existence of data bias in the input space and feature bias in the representation space. The feature bias is further decomposed into image-level attention bias and object-level prototype bias, as illustrated in Fig. \ref{intro} and Fig. \ref{intro2}:\\
\textbf{(1) Data bias.} The data distribution of unseen target domains is highly changeable, as shown in Fig. \ref{intro2}. Therefore, learning invariant features solely from single-domain data will exacerbate the statistical dependence \cite{pearl2000models} between the input data and labels, leading to biased learned results that favor the seen environments.\\
\textbf{(2) Attention bias.} When testing in unseen domains with complex context, as depicted in Fig. \ref{intro2}, the context features may confuse the object features and the network may exhibit attention bias, directing its focus towards the context of the object rather than the object itself. We refer to the scene context as \textbf{scene confounders}.\\
\textbf{(3) Prototype bias.} Each category possesses distinctive causal attributes, such as structural information, that are highly informative, as opposed to non-discriminative ones like view and color, which we refer to as \textbf{object attribute confounders}, as shown in Fig. \ref{intro2}. Therefore, if the learned category features are not constrained, the network may incorrectly take confounding attribute features that occur frequently in the source domain as category prototypes, which are biased and poorly generalizable.

To this end, we propose an Unbiased Faster R-CNN (UFR) model, as illustrated in Fig. \ref{intro}, with a comparison to vanilla Faster R-CNN \cite{Ren2015Faster}. The UFR model simulates diverse data distributions through data augmentation to mitigate the data bias and applies constraints on the features to learn scene-level causal attention and object-level causal prototype, thus eliminating the attention bias and prototype bias. 
Specifically, we define causality in the SDG task as the cause-and-effect relationship among scene, objects, causal features and non-causal features and construct a Structural Causal Model (SCM) to analyze the data bias and feature bias caused by scene confounders and object attribute confounders.
Additionally, we propose a Global-Local Transformation (GLT) module, which augments the data as a whole in the frequency domain and leverages the segmentation capability of SAM \cite{kirillov2023segment} to augment local objects in the spatial domain. To address scene confounders, we propose a Causal Attention Learning module with an attention invariance loss, which transforms the learning of scene-level causal features into the learning of causal attention.
To further learn causal features at the object level, we introduce a Causal Prototype Learning module, which encompasses an explicit constraint on the distribution of instance features and an implicit constraint on the relationship between prototypes.  
The effectiveness of our method is demonstrated through experiments conducted across five weather conditions.  The contributions  are summarized as follows:\\
\indent(1) We are the first to investigate single-domain generalized object detection from a causal perspective and analyze three biases that limit the generalization ability of the detector, including data bias, attention bias and prototype bias. \\
\indent(2) We construct a Structural Causal Model to analyze the biases caused by two types of confounders and further propose an Unbiased Faster R-CNN with a Global-Local Transformation module, a Causal Attention Learning module and a Causal Prototype Learning module to mitigate the data bias, attention bias and prototype bias respectively.\\
\indent(3) We evaluate our method on five different weather conditions to demonstrate its effectiveness and superiority and our method achieves a remarkable 3.9\% mAP improvement on the Night-Clear scene.

\section{Related Works}
\subsection{Domain Generalization}
Common strategies for addressing domain generalization problem include domain alignment \cite{li2018deep,shao2019multi, wang2021respecting,xu2023multi}, meta learning \cite{du2020learning,zhao2021learning,chen2023meta,Wang_2023_CVPR}, data augmentation \cite{Lee2022WildNet,Kang2022Style,vidit2023clip}, ensemble learning \cite{mancini2018best,seo2020learning,Peng2022SemanticAware}, self-supervised learning \cite{he2020momentum,grill2020bootstrap} and disentanglement learning \cite{wang2020cross,Lin2021DomainInvariant,zhang2022towards}.
As a special case of domain generalization, the solution for single-source domain generalization (SDG) can also be categorized into several of the aforementioned strategies. Many prior studies used data augmentation \cite{qiao2020learning,wang2021learning,vidit2023clip,li2021progressive} to generate out-of-distribution samples to extend the distribution of the source domain. For example, Wang et al. \cite{wang2021learning} proposed a style-complement module to generate diverse stylized images. Vidit et al. \cite{vidit2023clip} proposed a semantic augmentation method for SDG in object detection with the help of a pre-trained CLIP \cite{radford2021learning}. In addition, some works employed the feature normalization strategy \cite{Pan_2018_ECCV,Pan_2019_ICCV,Huang_2019_CVPR,Choi_2021_CVPR}  to learn domain-invariant features. For example, Fan et al. \cite{fan2021adversarially}  proposed  an ASR-Norm layer to learn standardization and rescaling statistics.
\subsection{Causality in Computer Vision}
Causal mechanism considers the fact that statistical dependence cannot reliably predict the labels of counterfactual inputs \cite{pearl2000models}. Thus, exploring causality \cite{buhlmann2020invariance} enables the acquisition of robust knowledge beyond what is supported by the observed data \cite{scholkopf2021towards}. Recently, causal mechanism has gathered significant attention in computer vision \cite{Zhang2020Causal,Yang2021CausalVAE,Hu2021Distilling,ouyang2022causality}. Many works use causal mechanism to tackle domain generalization \cite{Wang2023ContrastiveACE,xu2023multi,Li2023Deep,Lv2022Causalityb,qi2022class} and domain adaptation \cite{magliacane2018domain,Yue2021Transporting,jiang2022invariant,liu2023decoupled} problems. For example, Yue et al. \cite{Yue2021Transporting} applied the causal mechanism to domain adaptation and used the domain-invariant disentanglement to identify confounders. Lv et al. \cite{Lv2022Causalityb} proposed a representation learning framework for causality-inspired domain generalization. Besides, Liu et al. \cite{liu2023decoupled} proposed a Decoupled Unbiased Teacher to solve the source-free domain adaptation problem. Moreover, Xu et al. \cite{xu2023multi} introduced a causality-inspired data-augmentation strategy and eliminated non-causal factors by a Multi-view Adversarial Discriminator.  
In this paper, we first apply the causal mechanism to the single-source domain generalized object detection task and propose an Unbiased Faster R-CNN to mine causal features at the image level and object level.
\section{Structural Causal Model}
\begin{figure}[t]
\centering
\includegraphics[scale=0.6]{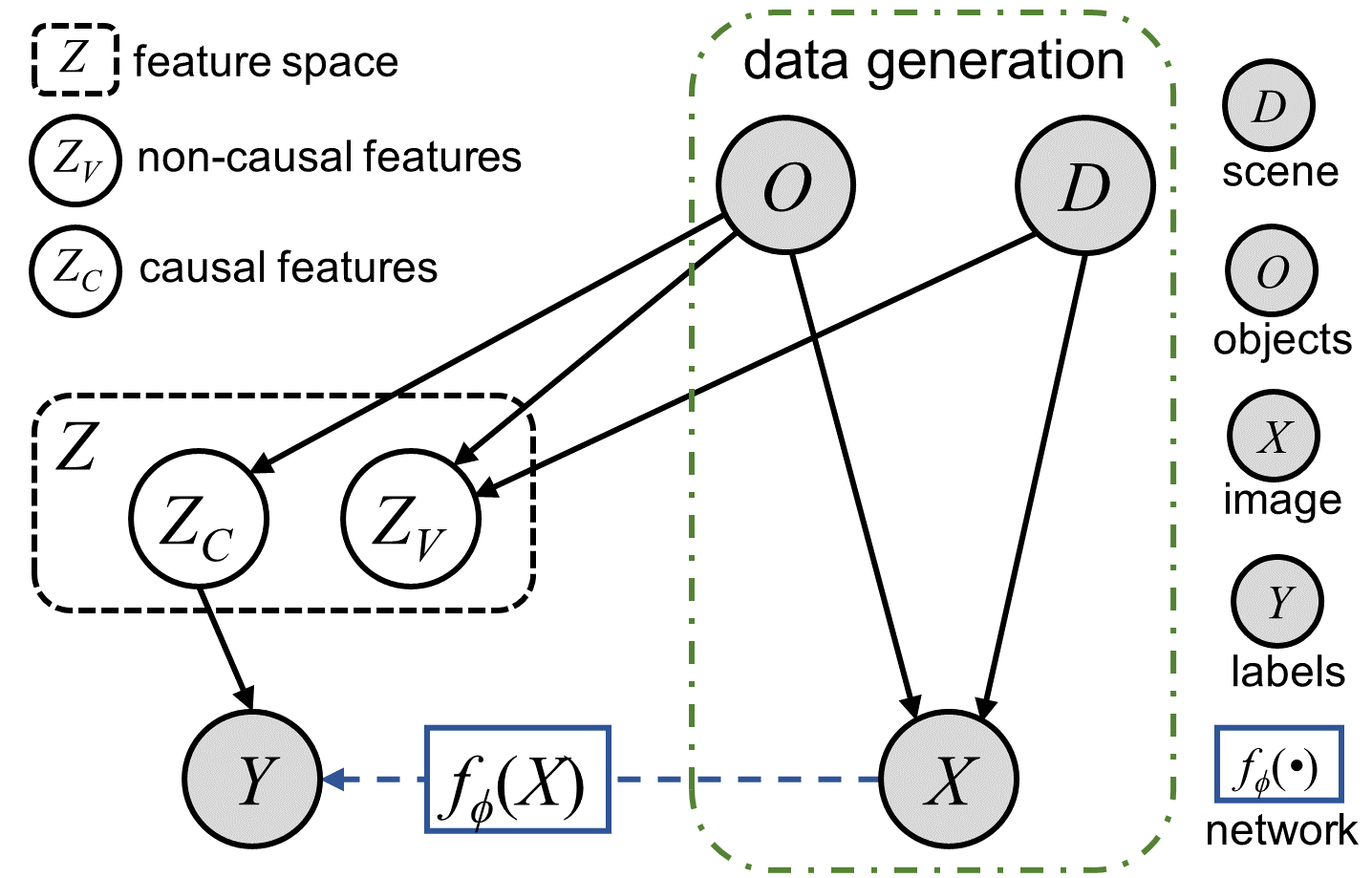}
\captionsetup{singlelinecheck=off}
\caption{The constructed Structural Causal Model (SCM) for the object detection task. The nodes denote variables, the solid arrows denote the direct causal effect and the dashed arrow indicates that there exists data dependence.}
\label{scm}
\vspace{-0.7em}
\end{figure}

Consider the data (e.g. images) from the observed environment (source domain) as $X$ and its target (e.g. detection labels) as $Y$, the objective of single-domain generalized object detection is to generalize the model trained with $(X, Y)$ in the observed environment to unseen environments (target domains). We represent images from a causal perspective and construct a Structural Causal Model (SCM) in Fig. \ref{scm} to describe the cause-and-effect relationship in object detection task and attempt to eliminate the data bias and feature bias caused by the scene confounders and object attribute confounders. The rationale behind the SCM is as follows:\\
(1) $\bm{O}\rightarrow \bm{X} \leftarrow \bm{D}$ denotes that an image is composed of scene $D$ and a set of objects $O$. $D$ is scene confounder, and the non-discriminative attributes of  $O$ are object attribute confounders. The GLT module is achieved by perturbing the scene confounders and object attribute confounders. \\
(2) $\bm{O}\rightarrow \bm{Z_V} \leftarrow \bm{D}$ denotes that the non-causal features $Z_V$ are derived from two components, including the scene confounders and non-causal object attribute confounders.\\
(3) $\bm{O}\rightarrow \bm{Z_C}\rightarrow \bm{Y}$ represents that the causal features $Z_C$ are determined by the discriminative attributes of objects, such as shape. And the prediction labels $Y$ are derived from the causal features $Z_C$. The complete feature space $Z$ consists of non-causal features $Z_V$ and causal features $Z_C$.\\
(4) $\bm{X}\rightarrow \bm{f_{\phi}(X)}\rightarrow \bm{Y}$ models the data stream of the object detection network parameterized by $\phi$. 

\begin{figure*}[h]
\centering
\includegraphics[scale=0.7]{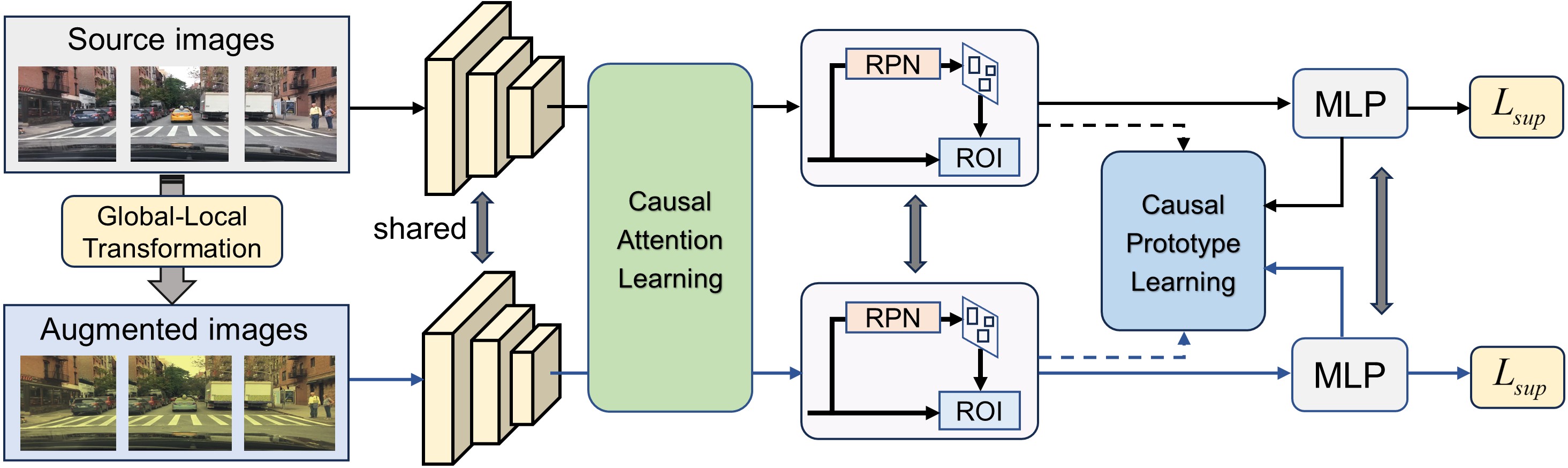}
\captionsetup{singlelinecheck=off}
\caption{The overall structure of the proposed Unbiased Faster R-CNN. The input source images are augmented through the Global-Local Transformation (GLT) module. Both the original images and augmented images are fed into the network for training. During training, the role of the Causal Attention Learning module is to constrain the network to learn scene-level causal attention and select causal features to feed into the RPN. The purpose of the Causal Prototype Learning module is to constrain the network to learn object-level causal features with the help of an explicit instance constraint (solid arrows) and an implicit prototype constraint (dashed arrows).}
\label{net}
\vspace{-0.7em}
\end{figure*}
Based on the constructed SCM, the ideal feature mapping of the network is $f_{\phi}(X)=Z_C$, which is also formulated as:
\begin{equation}
    f_{\phi}(X^{(a_l, d_l)}) =  f_{\phi}(X^{(a_k,d_k)}),
    \label{invariant}
\end{equation}
where $a_l \neq a_k$, $d_l \neq d_k$ and the superscript $(a_l, d_l)$ denotes assigning scene and object attributes the distribution $d_l$ and $a_l$, respectively. Eq. (\ref{invariant}) indicates that the ideal representation of an image learned by the model is invariant under different data augmentations. 

To achieve the above goal, we decompose the objective into image-level attention invariance learning and object-level prototype invariance learning and further propose an Unbiased Faster R-CNN model, as depicted in Fig. \ref{net}, and the details of which are illustrated in Section \ref{UFR}.

\section{Unbiased Faster R-CNN}
\label{UFR}
\subsection{Global-Local Transformation}
\begin{figure}[t]
\centering
\includegraphics[scale=0.4]{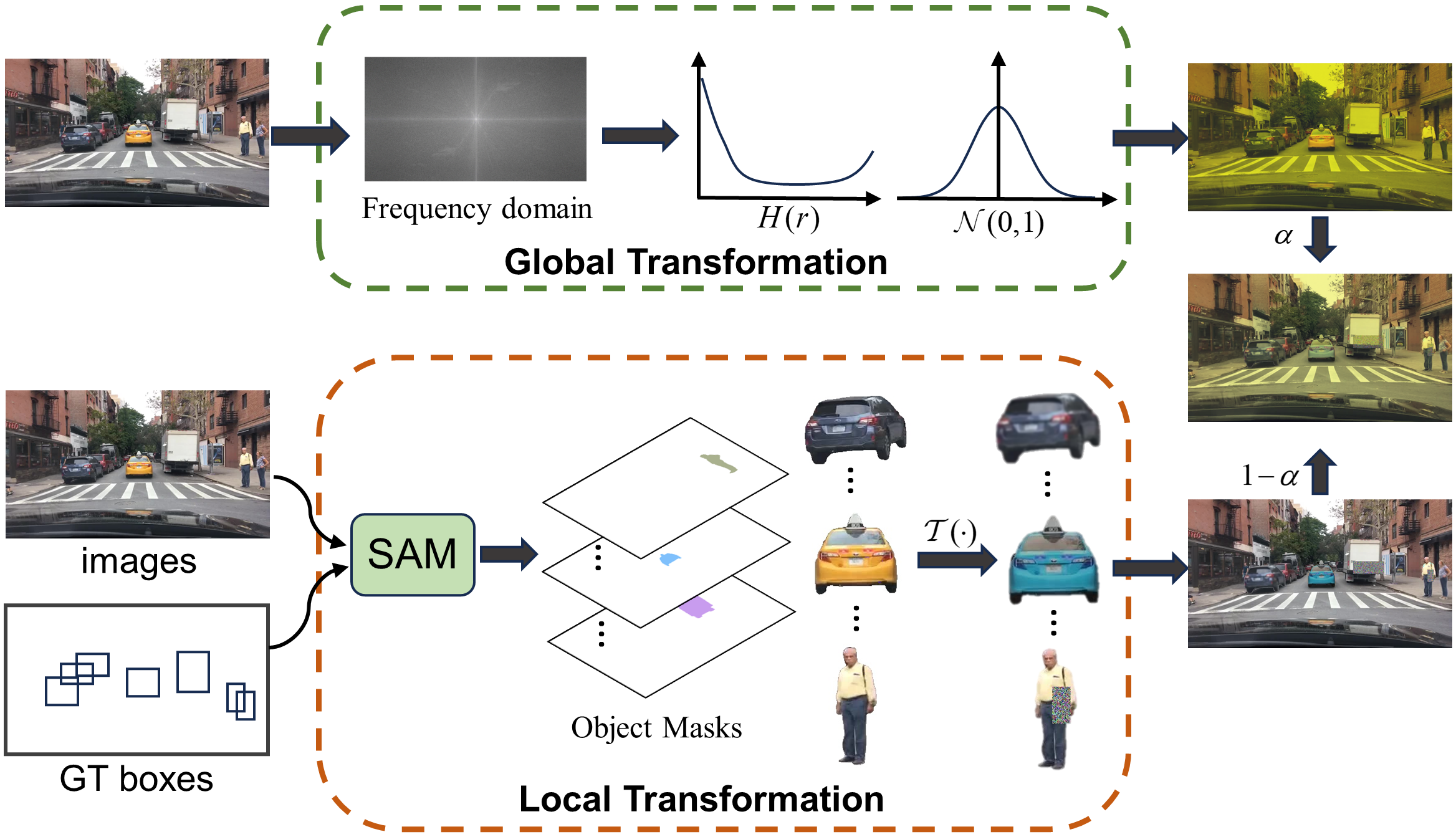}
\captionsetup{singlelinecheck=off}
\caption{Overview of  the Global-Local Transformation (GLT) module. The Global Transformation (GT) performs overall augmentation of an image in the frequency domain. And The Local-Transformation (LT) performs augmentation of local objects obtained by SAM \cite{kirillov2023segment} in the spatial domain. The final augmented image is obtained by fusing the GT image with the LT image.}
\label{GLT}
\vspace{-0.7em}
\end{figure}

The complete process of Global-Local Transformation is depicted in Fig. \ref{GLT}, which consists of Global Transformation (GT) in the frequency domain and Local Transformation (LT) in the spatial domain.

Inspired by the frequency-domain augmentation in \cite{xu2023multi}, the global transformation is formulated as:
\begin{equation}
    \text{GT}(x)=\mathcal{F'}(G(H(r)\cdot\mathcal{F}(x))+(1-H(r))\cdot\mathcal{F}(x)),
\end{equation}
where $\mathcal{F}$ denotes the Fourier Transformation and $\mathcal{F'}$ is the inverse one. $H(r)$ is the band-pass filter with a filter radius of $r$. $G(\cdot)$ is a randomization function according to a Gaussian distribution and $G(X)=X\cdot(1+\mathcal{N}(0,1))$.

For local transformation, as shown in Fig. \ref{GLT}, we first obtain the object masks with the help of Segment Anything Model (SAM) \cite{kirillov2023segment}:
\begin{equation}
    \text{SAM}(x,B)=m^O,
\end{equation}
where $x$ is the input image, $B$ is the bounding box set of the objects and $m^O$ is the set of the obtained masks, where $O=\{o_1,o_2,...o_{n_O}\}$ and $n_O$ is the number of objects in the input image. Then we extract object images by :
\begin{equation}
    x^{o_k}=x\odot m^{o_k}.
\end{equation}
Then the local transformation is formulated as:
\begin{equation}
    \text{LT}(x)=\mathcal{T}_{0}(x^{bg})+\sum_{k=1}^{n_O}\mathcal{T}_{k}(x^{o_k}),
\end{equation}
where $x^{bg}$ is the background image and $\mathcal{T}_{0}(\cdot)$ and $\mathcal{T}_{k}(\cdot)$ denote the randomly selected augmentation strategies for the background $x^{bg}$ and the object $x^{o_k}$.

The augmented image output by the GLT module is denoted as:
\begin{equation}
    \text{GLT}(x)=\alpha\cdot\text{GT}(x)+(1-\alpha)\cdot\text{LT}(x),
    \label{glt}
\end{equation}
where $\alpha$ is the fusion weight to balance the GT image and the LT image and the fusion strategy also increases the diversity of the augmented images.
\subsection{Causal Attention Learning}
We transform the learning of scene-level causal features into the learning of causal attention, which eliminates the need for explicit decoupling of causal features from non-causal features, instead focusing on selecting causal features based on attention.

Specifically, it is desirable for the network to capture object features precisely in different scene context. Thus, we enforce the feature attention maps of different data distribution images output by the backbone to converge and the attention invariance loss is defined as:
\begin{gather}
    \mathcal{L}_{att}=\text{Dice}(g(F_{att}^{(a_0,d_0)}),g(F_{att}^{(a_k,d_k)})),\\
    F_{att}^{(a,d)}=\sigma(E(x^{(a,d)})),
\end{gather}
where $g(\cdot)$ is a pixel-level binarization function, $F_{att}^{(a,d)}$ is the calculated attention map, $\sigma(\cdot)$ is the Sigmoid function, $E$ denotes the backbone, $x^{(a_0,d_0)}$ is the original input image and $x^{(a_k,d_k)}$ denotes the randomly augmented image obtained from GLT. Besides, $\text{Dice}(\cdot,\cdot)$ denotes the dice loss which is commonly used for measuring the regional similarity between two samples, which is defined as: 
\begin{equation}
    \text{Dice}(X_1,X_2) = 1-\frac{2\lvert X_1\cap X_2\lvert +1}{\lvert X_1\lvert +\lvert X_2\lvert +1},
\end{equation}
where $X_1$ and $X_2$ are two binary maps. The regions with a value of 1 for the maps are salient attention regions, while the regions with a value of 0 are non-salient attention regions. 

Then we select causal features according to the attention map for the subsequent Region Proposal Network (RPN) to generate object proposals $P$: 
\begin{equation}
    P = \text{RPN}(E(x^{(a,d)})\odot F_{att}^{(a,d)}).
    \label{rpn}
\end{equation}
\subsection{Causal Prototype Learning}
To facilitate the learning of object-level causal features, we introduce the Causal Prototype Learning module which encompasses  an explicit constraint and an implicit constraint. 

The explicit constraint is imposed on ROI features extracted from different data distributions.
Specifically, given the proposal set $P$ generated from the source image, we select a group of proposals with a confidence higher than the threshold $t$, denoted as $P(t)$. We represent the ROI features generated from $P(t)$ as $f(x, P(t))$ for simplicity. Then the explicit constraint is defined as:
\begin{equation}
    \mathcal{L}_{exp}=KL(p^{do(a_0,d_0)},p^{do(a_k,d_k)}),
    \label{explicit}
\end{equation}
where $KL$ denotes the Kullback-Leibler divergence, $p^{do(a_0,d_0)}$ is the short for $p(y|f(x^{(a_0,d_0)},P(t)))$, $p^{do(a_k,d_k)}$ is the short for $p(y|f(x^{(a_k,d_k)},P(t)))$and $p^{do(a,d)} = \text{MLP}(f(x^{(a,d)},P(t)))$. 

The explicit constraint $\mathcal{L}_{exp}$ encourages the within-class distance of the learned class representations from different data distributions to be concentrated and also gives supervisory information to the data-augmented image at the object feature level, which enhances the saliency of the object region, thereby improving the object localization performance of the data-augmented images. 

On the other hand, the implicit invariance constraint constrains the relationship between category prototypes across different data distributions. We hypothesize that the distance between causal prototypes of the same category for different data distributions is smaller compared to the distance between prototypes of different categories:
\begin{equation}
dist(v_{c_i}^{(a_0,d_0)},v_{c_j}^{(a_k,d_k)})>dist(v_{c_i}^{(a_0,d_0)},v_{c_i}^{(a_k,d_k)})\approx0,
\end{equation}
where $i\neq j$, $dist(\cdot)$ is a distance metric function, $v_{c_i}^{(a_0,d_0)}$ is the prototype of category $c_i$ of the source data distribution and $v_{c_j}^{(a_k,d_k)}$ is the prototype of category $c_j$ of the augmented data distribution. The prototype $v_c$ is computed by dynamically averaging the ROI features belonging to the category $c$.

We transform the implicit constraint that satisfy the above hypothesis into a prototype contrastive loss: 
\begin{gather}
     \mathcal{L}_{imp}=-\sum_i\text{log}\frac{\text{exp}(s(i,i))}{\text{exp}(s(i,i))+\sum_j \text{exp}(s(i,j))},\\
    s(i,j)={sim(v^{(a_0,d_0)}_{c_i},v^{(a_k,d_k)}_{c_j})/\tau},
 \label{implicit}
\end{gather}
where $i\neq j$, $sim(\cdot,\cdot)$ is the cosine similarity and $\tau$ is a temperature parameter.

The loss function of CPL module is presented as:
\begin{equation}
    \mathcal{L}_{prot}=\mathcal{L}_{exp}+\mathcal{L}_{imp}.
\end{equation}
The constraints of this module will further refine the image-level attention, thereby allowing the features fed into the RPN module to contain more discriminative information about the objects.
\subsection{Model Optimization}
The data streams of the original images and the augmented images share the network parameters and the total training loss is:
\begin{equation}
    \mathcal{L}=\mathcal{L}_{sup} + \lambda_1\mathcal{L}_{att} + \lambda_2\mathcal{L}_{prot},
    \label{total}
\end{equation}
where $\mathcal{L}_{sup}$ is the supervised object detection loss for the original images and the augmented images and $\lambda_{1/2}$ are hyperparameters.

During inference, our UFR model maintains the same parameter size as vanilla Faster R-CNN \cite{Ren2015Faster}, ensuring that it does not introduce additional space complexity. However, a key difference lies in the calculation of features that are fed into the RPN network, which is determined by Eq. (\ref{rpn}).
\section{Experiments}
\label{sec:formatting}
\subsection{Experimental Setup}
\textbf{Datasets.} We conduct experiments on the dataset built in \cite{Wu2022SingleDomaina}. The dataset consists of five different weather conditions, including Daytime-Clear, Daytime-Foggy, Dusk-Rainy, Night-Clear  and Night-Rainy. The Daytime-Clear scene is used as the source domain, comprising 19,395 images for training and 8,313 images for testing. 
The other four scenes are used as unseen target domains, consisting of 3,775 images in Daytime-Foggy condition, 3,501 images in Dusk-Rainy condition, 26,158 images in Night-Clear  condition and 2,494 images in Night-Rainy condition. The dataset contains annotations for seven categories of objects, including person, car, bike, rider, motor, bus and truck. 

\begin{figure*}[t]
\centering
\subfloat
{
    \begin{minipage}[b]{.24\linewidth}
        \centering
        \includegraphics[scale=0.13]{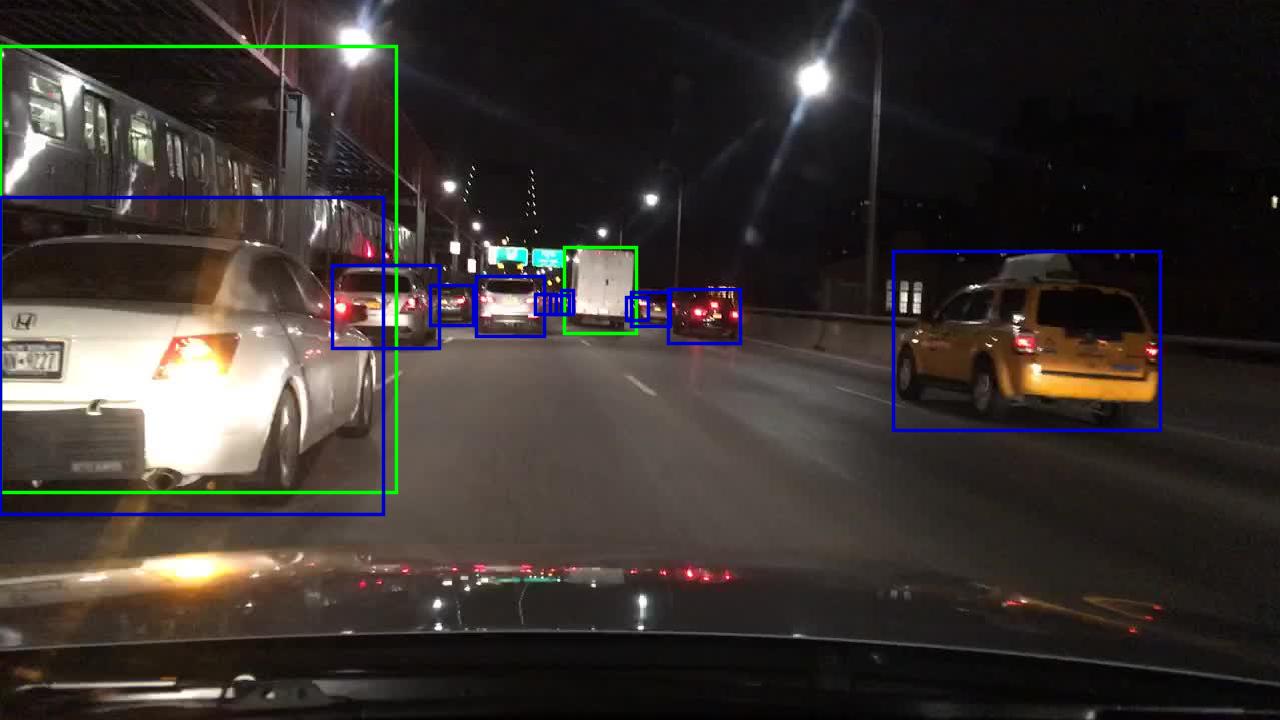}
    \end{minipage}
    \begin{minipage}[b]{.24\linewidth}
        \centering
        \includegraphics[scale=0.13]{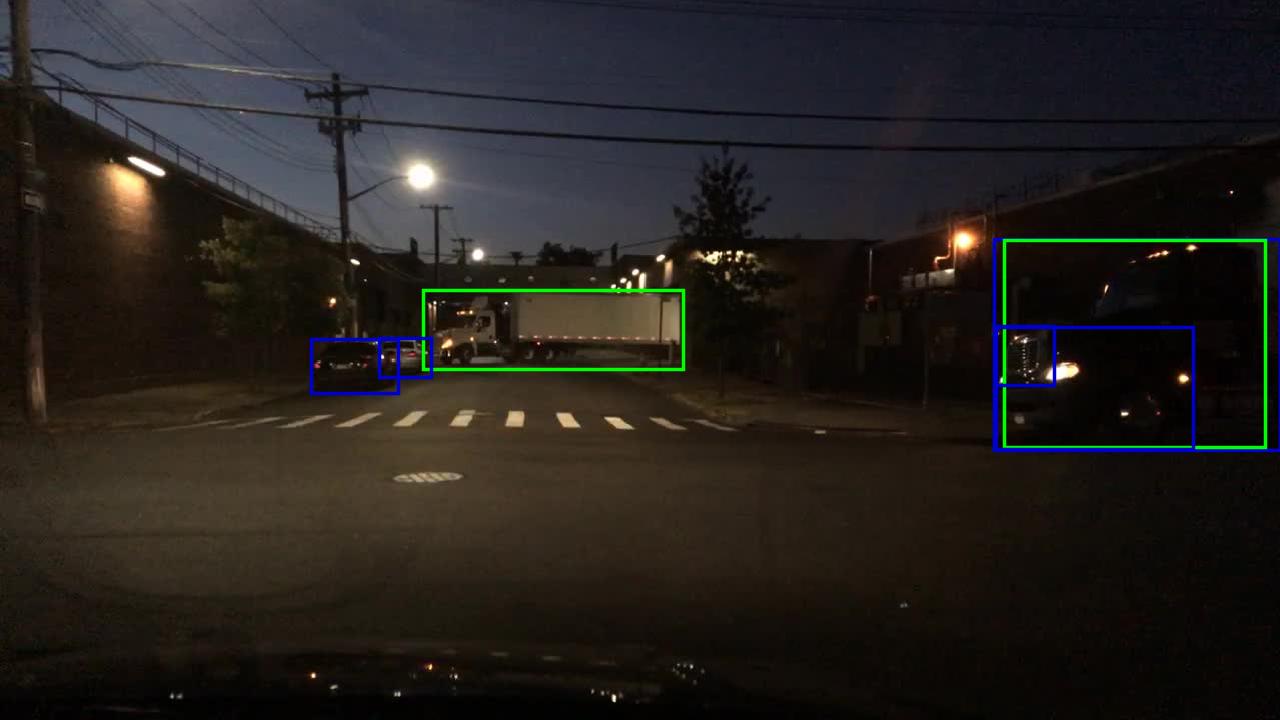}
    \end{minipage}
    \begin{minipage}[b]{.24\linewidth}
        \centering
        \includegraphics[scale=0.13]{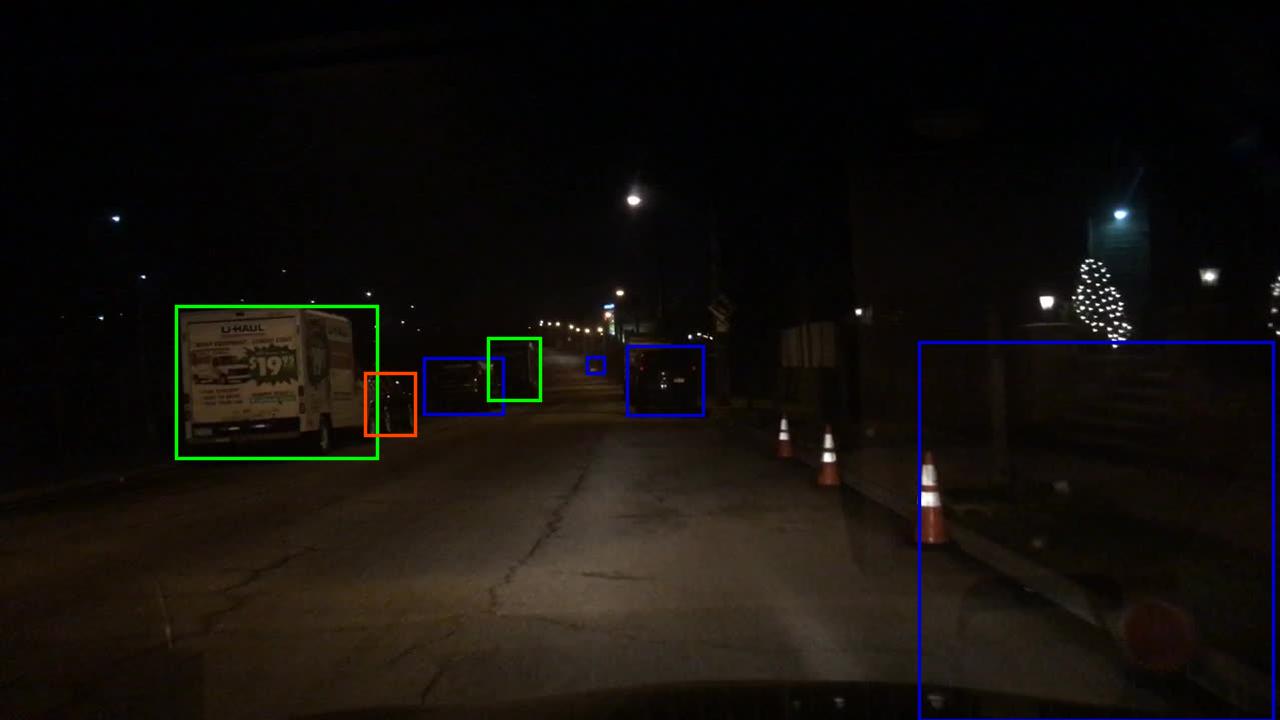}
    \end{minipage}
    \begin{minipage}[b]{.24\linewidth}
        \centering
        \includegraphics[scale=0.13]{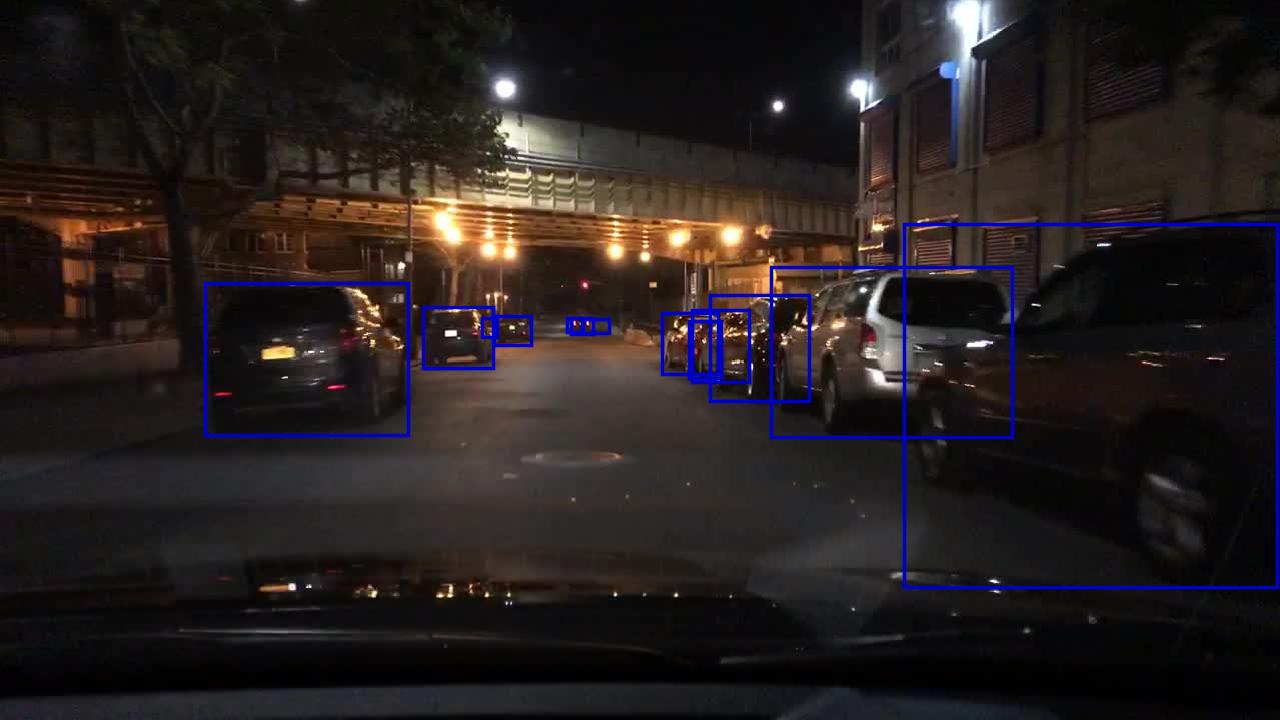}
    \end{minipage}
}

\subfloat
{
    \begin{minipage}[b]{.24\linewidth}
        \centering
        \includegraphics[scale=0.13]{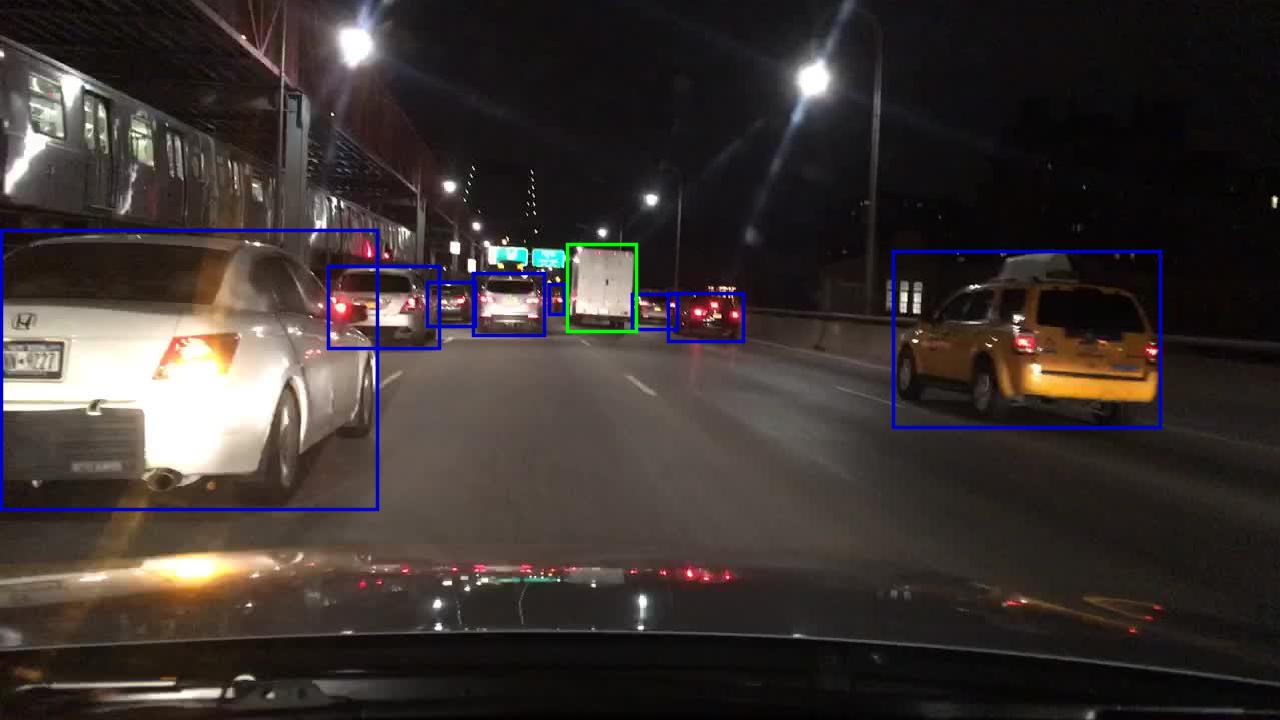}
    \end{minipage}
    \begin{minipage}[b]{.24\linewidth}
        \centering
        \includegraphics[scale=0.13]{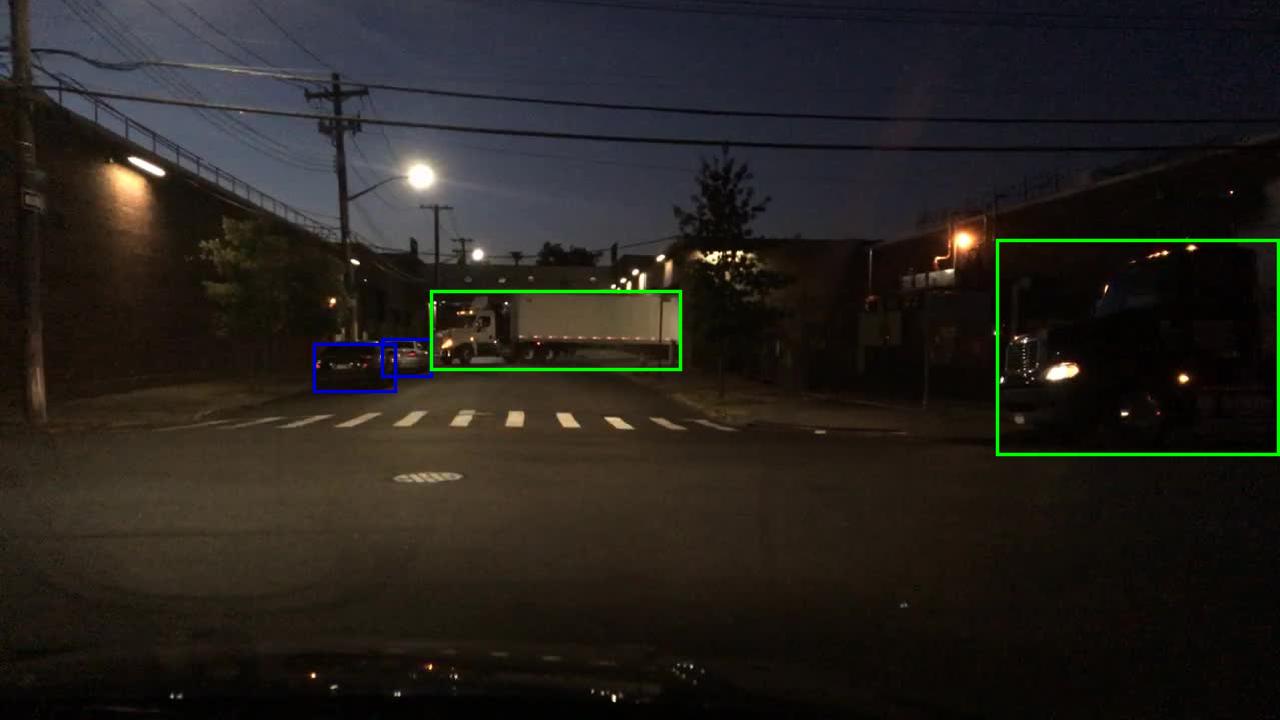}
    \end{minipage}
    \begin{minipage}[b]{.24\linewidth}
        \centering
        \includegraphics[scale=0.13]{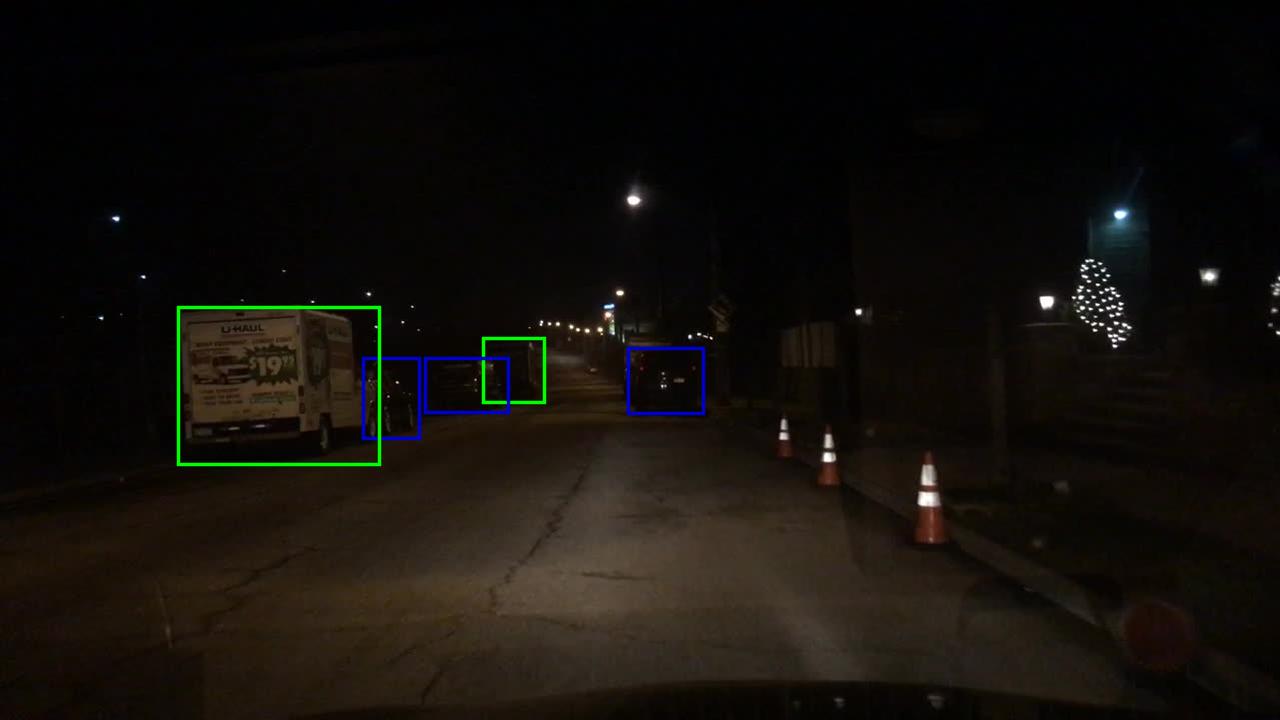}
    \end{minipage}
    \begin{minipage}[b]{.24\linewidth}
        \centering
        \includegraphics[scale=0.13]{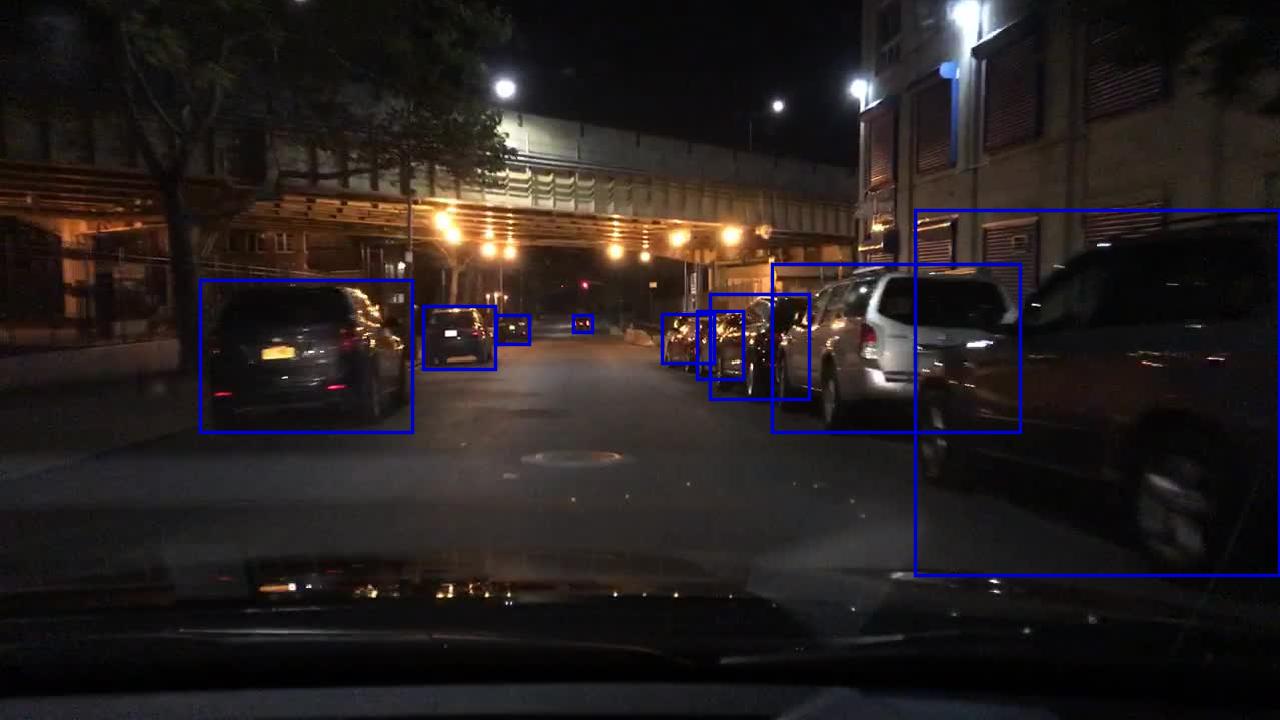}
    \end{minipage}
}
\caption{Qualitative evaluation results of the model's generalization ability on the Night-Clear scene. The \textbf{top-row} images are the results of \textbf{vanilla Faster R-CNN} \cite{Ren2015Faster}. The \textbf{bottom-row} images are the results of \textbf{our method}. }
\label{nightfig}
\vspace{-0.6em}
\end{figure*}
\noindent \textbf{Implementation Details.} We adopt Faster R-CNN \cite{Ren2015Faster} with ResNet-101 \cite{he2016deep} as the object detector. The backbone is initialized with weights pre-trained on ImageNet \cite{deng2009imagenet}. We train the model with Stochastic Gradient Descent (SGD) optimizer with a momentum of 0.9 for 80$k$ iterations. During training, the learning rate is set to 0.001, the batch size is set to 4. Besides, the threshold $t$ is set to 0.7, the $\lambda_1$ and $\lambda_2$ are both set to 0.1.\\
\noindent \textbf{Data Augmentation Setting.} For local transformation in spatial domain, we randomly apply the augmentation strategies consisting of gaussian blurring, color jittering, random erasing and grayscale. The fusion weight $\alpha$ in Eq. (\ref{glt}) is a random scalar in $[0,1]$. 
\newcolumntype{g}{>{\columncolor{Gray}}c}
\begin{table}[t]
\Large
\resizebox{\linewidth}{!}{
    \begin{tabular}{c|ccccccc|g}
        \toprule[1.5pt]
Methods &Bus &Bike & Car	&Motor	&\makebox[0.01\textwidth][c]{Person}	&Rider	&\makebox[0.01\textwidth][c]{Truck}	&mAP \\
\hline
FR \cite{Ren2015Faster} &66.9	&45.9	&69.8	&46.5	&\textbf{50.6}	&49.4	&64.0	&56.2\\
SW	\cite{Pan_2019_ICCV} &62.3	&42.9	&53.3	&49.9	&39.2	&46.2	&60.6	&50.6\\
IBN-Net \cite{Pan_2018_ECCV}	&63.6	&40.7	&53.2	&45.9	&38.6	&45.3	&60.7	&49.7\\
IterNorm \cite{Huang_2019_CVPR}	&58.4	&34.2	&42.4	&44.1	&31.6	&40.8	&55.5	&43.9\\
ISW \cite{Choi_2021_CVPR}	&62.9	&44.6	&53.5	&49.2	&39.9	&48.3	&60.9	&51.3\\
SDGOD \cite{Wu2022SingleDomaina}	&\textbf{68.8}	&50.9	&53.9	&\textbf{56.2}	&41.8	&\textbf{52.4}	&\textbf{68.7}	&56.1\\
CLIP-Gap \cite{vidit2023clip}	&55.0	&47.8	&67.5	&46.7	&49.4	&46.7	&54.7	&52.5\\
\hline\hline
UFR (Ours)	&66.8	&\textbf{51.0}	&\textbf{70.6}	&55.8	&49.8	&48.5	&67.4	&\textbf{58.6}\\
\bottomrule[1.5pt]
    \end{tabular}
    }
\caption{Quantitative results (\%) on the Daytime-Clear scene.}
\centering
\label{dayclear}
\vspace{-0.8em}
\end{table}
\subsection{Comparison with State-of-the-arts}
Following the setting in \cite{Wu2022SingleDomaina}, we use the Mean Average Precision (mAP) as our metric and report the mAP@0.5 results. We compare our method with feature normalization-based SDG methods, including SW \cite{Pan_2019_ICCV}, IBN-Net \cite{Pan_2018_ECCV}, IterNorm \cite{Huang_2019_CVPR}, and ISW \cite{Choi_2021_CVPR}, and the existing two single domain generalized object detection methods, including SDGOD  \cite{Wu2022SingleDomaina} and CLIP-Gap \cite{vidit2023clip}. Besides, we also compare the performance of our model with vanilla Faster R-CNN (FR) \cite{Ren2015Faster}. 

\noindent \textbf{Results on Daytime-Clear Scene.}
We evaluate the model performance on the source domain. As shown in Table \ref{dayclear}, our method achieves an optimal result of 58.6\% mAP and has a 2.4\% mAP gain against vanilla FR \cite{Ren2015Faster}. It shows that our method is able to maintain or even improve the performance of the FR \cite{Ren2015Faster} model in the source domain while improving its generalization ability in unseen domains.
\begin{table}[t]
\Large
\resizebox{\linewidth}{!}{
    \begin{tabular}{c|ccccccc|g}
        \toprule[1.5pt]
Methods &Bus &Bike & Car	&Motor	&\makebox[0.01\textwidth][c]{Person}	&Rider	&\makebox[0.01\textwidth][c]{Truck}	&mAP \\
\hline
FR \cite{Ren2015Faster}	&43.5	&31.2	&49.8  &17.5 &36.3	&29.2	&43.1 &35.8 \\
SW \cite{Pan_2019_ICCV}	&38.7&	29.2	&49.8	&16.6	&31.5	&28&	40.2&	33.4\\
IBN-Net \cite{Pan_2018_ECCV}	&37.8	&27.3	&49.6	&15.1	&29.2	&27.1	&38.9	&32.1\\
IterNorm \cite{Huang_2019_CVPR}	&38.5	&23.5	&38.9	&15.8	&26.6	&25.9	&38.1	&29.6\\
ISW \cite{Choi_2021_CVPR}	&38.5	&28.5	&49.6	&15.4	&31.9	&27.5	&41.3	&33.2\\
SDGOD \cite{Wu2022SingleDomaina}	&40.6	&35.1	&50.7	&\textbf{19.7}	&34.7	&\textbf{32.1}&43.4	&36.6\\
CLIP-Gap \cite{vidit2023clip}	&37.7	&34.3	&58.0 &	19.2	&37.6	&28.5	&42.9	&36.9\\
\hline\hline
Ours	&\textbf{43.6}	&\textbf{38.1}	&\textbf{66.1}	&14.7	&\textbf{49.1}	&26.4	&\textbf{47.5}	&\textbf{40.8}\\
\bottomrule[1.5pt]
    \end{tabular}
    }
\caption{Quantitative results (\%) on the Night-Clear scene.}
\centering
\label{nightclear}
\vspace{-0.8em}
\end{table}

\noindent\textbf{Results on Night-Clear Scene.} The qualitative and quantitative evaluation results of the model's generalization performance in the Night-Clear  scene are shown in Table \ref{nightclear} and Fig. \ref{nightfig}, respectively. As shown in Table \ref{nightclear}, our method achieves the best result of 40.8\% mAP, outperforming Clip-Gap \cite{vidit2023clip} by 3.9\% mAP. Besides, compared with vanilla FR \cite{Ren2015Faster}, our method achieves a 5.0\% mAP gain, which indicates that the learned features is beneficial for improving model generalization ability. Besides, compared with SDGOD \cite{Wu2022SingleDomaina} that learns domain-invariant features, our method has an improvement of 4.2\% mAP, which suggests that causal features are more discriminative features and can generalize better. 
For qualitative results demonstrated in Fig. \ref{nightfig}, it is obviously observed that our method detects objects more accurately and has fewer false positives compared with vanilla FR \cite{Ren2015Faster}.

\begin{figure*}[t]
\centering
\subfloat
{
    \begin{minipage}[b]{.24\linewidth}
        \centering
        \includegraphics[scale=0.13]{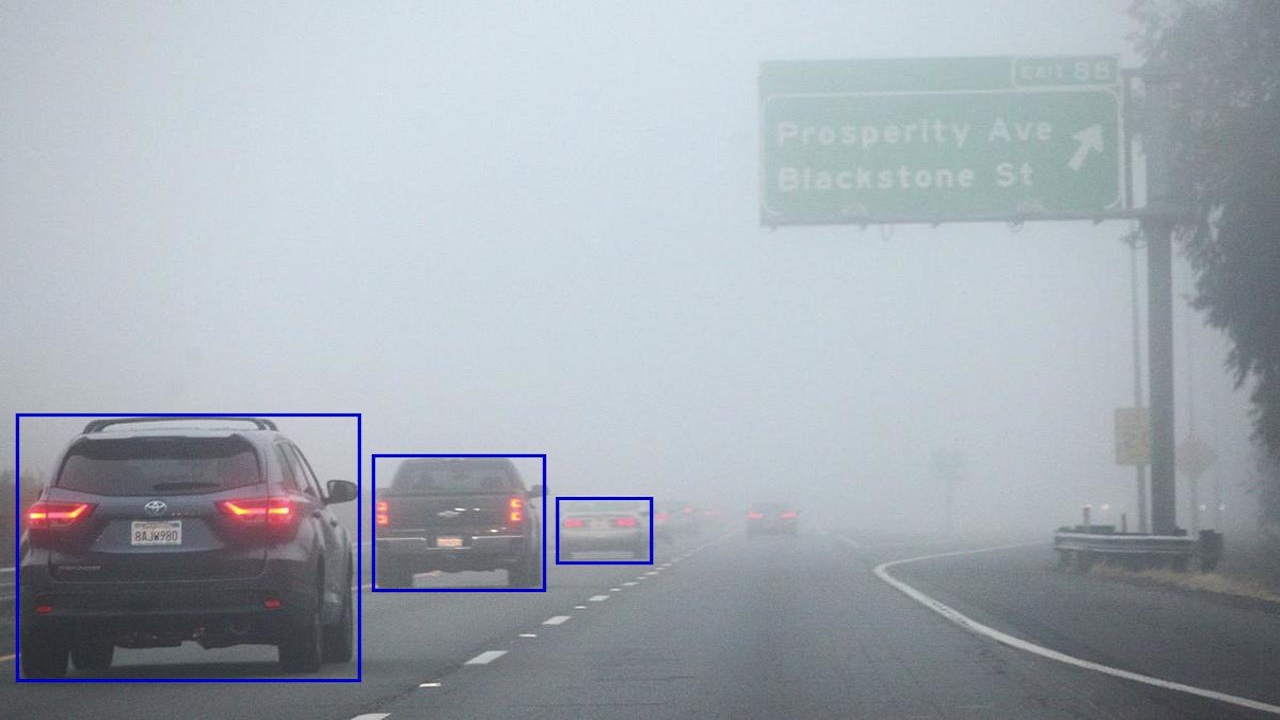}
    \end{minipage}
    \begin{minipage}[b]{.24\linewidth}
        \centering
        \includegraphics[scale=0.13]{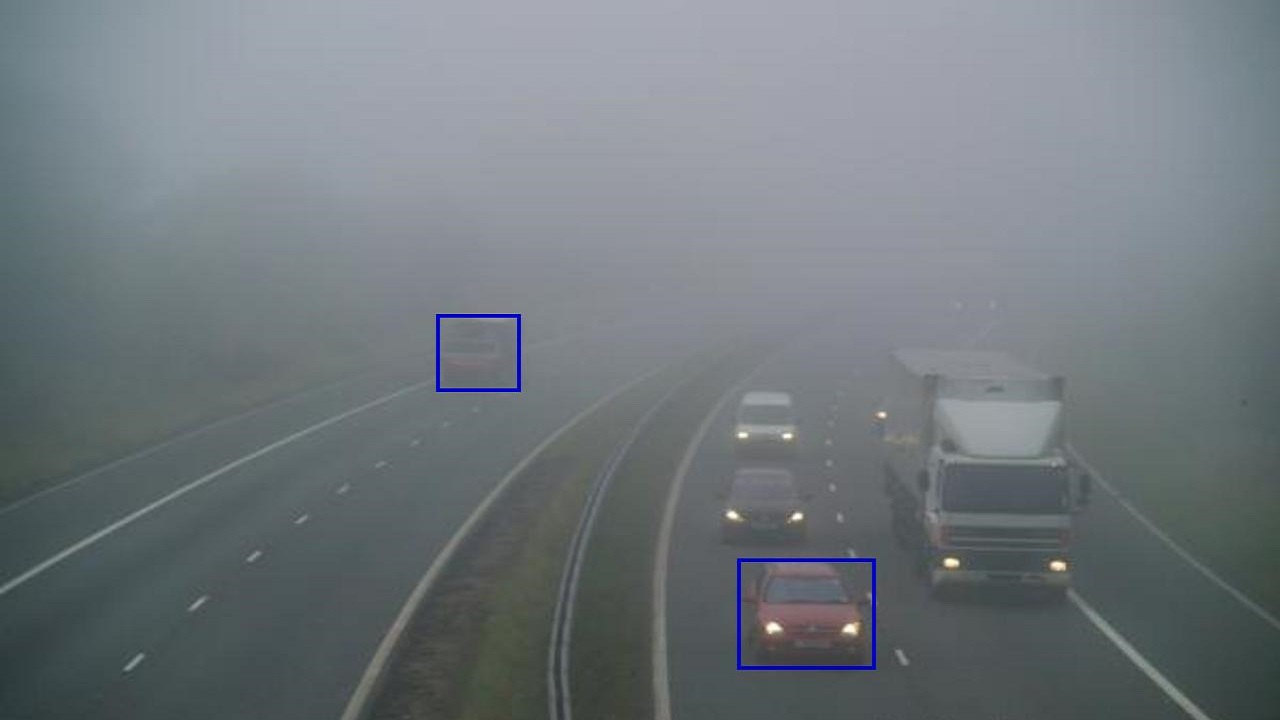}
    \end{minipage}
    \begin{minipage}[b]{.24\linewidth}
        \centering
        \includegraphics[scale=0.13]{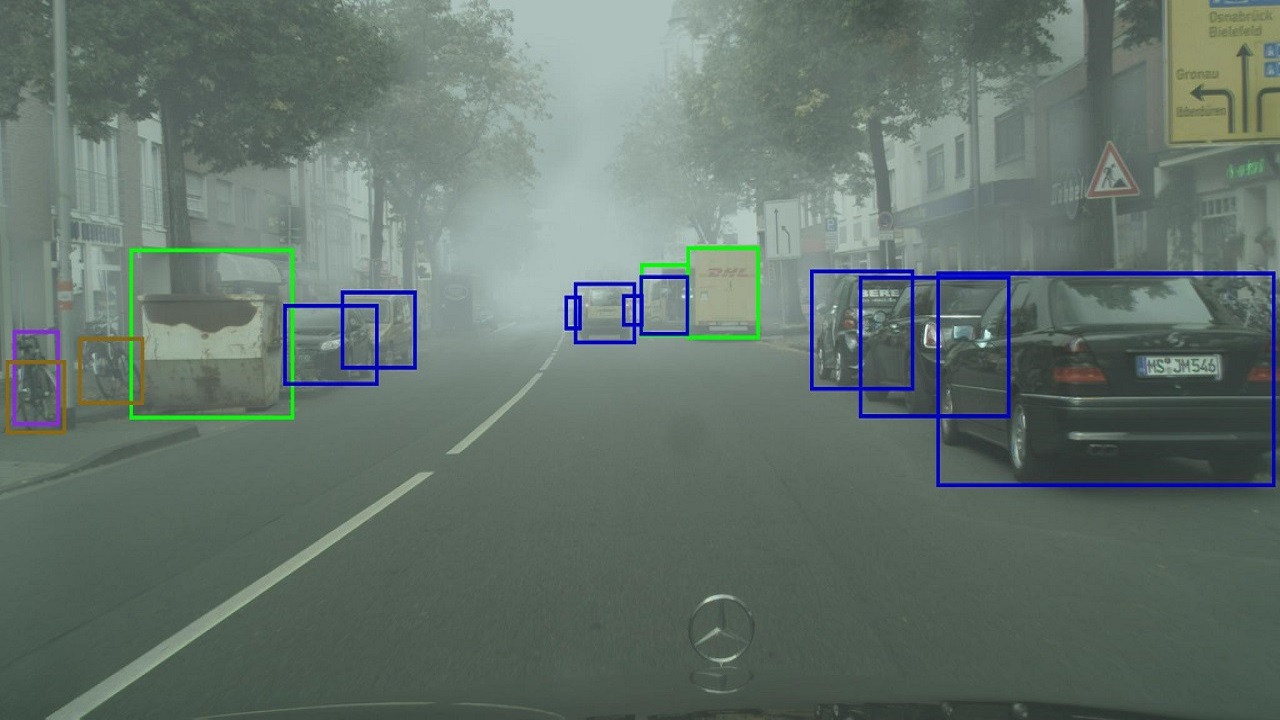}
    \end{minipage}
    \begin{minipage}[b]{.24\linewidth}
        \centering
        \includegraphics[scale=0.13]{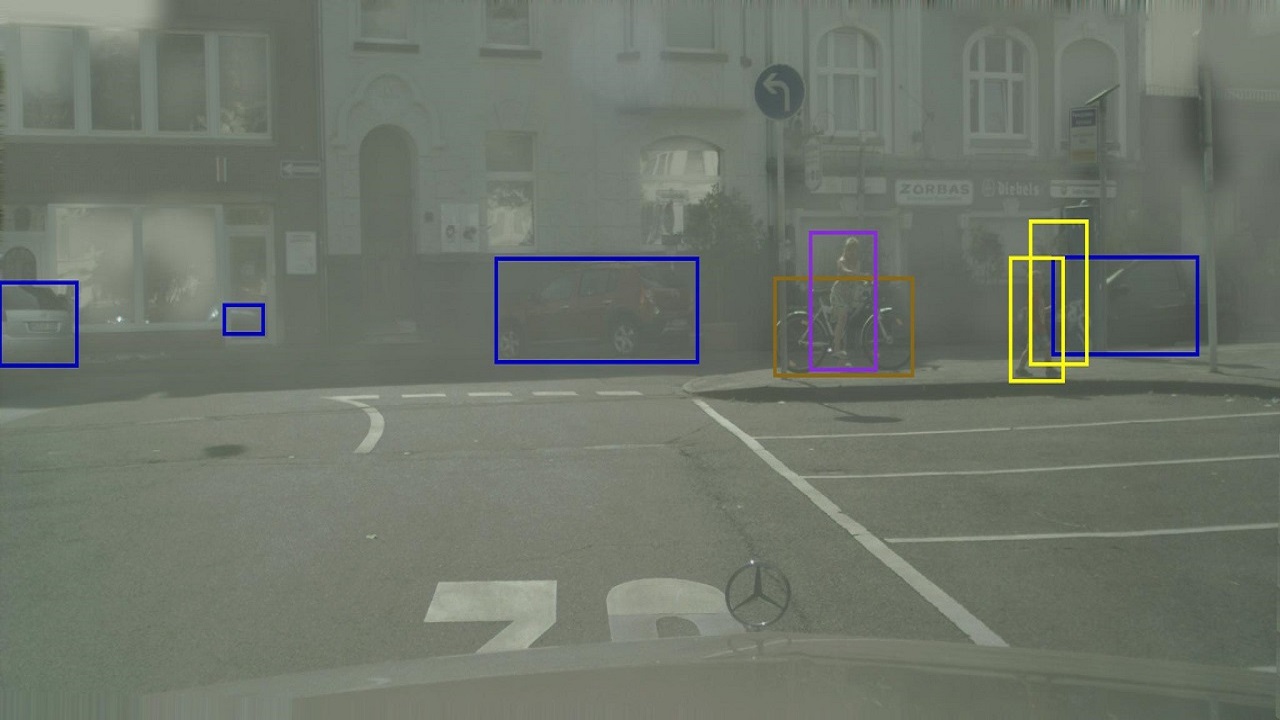}
    \end{minipage}
}

\subfloat
{
    \begin{minipage}[b]{.24\linewidth}
        \centering
        \includegraphics[scale=0.13]{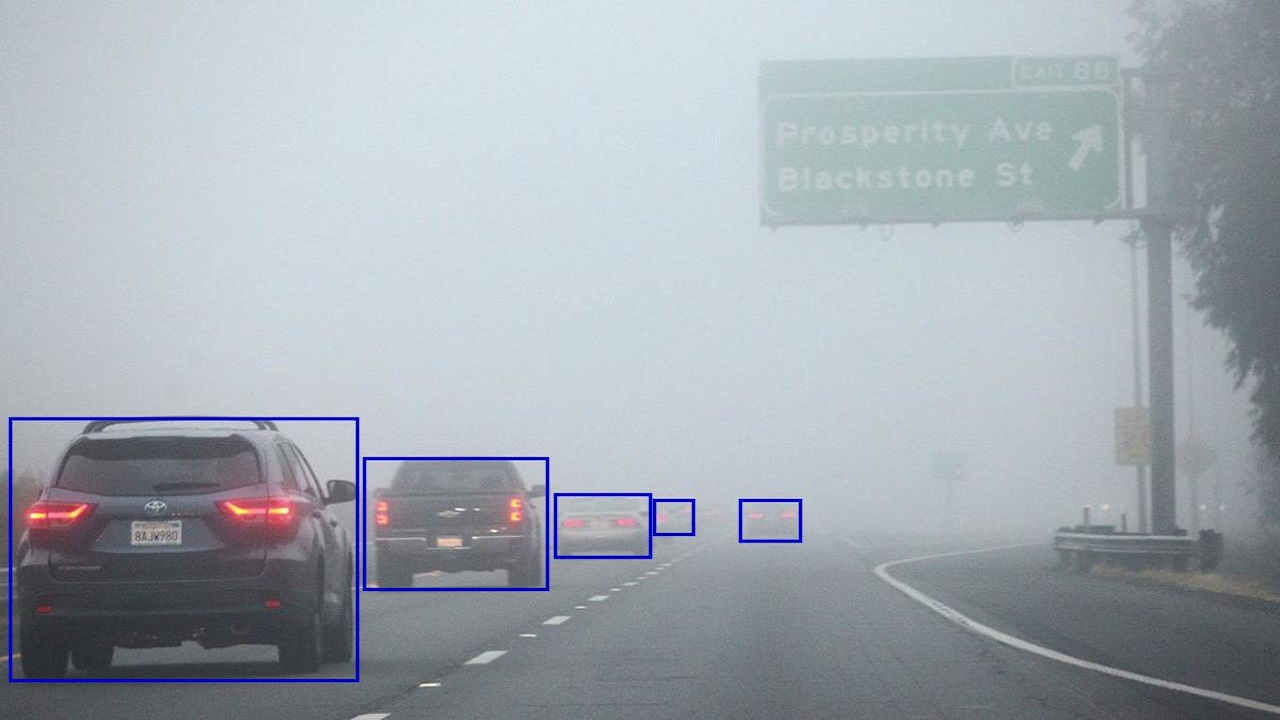}
    \end{minipage}
        \begin{minipage}[b]{.24\linewidth}
        \centering
        \includegraphics[scale=0.13]{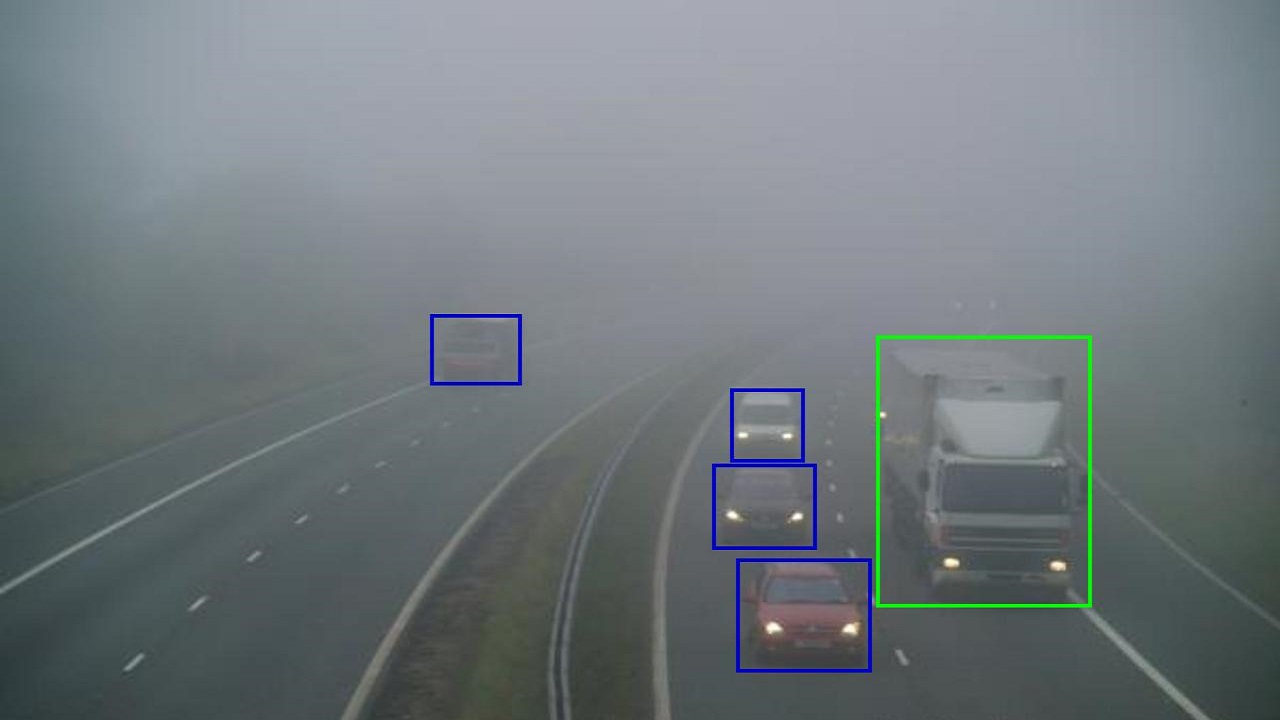}
    \end{minipage}
    \begin{minipage}[b]{.24\linewidth}
        \centering
        \includegraphics[scale=0.13]{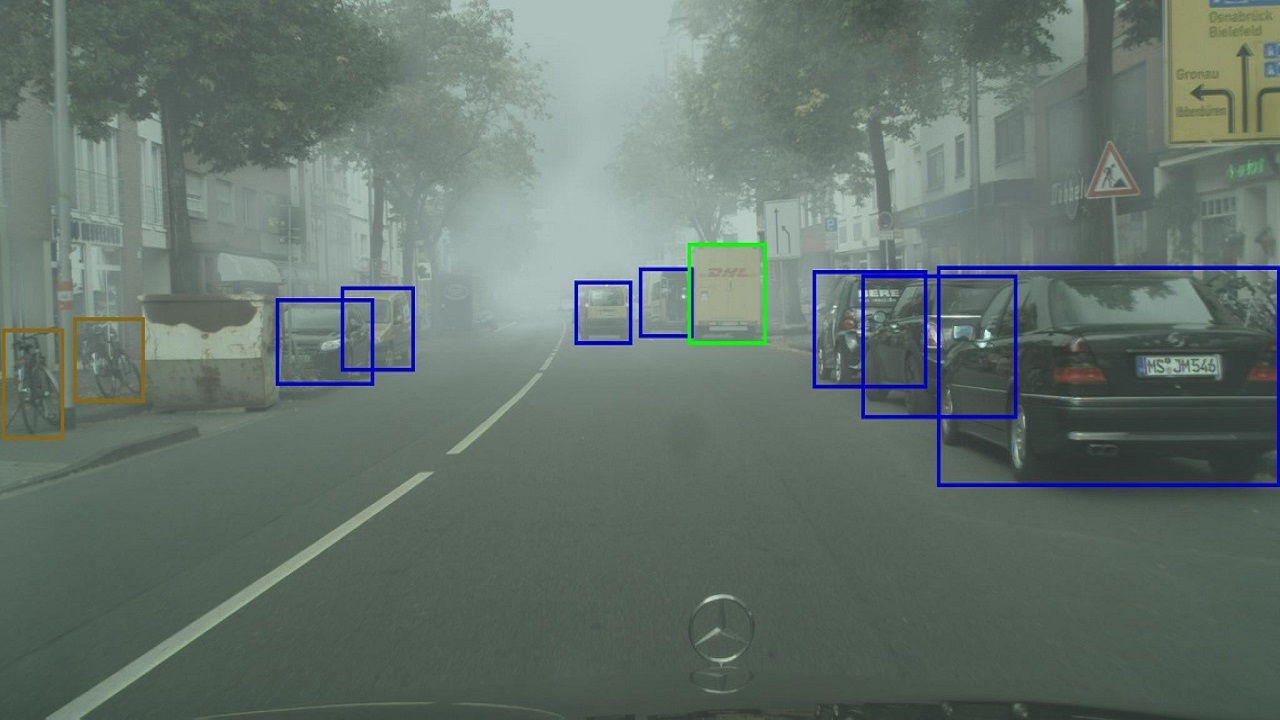}
    \end{minipage}
    \begin{minipage}[b]{.24\linewidth}
        \centering
        \includegraphics[scale=0.13]{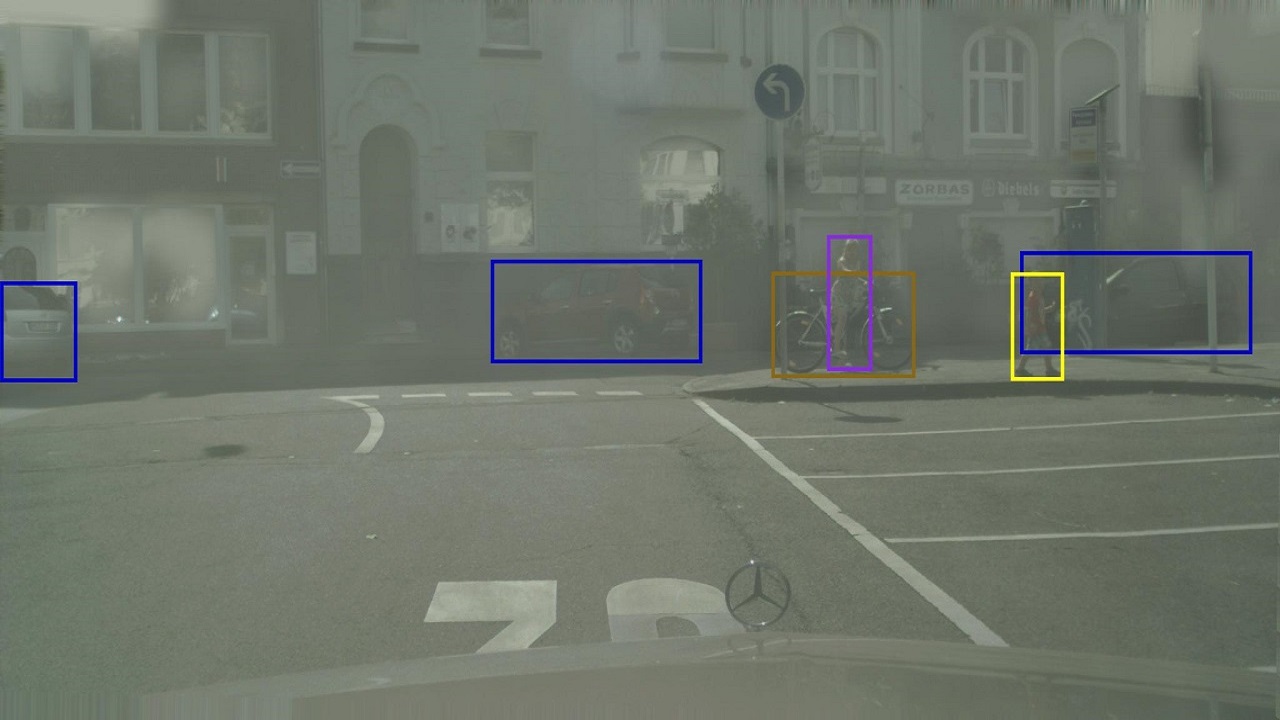}
    \end{minipage}
}
\caption{Qualitative evaluation results of the model's generalization ability on the Daytime-Foggy scene. The top-row and bottom-row images demonstrate the results of vanilla Faster R-CNN \cite{Ren2015Faster} and our method, respectively. Besides, the first two columns are the results of real foggy images and the last two columns are the results of synthetic foggy images.}
\label{fogfig}
\vspace{-0.6em}
\end{figure*}
\noindent \textbf{Results on Dusk-Rainy and Night-Rainy Scenes.} Table \ref{duskrainy} and Table \ref{nightrainy} demonstrate the model's ability to generalize to Dusk-Rainy and Night-Rainy scenes. Our method achieves the best mAP in both scenes, with gains of 0.9\% mAP and 0.5\% mAP respectively compared with CLIP-Gap \cite{vidit2023clip}. Besides, compared with vanilla FR \cite{Ren2015Faster}, our method achieves improvements of 5.2\% mAP and 5.0\% mAP respectively. Furthermore, the feature normalization-based methods \cite{Pan_2019_ICCV,Pan_2018_ECCV,Huang_2019_CVPR,Choi_2021_CVPR} have a poor performance in both scenes, especially in motor category in Night-Rainy scene, with mAP of less than 1.0\%, which reinforces the challenge of these two scenes and the superiority of our method.

\begin{table}[t]
\Large
\resizebox{\linewidth}{!}{
    \begin{tabular}{c|ccccccc|g}
        \toprule[1.5pt]
Methods &Bus &Bike & Car	&Motor	&\makebox[0.01\textwidth][c]{Person}	&Rider	&\makebox[0.01\textwidth][c]{Truck}	&mAP \\
\hline
FR \cite{Ren2015Faster} &34.2	&21.8	&47.9	&16.0	&22.9	&18.5	&34.9	&28.0\\
SW \cite{Pan_2019_ICCV}	&35.2	&16.7	&50.1	&10.4	&20.1	&13.0	&38.8	&26.3\\
IBN-Net \cite{Pan_2018_ECCV}	&37.0	&14.8	&50.3	&11.4	&17.3	&13.3	&38.4	&26.1\\
IterNorm \cite{Huang_2019_CVPR}	&32.9	&14.1	&38.9	&11.0	&15.5	&11.6	&35.7	&22.8\\
ISW	\cite{Choi_2021_CVPR} &34.7	&16.0	&50.0	&11.1	&17.8	&12.6	&38.8	&25.9\\
SDGOD \cite{Wu2022SingleDomaina}	&37.1	&19.6	&50.9	&13.4	&19.7	&16.3	&40.7	&28.2\\
CLIP-Gap \cite{vidit2023clip}	&\textbf{37.8}	&\textbf{22.8}	&60.7	&\textbf{16.8}	&26.8	&\textbf{18.7}	&42.4	&32.3\\
\hline\hline
Ours	&37.1	&21.8	&\textbf{67.9}	&16.4	&\textbf{27.4}	&17.9	&\textbf{43.9}	&\textbf{33.2}\\
\bottomrule[1.5pt]
    \end{tabular}
    }
\caption{Quantitative results (\%) on the Dusk-Rainy scene.}
\centering
\label{duskrainy}
\vspace{-0.5em}
\end{table}

\begin{table}[t]
\Large
\resizebox{\linewidth}{!}{
    \begin{tabular}{c|ccccccc|g}
        \toprule[1.5pt]
Methods &Bus &Bike & Car	&Motor	&\makebox[0.01\textwidth][c]{Person}	&Rider	&\makebox[0.01\textwidth][c]{Truck}	&mAP \\
\hline
FR \cite{Ren2015Faster}	&21.3	&7.7	&28.8	&6.1	&8.9	&10.3	&16.0	&14.2 \\
SW \cite{Pan_2019_ICCV}	&22.3	&7.8	&27.6	&0.2	&10.3	&10.0	&17.7	&13.7\\
IBN-Net \cite{Pan_2018_ECCV}	&24.6	&10.0	&28.4	&0.9	&8.3	&9.8	&18.1	&14.3\\
IterNorm \cite{Huang_2019_CVPR}	&21.4	&6.7	&22.0	&0.9	&9.1	&10.6	&17.6	&12.6\\
ISW	\cite{Choi_2021_CVPR} &22.5	&11.4	&26.9	&0.4	&9.9	&9.8	&17.5	&14.1\\
SDGOD \cite{Wu2022SingleDomaina}	&24.4	&11.6	&29.5	&\textbf{9.8}	&10.5	&\textbf{11.4}	&19.2	&16.6\\
CLIP-Gap \cite{vidit2023clip}	&28.6	&\textbf{12.1}	&\textbf{36.1}	&9.2	&12.3	&9.6	&22.9	&18.7\\
\hline\hline
Ours	&\textbf{29.9}	&11.8	&\textbf{36.1}	&9.4	&\textbf{13.1}	&10.5	&\textbf{23.3}	&\textbf{19.2}\\
\bottomrule[1.5pt]
    \end{tabular}
    }
\caption{Quantitative results (\%) on the Night-Rainy scene.}
\centering
\label{nightrainy}
\vspace{-0.8em}
\end{table}
\noindent \textbf{Results on Daytime-Foggy Scene.} Table \ref{dayfoggy} shows the generalization results of our model in Daytime-Foggy scene. We can see that our method outperforms all the methods in the table and achieves 39.6\% mAP. Specifically, compared with SDGOD \cite{Wu2022SingleDomaina} and CLIP-Gap \cite{vidit2023clip}, our method achieves improvements of 6.1\% mAP and 1.1\% mAP, respectively. 
Besides, some visualization results are demonstrated in Fig. \ref{fogfig}. Compared with vanilla FR \cite{Ren2015Faster}, our method achieves more accurate object localization and classification in both real and synthetic foggy environments.

\begin{figure*}[t]
\centering
\subfloat
{
    \begin{minipage}[b]{.24\linewidth}
        \centering
        \includegraphics[scale=0.13]{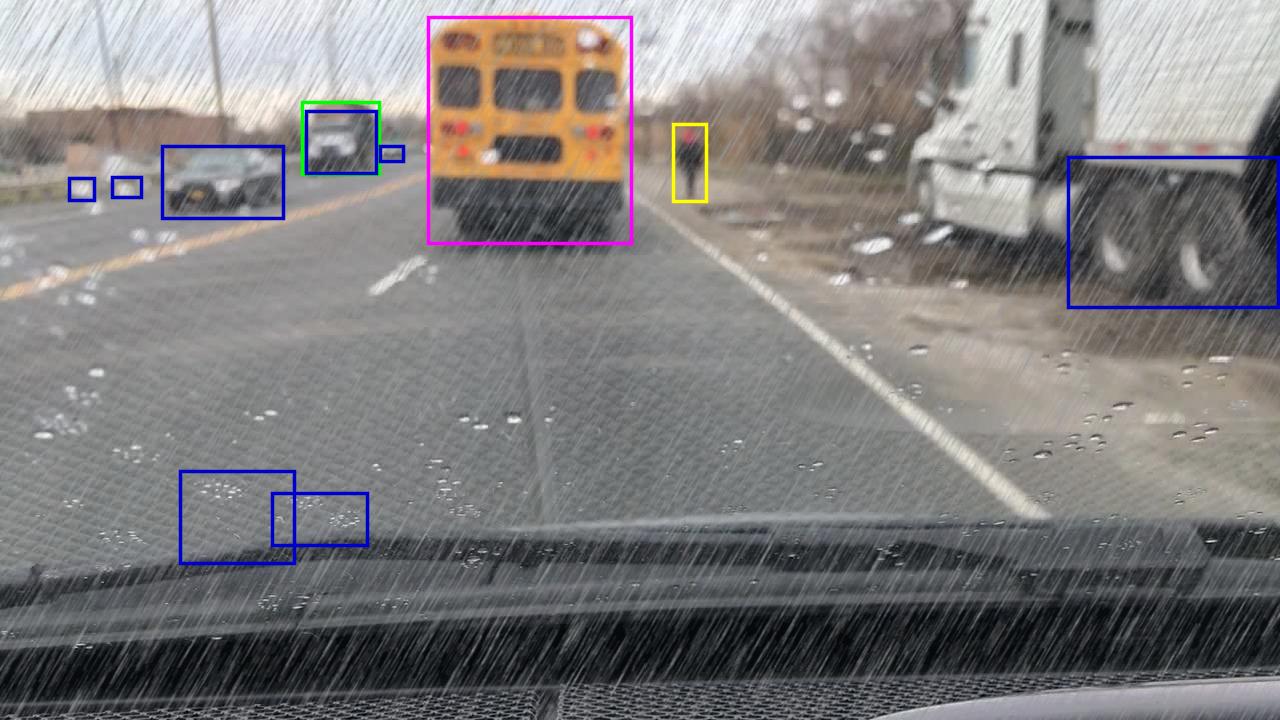}
    \end{minipage}
    \begin{minipage}[b]{.24\linewidth}
        \centering
        \includegraphics[scale=0.3355]{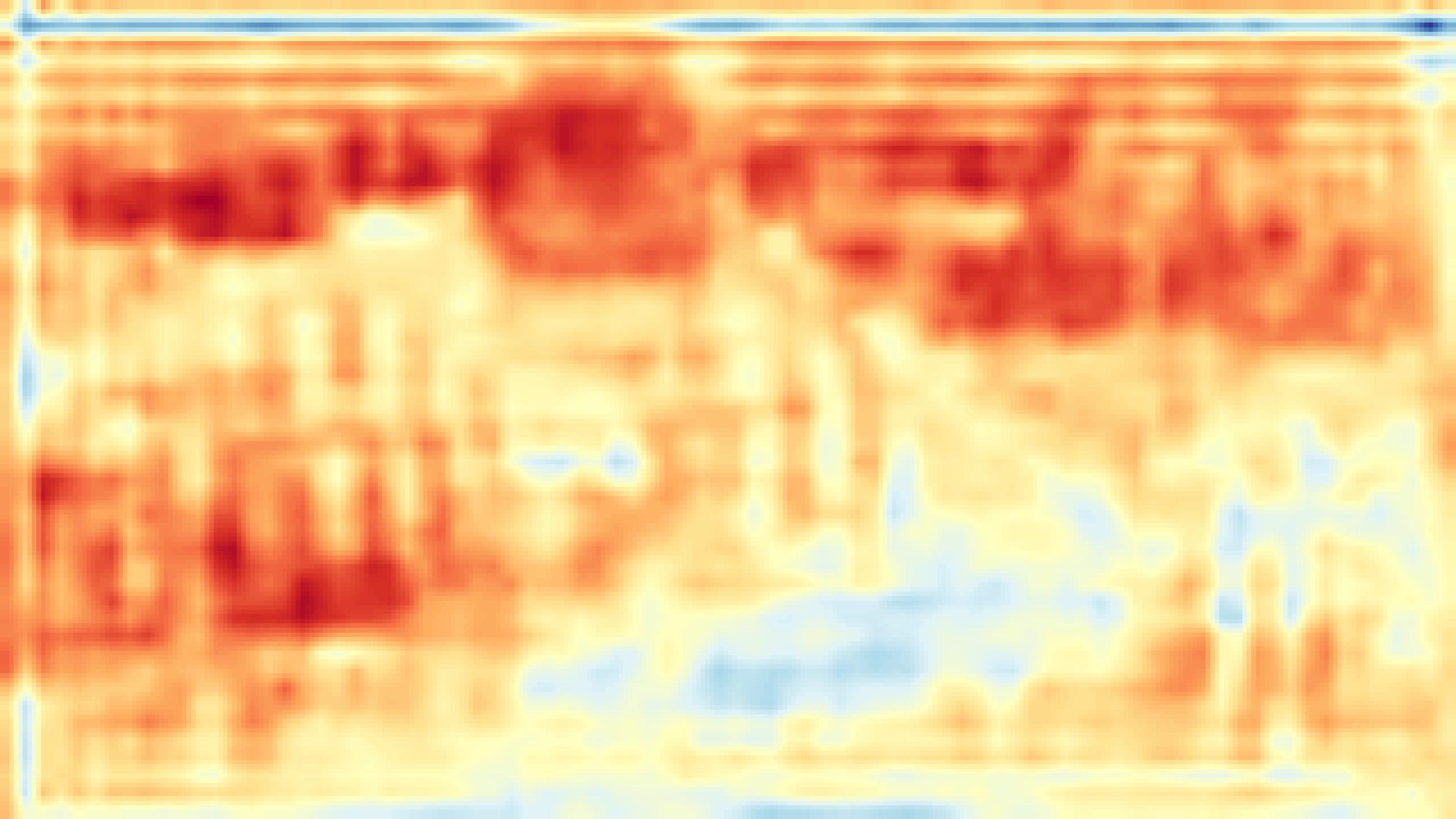}
    \end{minipage}
    \begin{minipage}[b]{.24\linewidth}
        \centering
        \includegraphics[scale=0.26]{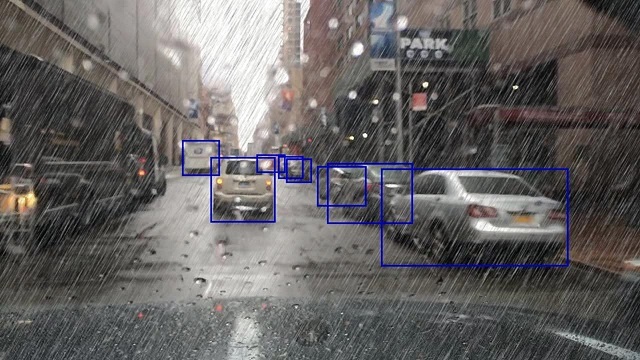}
    \end{minipage}
    \begin{minipage}[b]{.24\linewidth}
        \centering
        \includegraphics[scale=0.335]{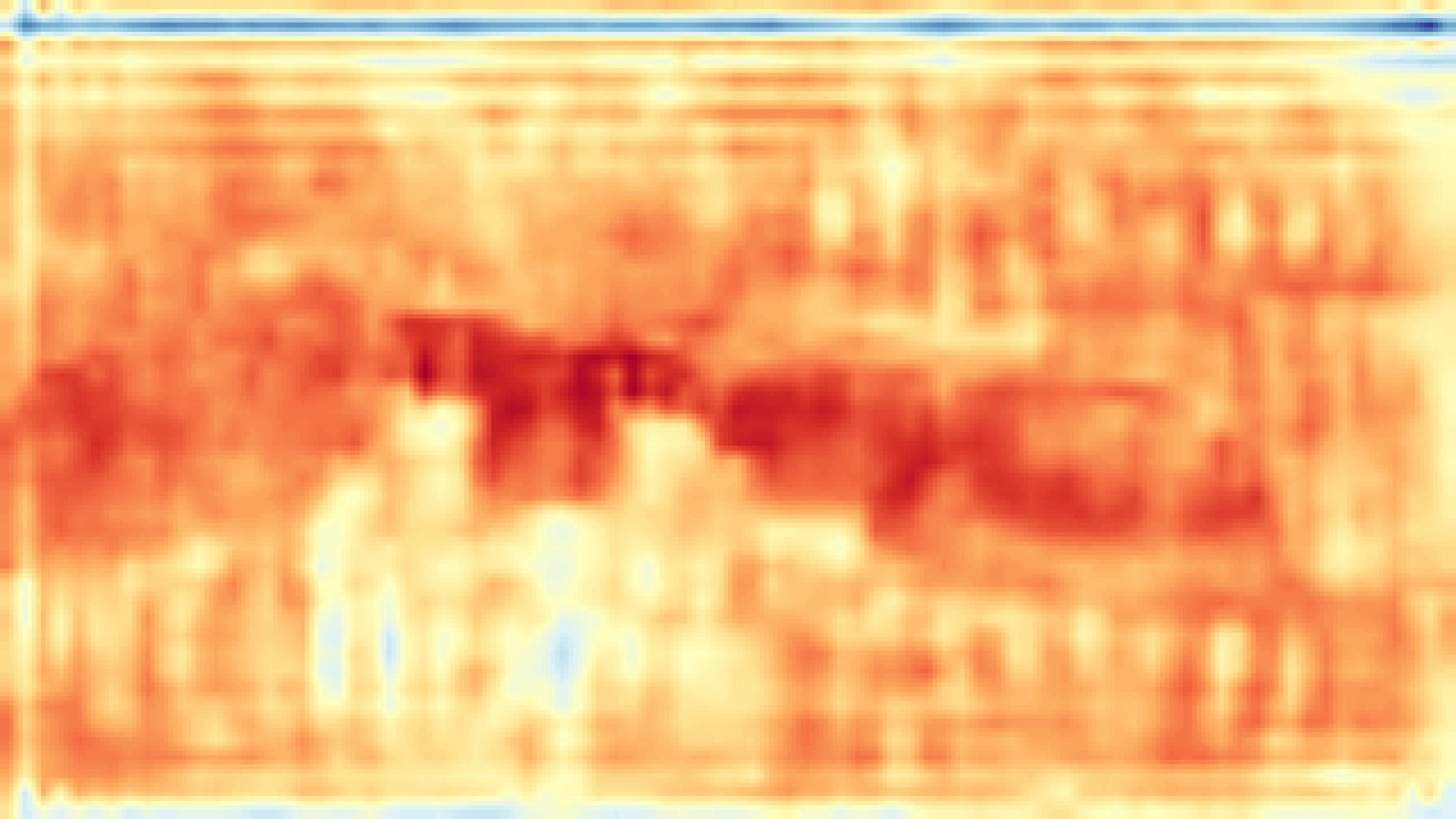}
    \end{minipage}
}

\subfloat
{
    \begin{minipage}[b]{.24\linewidth}
        \centering
        \includegraphics[scale=0.13]{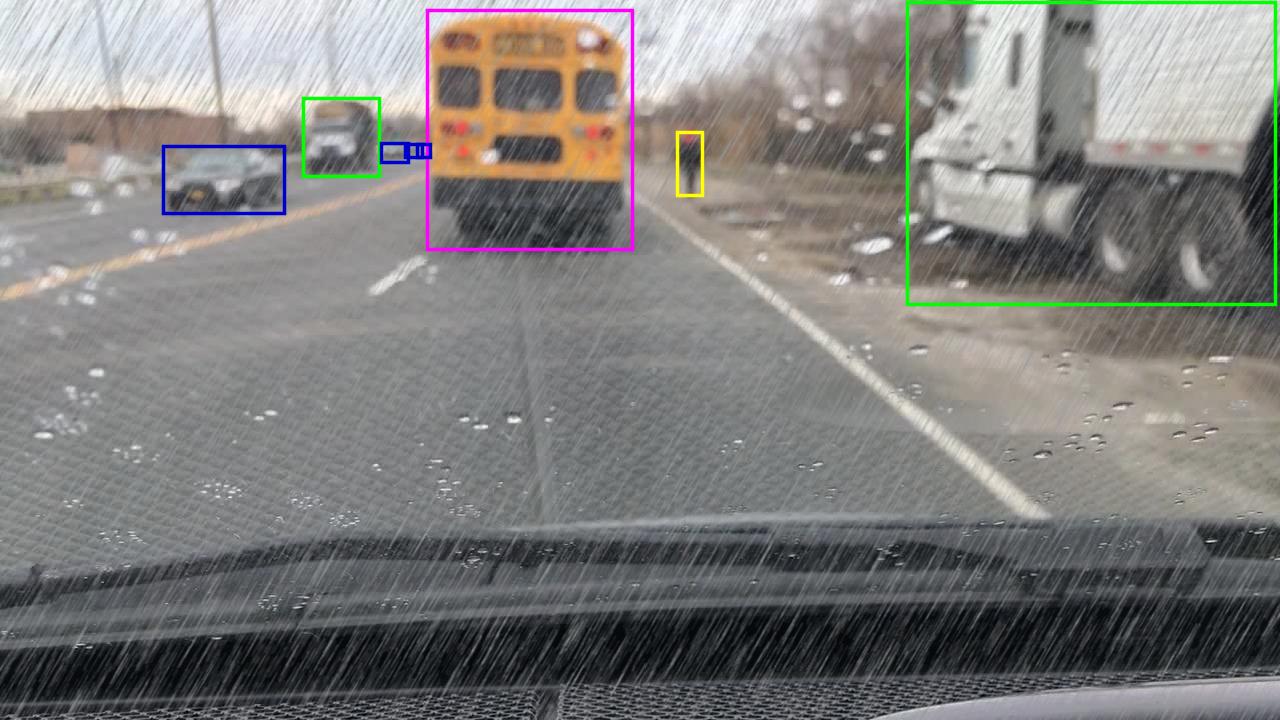}
    \end{minipage}
        \begin{minipage}[b]{.24\linewidth}
        \centering
        \includegraphics[scale=0.33555]{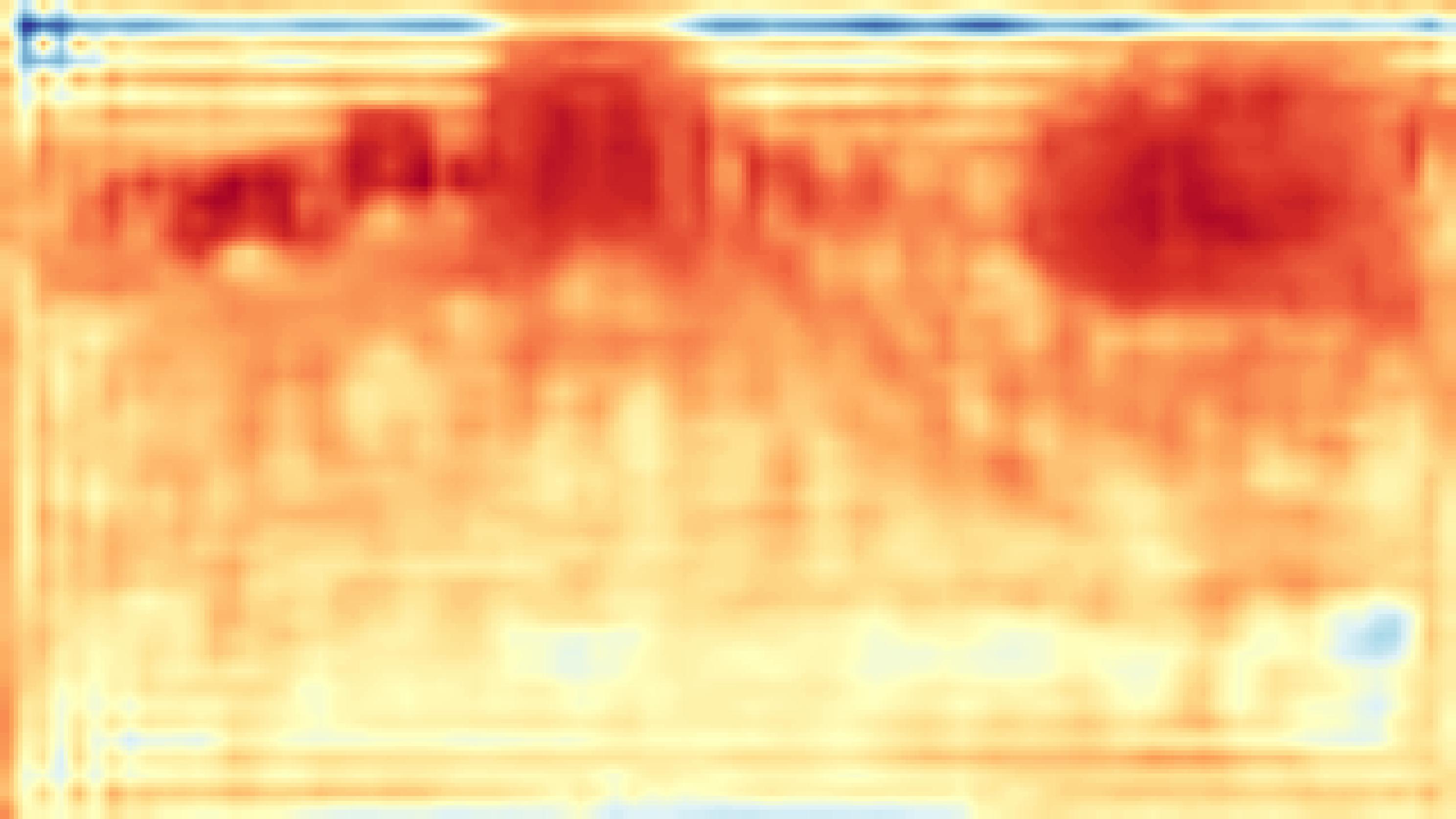}
    \end{minipage}
    \begin{minipage}[b]{.24\linewidth}
        \centering
        \includegraphics[scale=0.26]{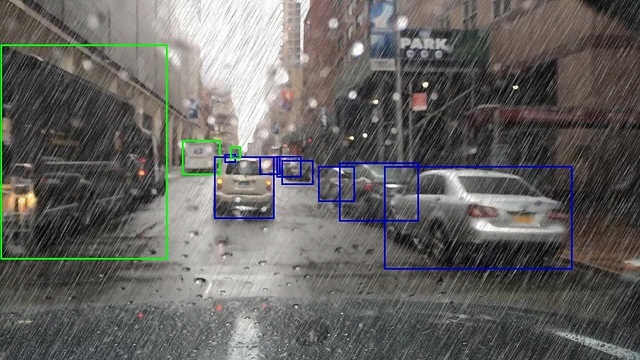}
    \end{minipage}
    \begin{minipage}[b]{.24\linewidth}
        \centering
        \includegraphics[scale=0.3355]{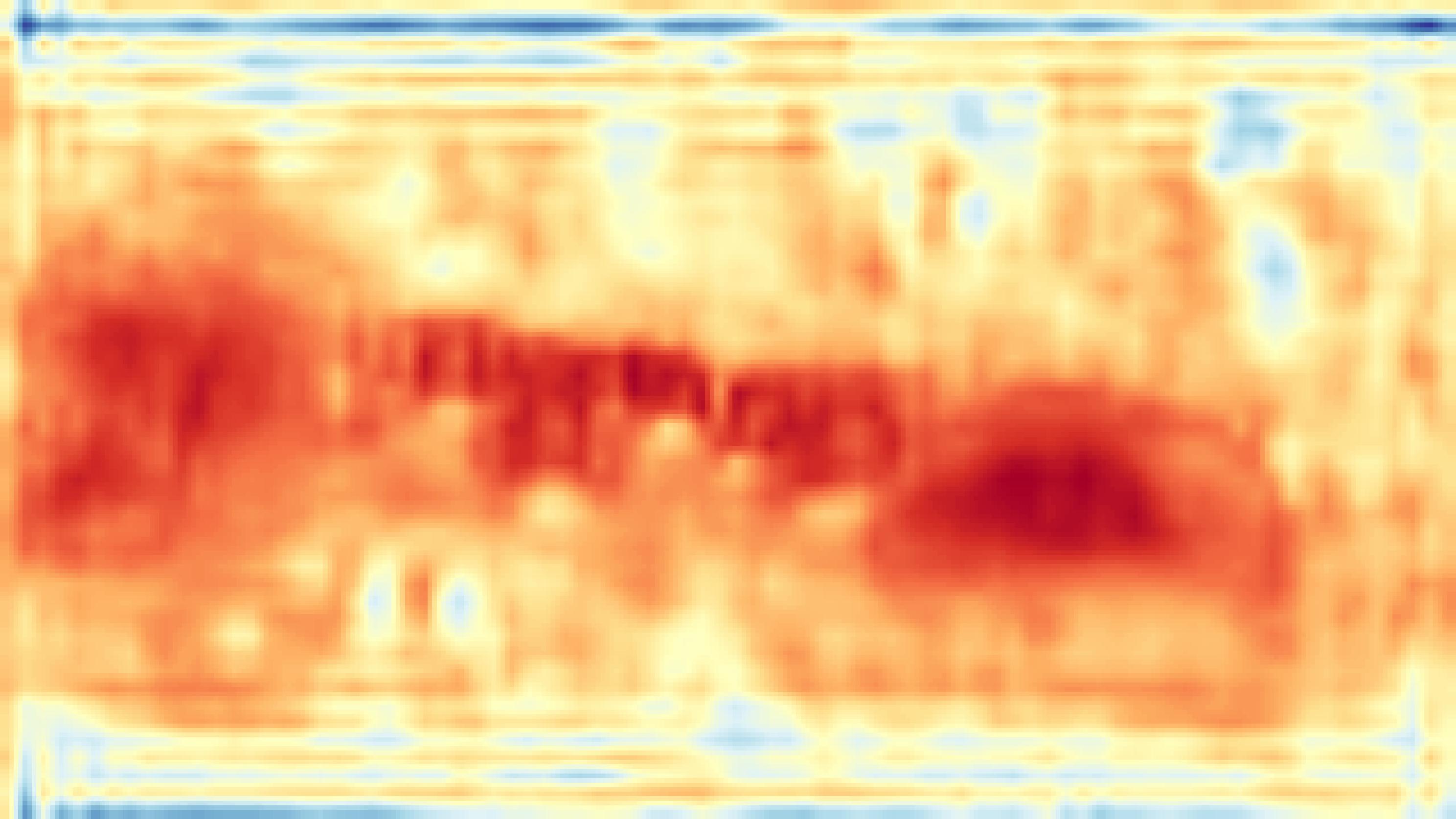}
    \end{minipage}
}
\caption{Visualization of the detection results and attention maps on the Dusk-Rainy scene. The top-row and bottom-row images are the results of vanilla Faster R-CNN  \cite{Ren2015Faster} and our method respectively. The dark red region represents the area where the attention is salient.}
\label{heatmap}
\vspace{-0.6em}
\end{figure*}

\subsection{Ablation Study}
\label{secablation}
In this section, we conduct several experiments on Daytime-Clear, Night-Clear  and Daytime-Foggy scenes to analyze the role of each component of the Unbiased Faster R-CNN model. Specifically, the experimental results in Table \ref{ablation} are obtained by training on the Daytime-Clear scene and testing on the three weather conditions.\\
\noindent \textbf{Analysis of the GLT module.} As shown in Table \ref{ablation}, the results of vanilla FR \cite{Ren2015Faster} are improved when combined with the GLT module, with gains of 4.2\% mAP in Daytime-Clear scene, 2.8\% mAP in Night-Clear scene, and 2.6\% mAP in Daytime-Foggy scene.  Besides, the 60.4\% mAP in Daytime-Clear scene is optimal, which indicates that the GLT  module is an effective data augmentation method that increases data diversity and performs exceptionally well in the source domain with supervised learning. 

\begin{table}[t]
\Large
\resizebox{\linewidth}{!}{
    \begin{tabular}{c|ccccccc|g}
        \toprule[1.5pt]
Methods &Bus &Bike & Car	&Motor	&\makebox[0.01\textwidth][c]{Person}	&Rider	&\makebox[0.01\textwidth][c]{Truck}	&mAP \\
\hline
FR \cite{Ren2015Faster}	&34.5	&29.6	&49.3	&26.2	&33.0	&35.1	&26.7	&33.5\\
SW \cite{Pan_2019_ICCV}	&30.6	&36.2	&44.6	&25.1	&30.7	&34.6	&23.6	&30.8\\
IBN-Net \cite{Pan_2018_ECCV}	&29.9	&26.1	&44.5	&24.4	&26.2	&33.5	&22.4	&29.6\\
IterNorm \cite{Huang_2019_CVPR}	&29.7	&21.8	&42.4	&24.4	&26.0	&33.3	&21.6	&28.4\\
ISW	\cite{Choi_2021_CVPR} &29.5	&26.4	&49.2	&27.9	&30.7	&34.8	&24.0	&31.8\\
SDGOD \cite{Wu2022SingleDomaina}	&32.9	&28.0	&48.8	&29.8	&32.5	&38.2	&24.1	&33.5\\
CLIP-Gap \cite{vidit2023clip}	&36.2	&34.2	&57.9	&\textbf{34.0}	&38.7	&\textbf{43.8}	&25.1	&38.5\\
\hline\hline
Ours	&\textbf{36.9}	&\textbf{35.8}	&\textbf{61.7}	&33.7	&\textbf{39.5}	&42.2	&\textbf{27.5}	&\textbf{39.6}\\
\bottomrule[1.5pt]
    \end{tabular}
    }
\caption{Quantitative results (\%) on the Daytime-Foggy scene.}
\centering
\label{dayfoggy}
\vspace{-0.8em}
\end{table}
\noindent \textbf{Analysis of $\mathcal{L}_{att}$, $\mathcal{L}_{exp}$ and $\mathcal{L}_{imp}$.} As depicted in Table \ref{ablation}, incorporating $\mathcal{L}_{att}$, $\mathcal{L}_{exp}$ and $\mathcal{L}_{imp}$ in addition to the  GLT module decreases the model performance in the Daytime-Clear scene while improving the performance in the Night-Clear and Daytime-Foggy scenes, which suggests that these constraints limit the model's performance in the seen environment with supervised learning, but improve the model's performance in unseen environments. The phenomenon is due to that these constraints encourage the model to extract causal knowledge, thereby preventing the model from acquiring domain-specific knowledge. The domain-specific knowledge, although not generalizable, carries some supplementary information for supervised learning. 
Furthermore, we can observe that the attention invariance loss $\mathcal{L}_{att}$ improves the model generalization performance more significantly, which indicates that the scene confounders have a greater impact on the model generalization ability. Besides, compared to the implicit constraint $\mathcal{L}_{imp}$, the explicit constraint $\mathcal{L}_{exp}$ is more binding and contributes more to generalization. Therefore, purely implicit constraint cannot guarantee the model to learn causal prototypes, and explicit constraint on the predicted distribution is necessary.
\begin{table}
    \centering
    \Large
\resizebox{\linewidth}{!}{
    \begin{tabular}{c|ccc|ccc}
    \toprule[1.5pt]
        Methods &$\mathcal{L}_{att}$ &$\mathcal{L}_{exp}$ &$\mathcal{L}_{imp}$ & D-Clear &N-Clear  &D-Foggy \\
        \hline
          FR \cite{Ren2015Faster}& &  & & 56.2	&35.8	&33.5\\
          \hline
          \multirow{5}{*}{+GLT}&   & & &  \textbf{60.4}	&38.6	&36.1\\
          &\Checkmark &   &  &59.5	&39.7	&38.1\\
          &\Checkmark &\Checkmark &  & 59.1	&40.4	&39.0\\
          &\Checkmark & &\Checkmark  & 59.3	&40.0	&38.5\\
          &\Checkmark &\Checkmark &\Checkmark   &58.6	&\textbf{40.8}	&\textbf{39.6} \\
    \bottomrule[1.5pt]
    \end{tabular}
    }
    \caption{Ablation study results (\%) of our proposed UFR method. `+GLT' denotes incorporating the GLT to vanilla FR\cite{Ren2015Faster}. The loss check mark denotes the addition of the loss to FR \cite{Ren2015Faster} with GLT.}
    \label{ablation}
    \vspace{-0.7em}
\end{table}
\subsection{Further Analysis}
\begin{figure}[t]
	\centering
	\begin{subfigure}{0.49\linewidth}
		\centering
		\includegraphics[width=1.0\linewidth]{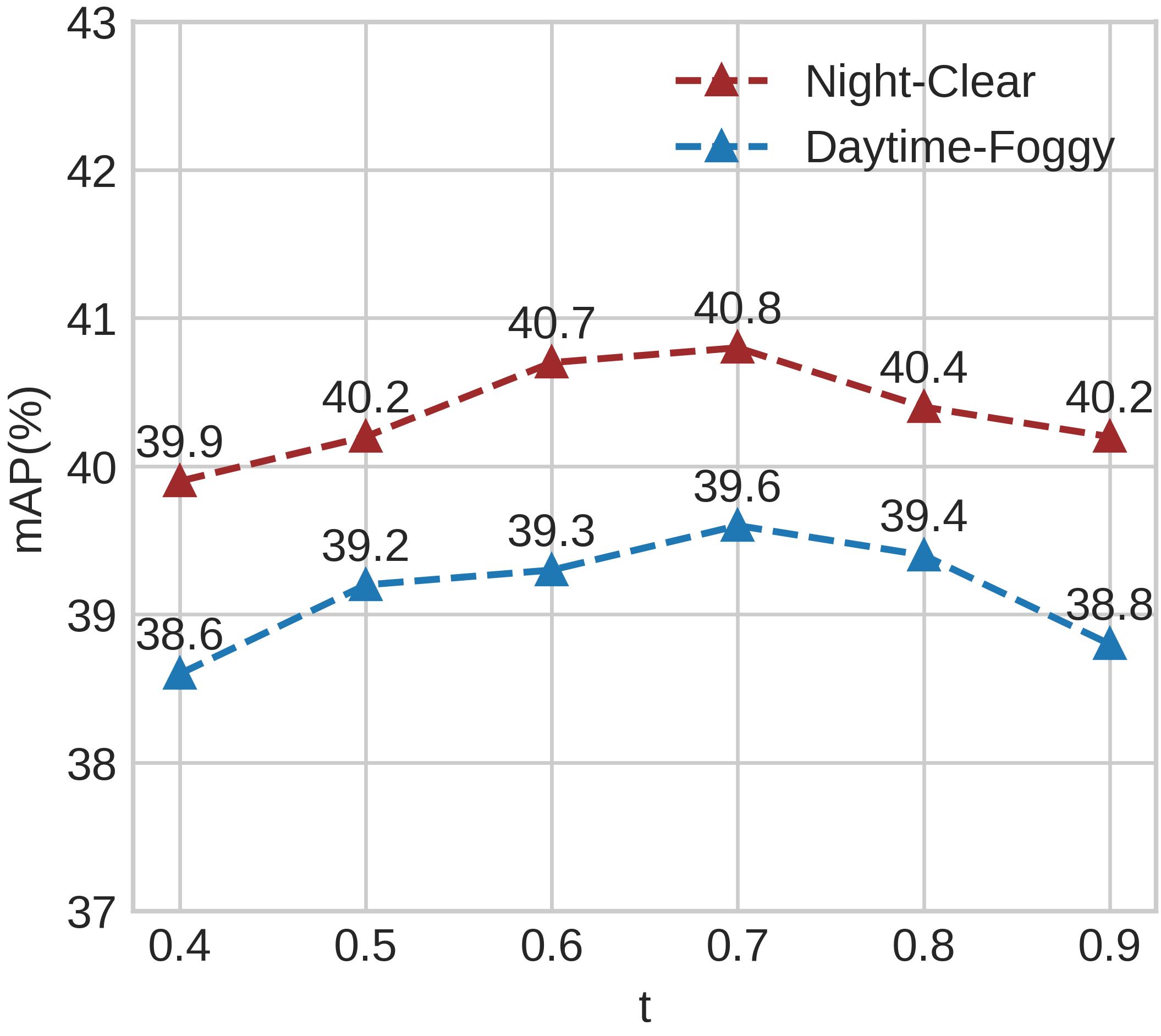}
        \caption{\label{t}Analysis on $t$}
	\end{subfigure}
	\centering
	\begin{subfigure}{0.49\linewidth}
		\centering
		\includegraphics[width=1.0\linewidth]{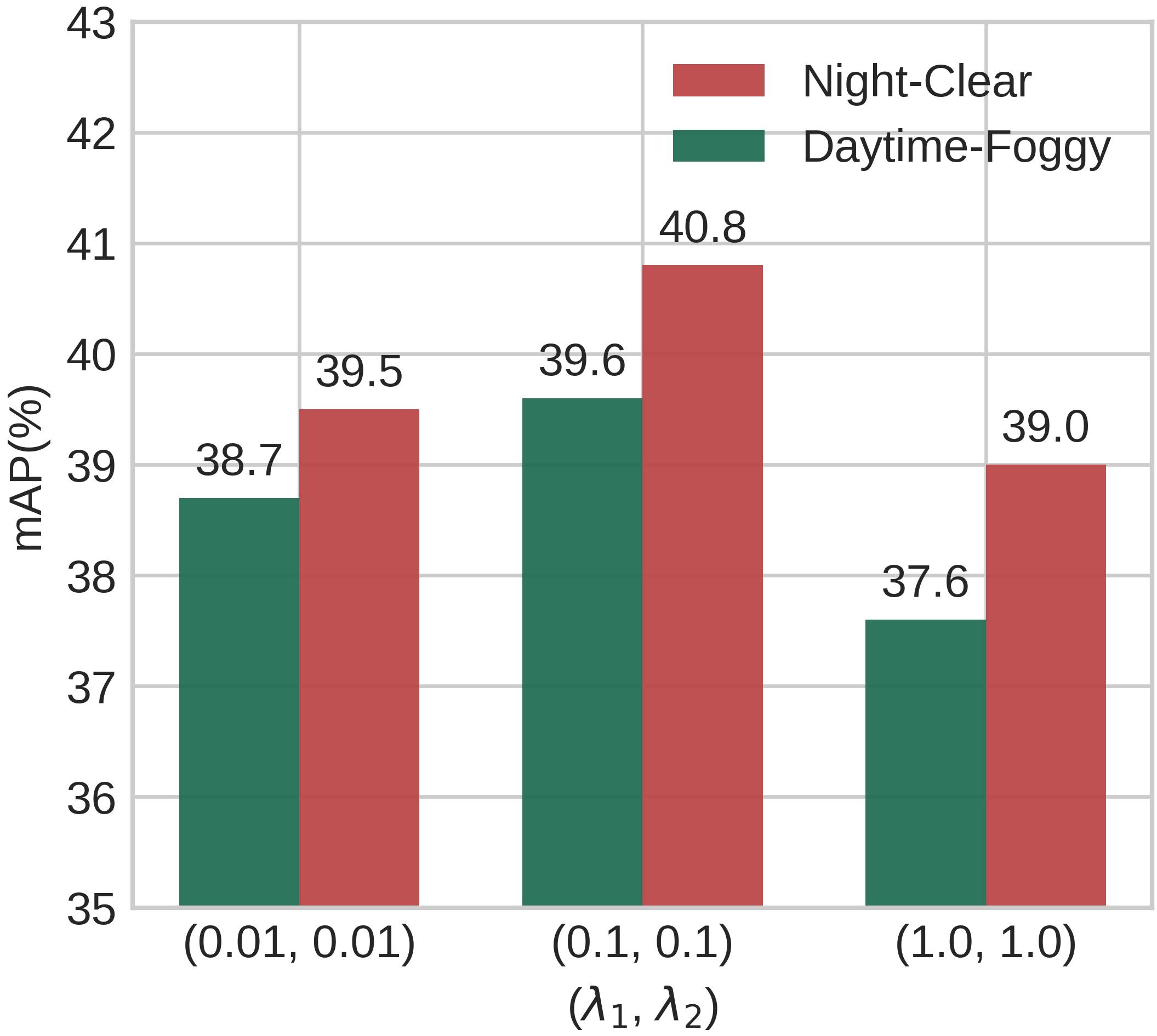}
		\caption{\label{lambda}Analysis on $\lambda_1$ and $\lambda_2$}
	\end{subfigure}
	\caption{Results of hyperparameter analysis.}
	\label{param}
 \vspace{-0.7em}
\end{figure}

\noindent\textbf{Attention visualization.} We conduct experiments on Dusk-Rainy scene and compare the attention maps produced by  vanilla FR \cite{Ren2015Faster} and our UFR model, as shown in Fig. \ref{heatmap}. We can observe that the attention maps generated by FR \cite{Ren2015Faster} are unfocused and have more attention on irrelevant background areas. In contrast, our method produces more effective category-related attention and has less attention on background, which suggests that our method can capture discriminative object features in unseen environments with a superior generalization performance.\\
\noindent \textbf{Hyperparameter analysis.} As demonstrated in Fig. \ref{param}, we train our model with different settings of $t$, $\lambda_1$ and $\lambda_2$ and test the generalization performance on the Night-Clear and Daytime-Foggy scenes. The model performance with different settings of threshold $t$ in Eq. (\ref{explicit}) is reported in Fig. \ref{t}. The best threshold $t$ is 0.7. Besides, we also report the results of our model in Fig. \ref{lambda} with different settings of $\lambda_1$ and $\lambda_2$ in Eq. (\ref{total}). It is obvious that the model achieves the best performance with $\lambda_1=0.1$ and $\lambda_2=0.1$.
\section{Conclusions}
In this paper, we analyze the factors that contribute to the limited single-domain generalization ability of Faster R-CNN from a causal perspective and summarize them as data bias, attention bias and prototype bias. To tackle these challenges, we propose a novel Unbiased Faster R-CNN. Our model leverages a Structural Causal Model to analyze the biases in the  task that arise from both scene confounders and object attribute confounders. We design a Global-Local Transformation module for data augmentation to mitigate the data bias. To learn features that are robust to scene confounders, we introduce a Causal Attention Learning module for image-level  causal feature selection. To further mitigate the influence of object attribute confounders, we develop a Causal Prototype Learning module that learns object-level causal features. Experimental results and further analysis demonstrate the effectiveness of our method.
\section*{Acknowledgements}
This work was supported by National Natural Science Foundation of China under Grants U2013210 and 62303447; Liaoning Science Fund for Distinguished Young Scholars under Grant 2023JH6/100500005; Postdoctoral Innovation Talents Support Program, No. BX20230399.

{
    \small
    \bibliographystyle{ieeenat_fullname}
    \bibliography{main}
}

\end{document}


\maketitle
\thispagestyle{empty}
\subsection{Connection between causality and our method} 
In our method, the non-causal factors are scenes and non-discriminative object attributes, which lead to attention bias and prototype bias in feature space.
Our method is to eliminate the two biases by \textbf{learning generalized invariance features robust to the change of non-causal factors} in unseen domains. However, the training data is biased and can't cover a rich set of these non-causal factors. Thus the GLT is proposed to reduce \textbf{data bias}. The causal attention learning (\textbf{attention debias}) and causal prototype learning (\textbf{prototype debias}) modules are designed to learn invariance features from a rich data distribution with diverse non-causal factors. 
Following the causal theory \cite{buhlmann2020invariance} , the probabilistic invariance constraint $\mathcal{L}_{exp}$ of the predicted results in Eq. (11) is an explicit constraint for causal learning and the feature invariance constraints $\mathcal{L}_{att}$ and $\mathcal{L}_{imp}$in the representation space are implicit ones.\\
\vspace{-1em}
\subsection{Reasons of using Dice loss with binary significance maps instead of using MSE loss with attention maps}
The binary significance map provides a good representation of the activated and inactivated regions. And the MSE loss constrains the difference in attention values more. For this task, \textbf{it is sufficient that the activated regions} are consistency with dice loss though there are differences in the attention values. I think the MSE loss on attention maps is a hard constraint and the dice loss on the significance maps is a soft constraint.
Besides, we conduct experiments to analyze the impact of \textbf{constraining the attention values with  MSE loss} and the mAP on Night-Clear scene decreases by 1.03\%.\\
\begin{table}[t]
\centering
\setlength{\tabcolsep}{1.5mm}{
    \begin{tabular}{ccc}
    \hline
         Methods &Night-Clear  &Day-Foggy \\\hline
         Baseline&11.93  &8.47\\\hline
         +GLT& 7.25 &5.18\\\hline
    \end{tabular}}
        \captionof{table}{$L_1$ distance to $F_3$}
        \label{tab}
\vspace{-0.5em}
\end{table}
\begin{table}[t]
\small
\centering
\setlength{\tabcolsep}{1.0mm}{
    \begin{tabular}{cccccccc}
    \hline
         Methods &Bus &Bike & Car	&Motor	&Person	&Rider	&Truck \\\hline
         UFR w/o $\mathcal{L}_{prot}$& 5.16 &5.90  &5.10 &4.86 & 5.88 &6.34 &4.95\\\hline
         UFR w/ $\mathcal{L}_{prot}$&3.69  &3.17 &2.94  &4.15 & 2.72 &3.11 &4.02\\\hline
    \end{tabular}}
        \captionof{table}{$L_1$ distance between $p_i^c$ and $p_{avg}^c$}
        \label{tab2}
\vspace{-0.5em}
\end{table}

\section{Experiments}
\subsection{More Implementation Details}
\begin{figure*}[htbp]
\centering
\subfloat[original Objects]
{
    \begin{minipage}[b]{.24\linewidth}
        \centering
        \includegraphics[scale=0.0935]{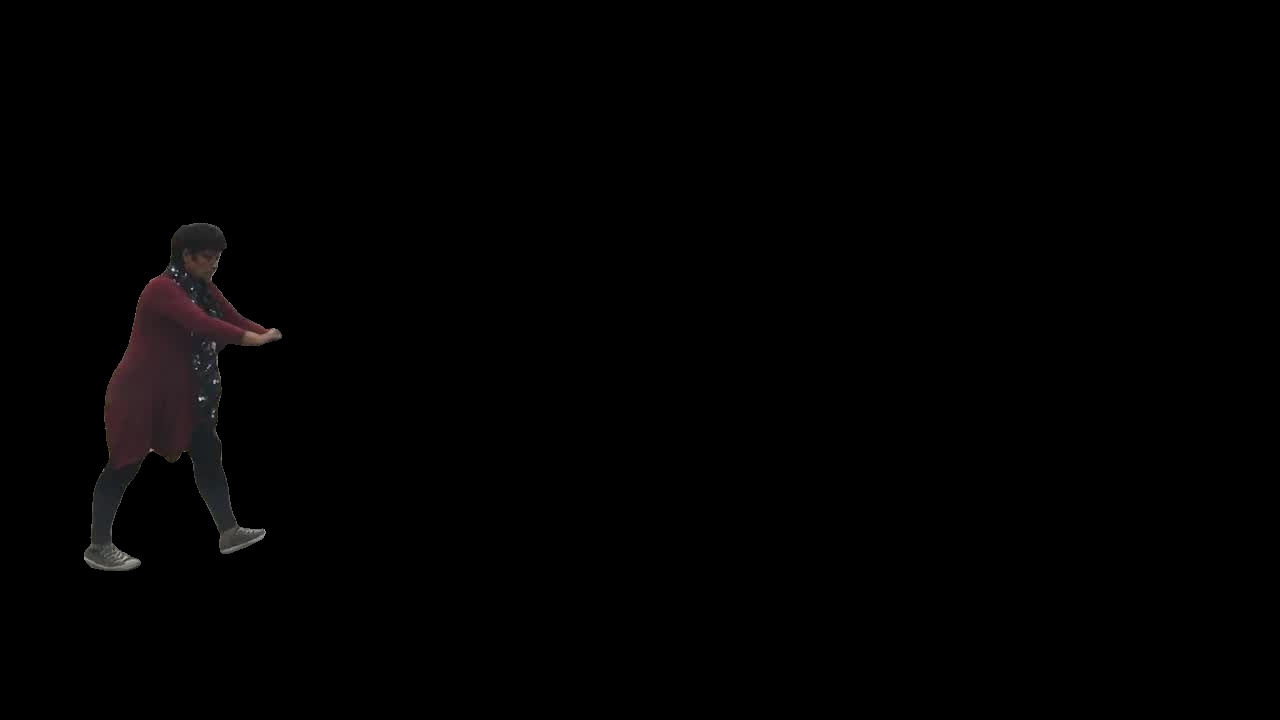}
    \end{minipage}
        \begin{minipage}[b]{.24\linewidth}
        \centering
        \includegraphics[scale=0.0935]{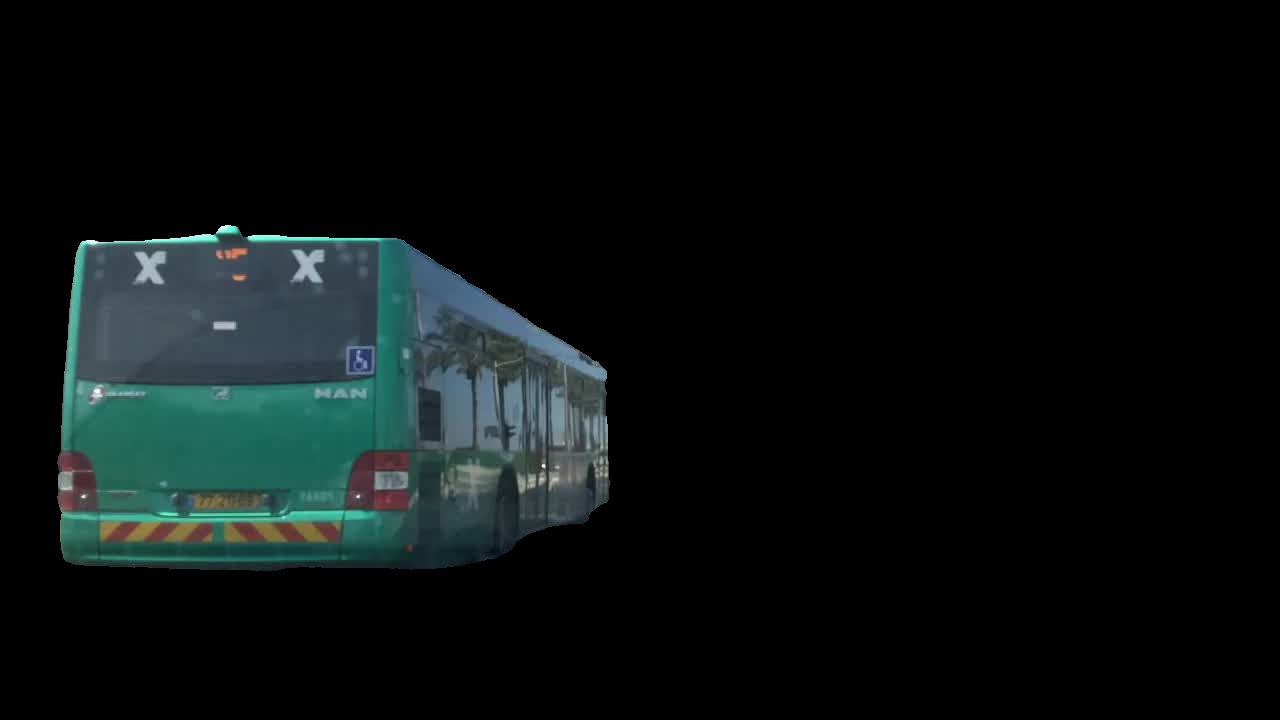}
    \end{minipage}
        \begin{minipage}[b]{.24\linewidth}
        \centering
        \includegraphics[scale=0.0935]{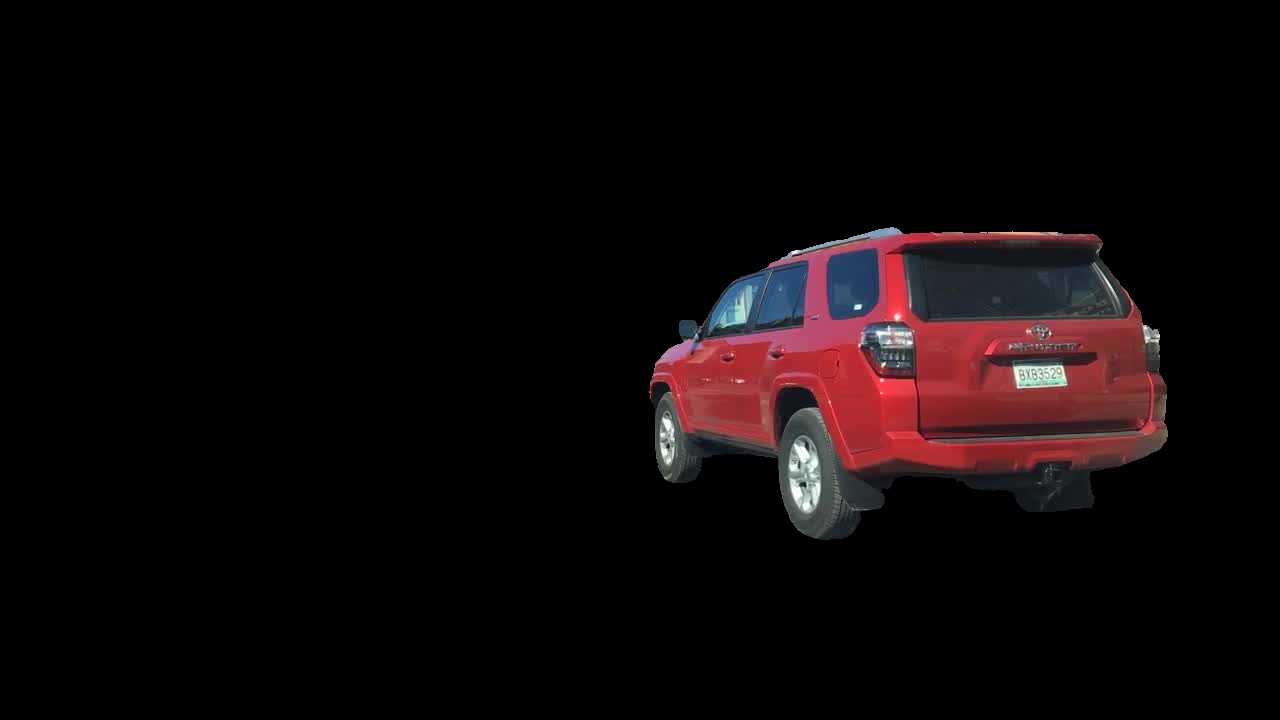}
    \end{minipage}
    \begin{minipage}[b]{.24\linewidth}
        \centering
        \includegraphics[scale=0.0935]{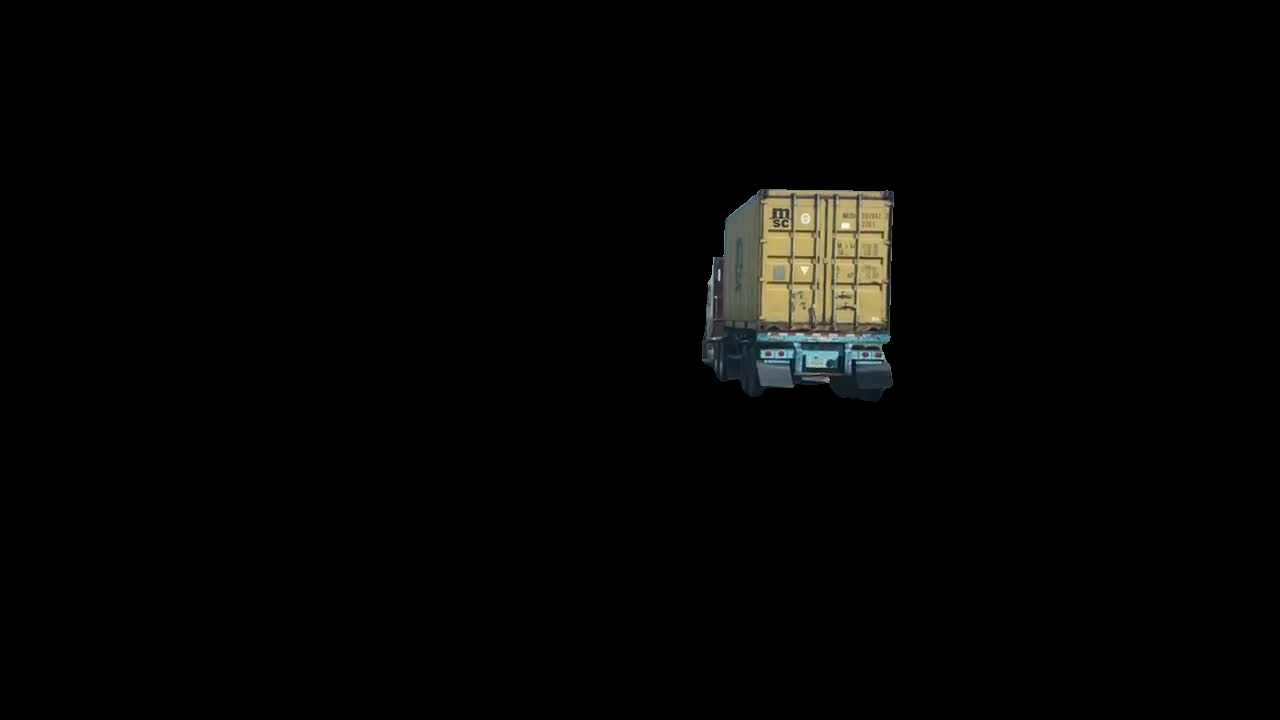}
    \end{minipage}
\label{orin}
}

\subfloat[Gaussian Blurring]
{
    \begin{minipage}[b]{.24\linewidth}
        \centering
        \includegraphics[scale=0.0935]{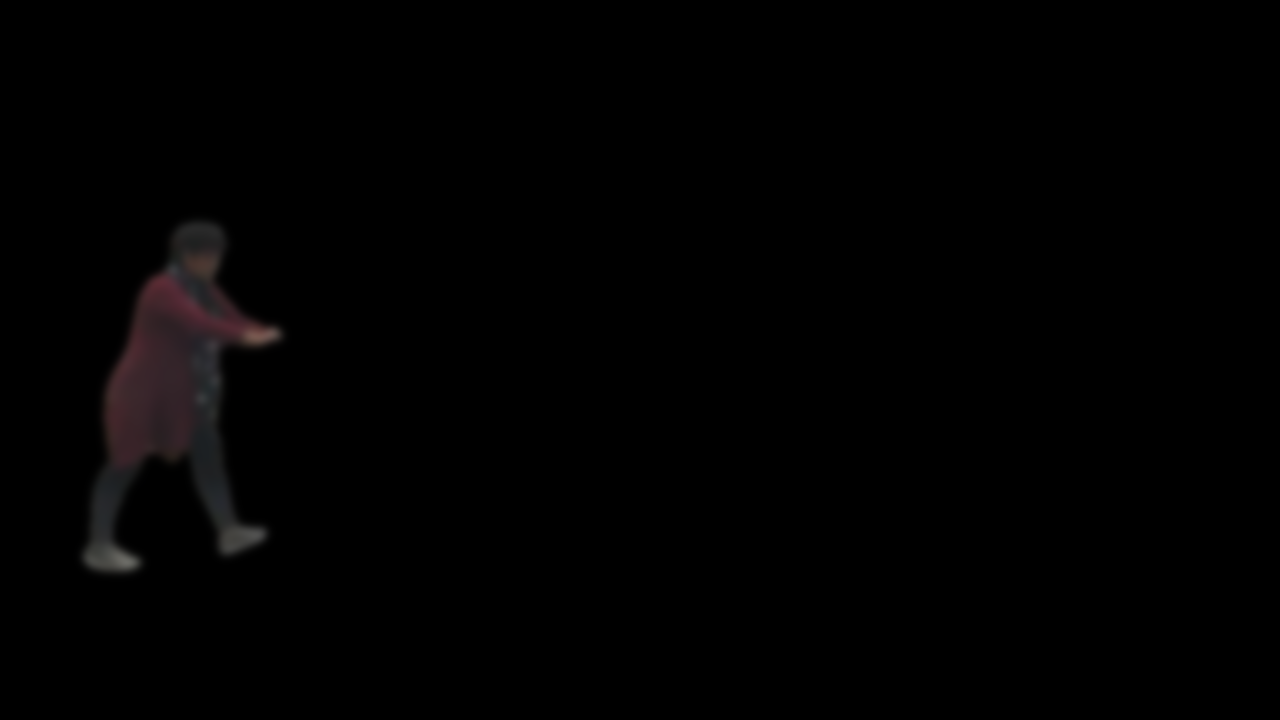}
    \end{minipage}
        \begin{minipage}[b]{.24\linewidth}
        \centering
        \includegraphics[scale=0.0935]{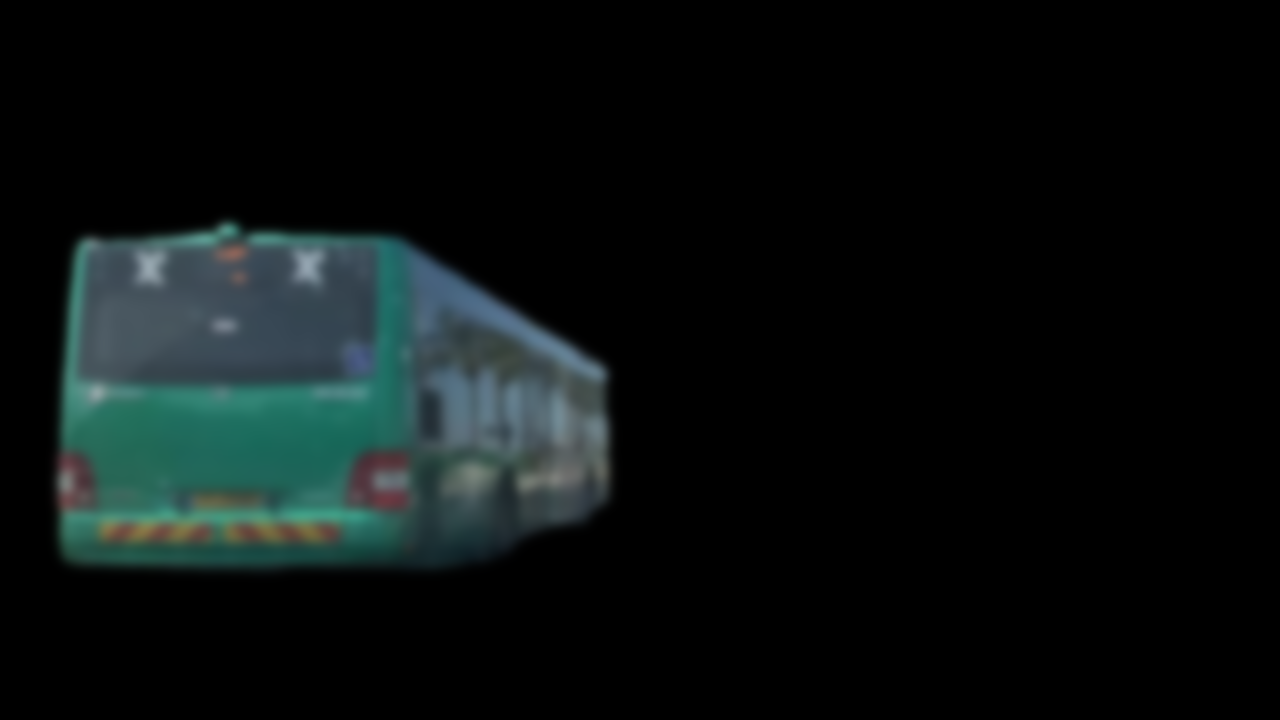}
    \end{minipage}
        \begin{minipage}[b]{.24\linewidth}
        \centering
        \includegraphics[scale=0.0935]{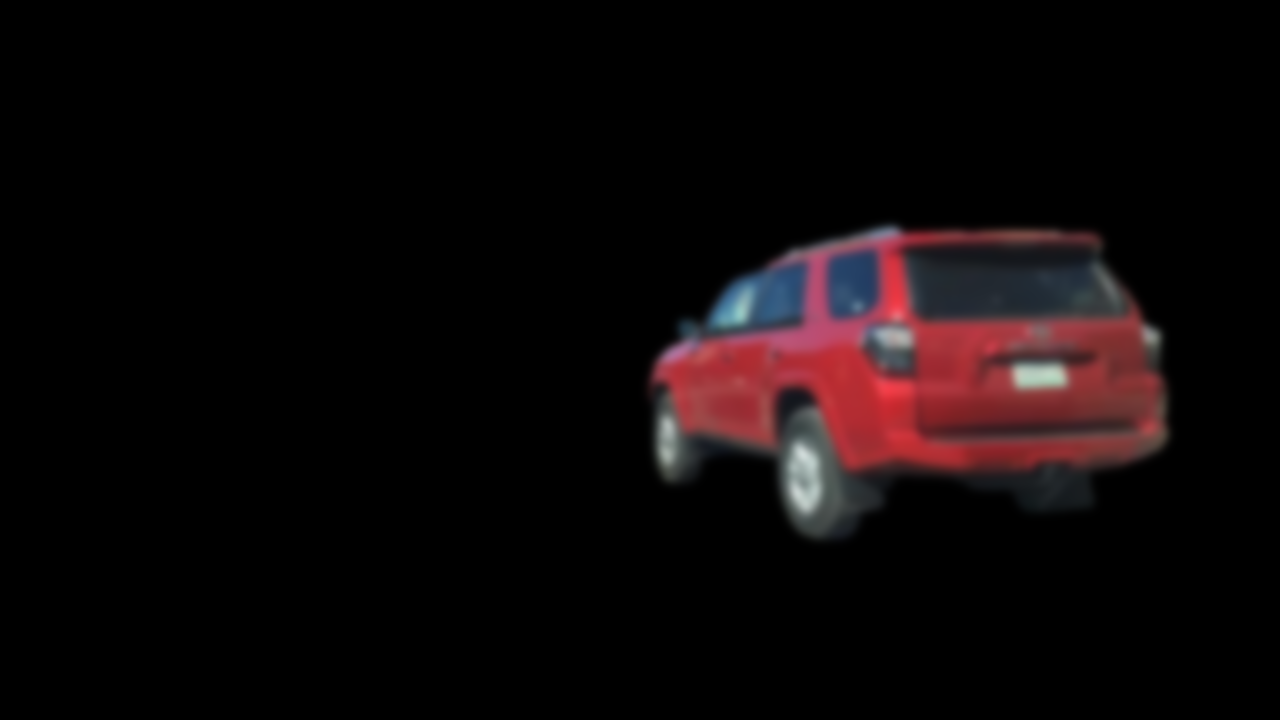}
    \end{minipage}
    \begin{minipage}[b]{.24\linewidth}
        \centering
        \includegraphics[scale=0.0935]{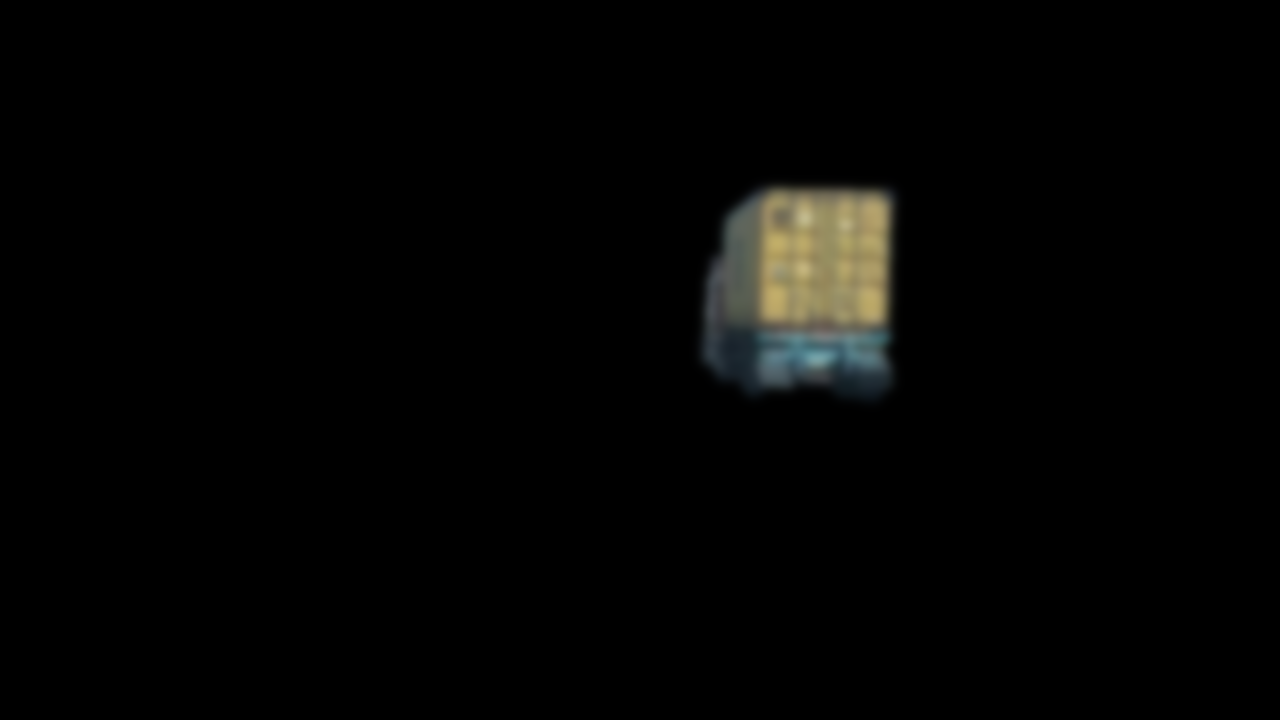}
    \end{minipage}
\label{gaussian blur}
}

\subfloat[Color Jittering]
{
    \begin{minipage}[b]{.24\linewidth}
        \centering
        \includegraphics[scale=0.0935]{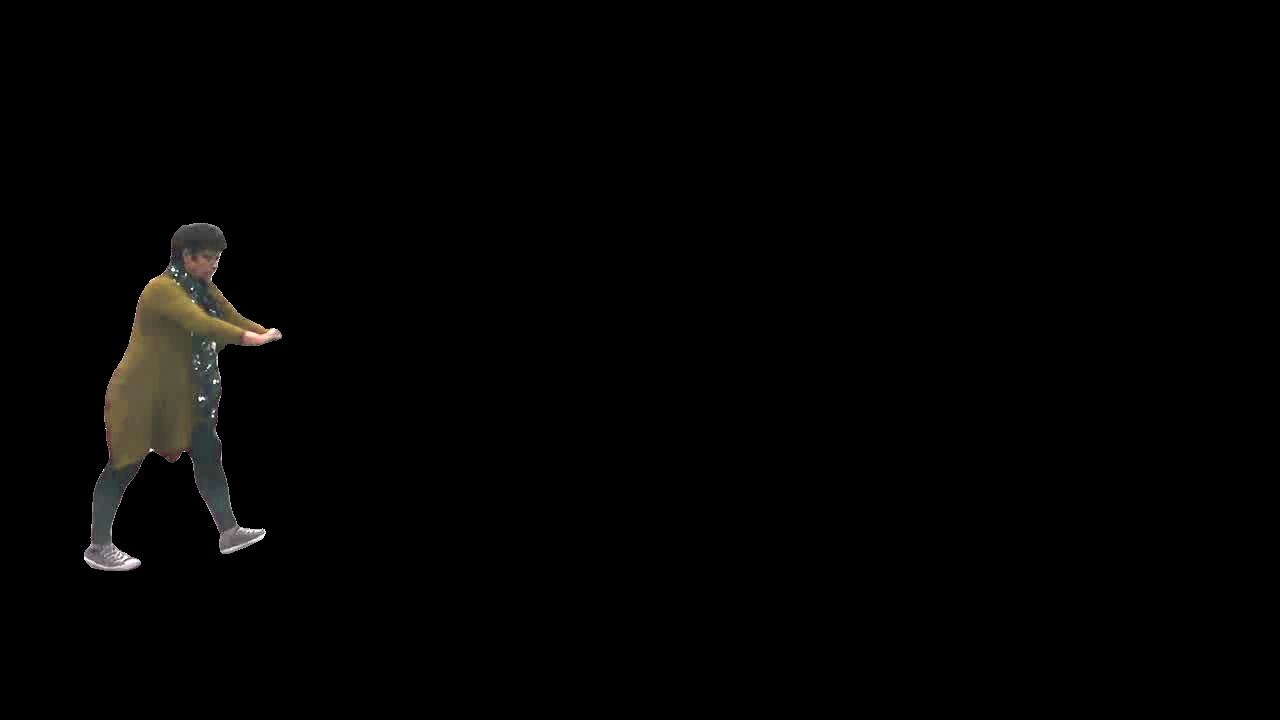}
    \end{minipage}
        \begin{minipage}[b]{.24\linewidth}
        \centering
        \includegraphics[scale=0.0935]{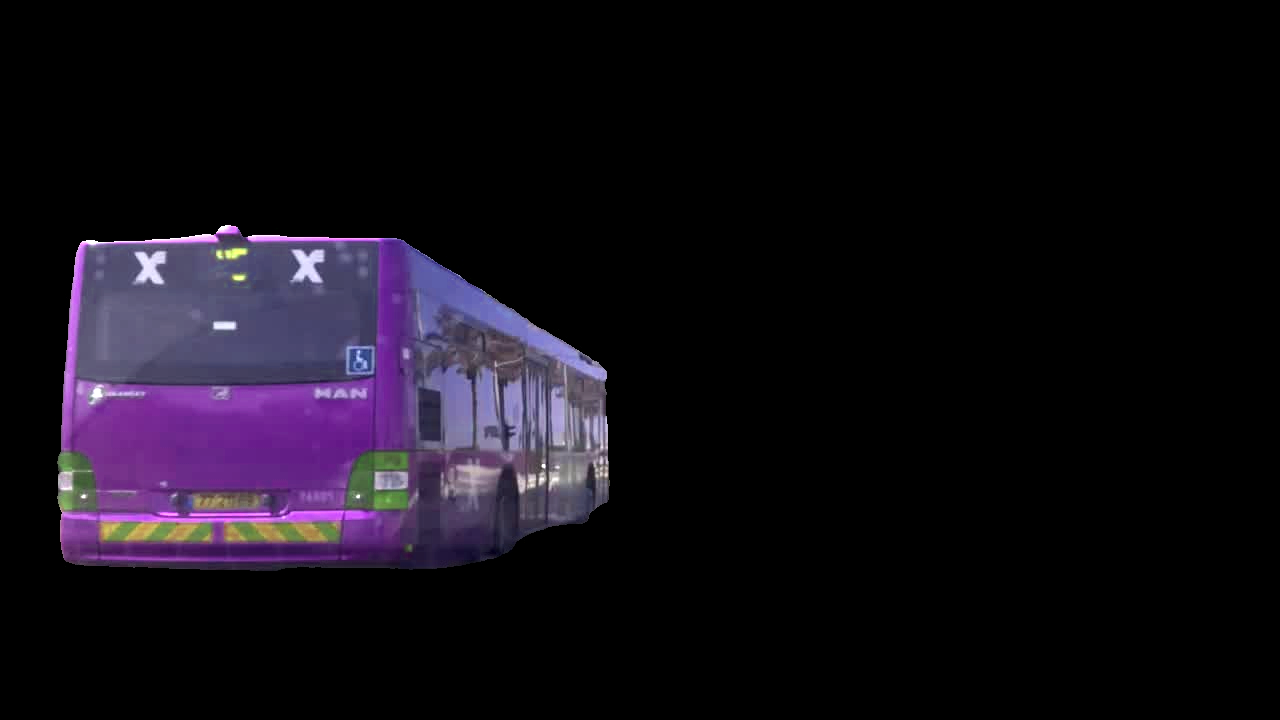}
    \end{minipage}
        \begin{minipage}[b]{.24\linewidth}
        \centering
        \includegraphics[scale=0.0935]{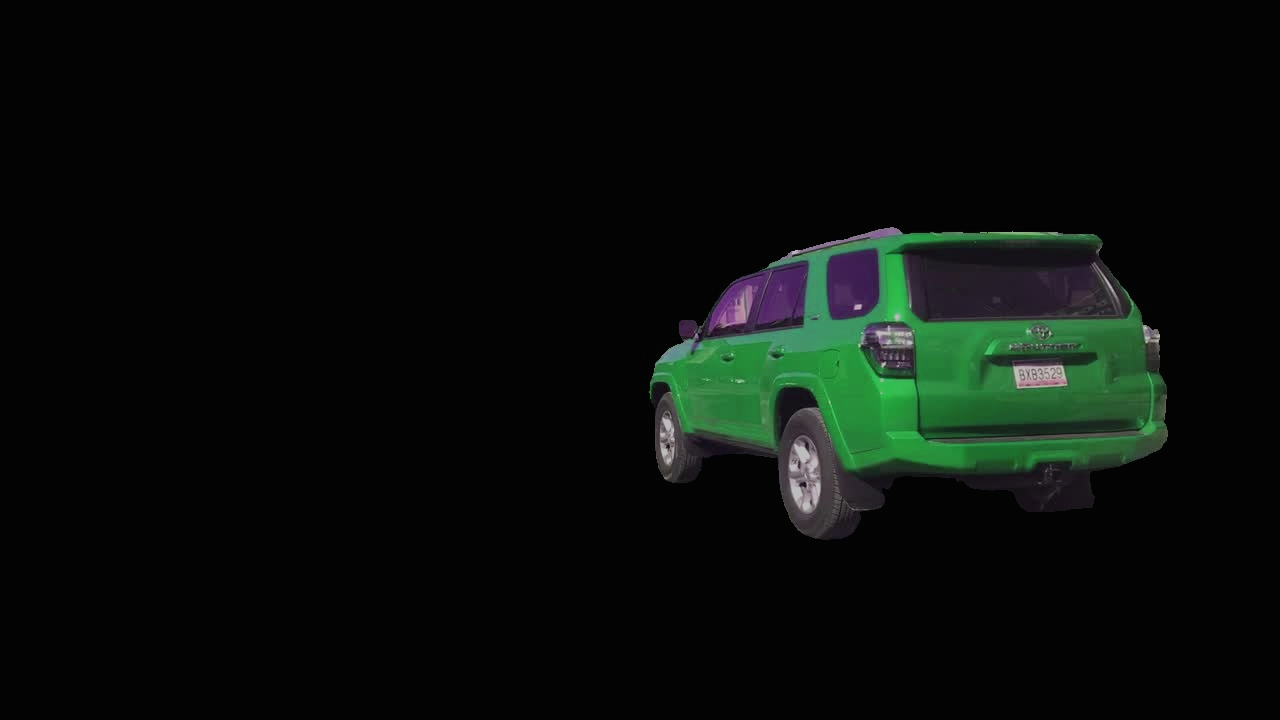}
    \end{minipage}
    \begin{minipage}[b]{.24\linewidth}
        \centering
        \includegraphics[scale=0.0935]{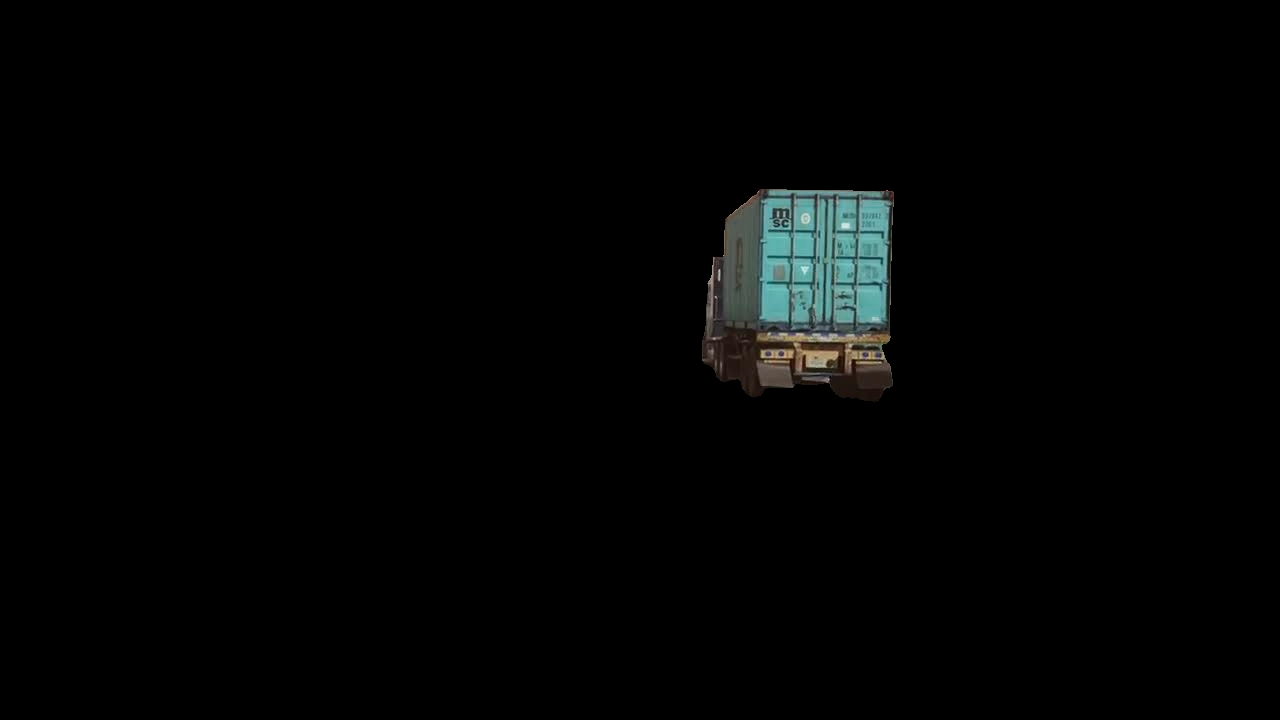}
    \end{minipage}
\label{color jitter}
}

\subfloat[Random Erasing]
{
    \begin{minipage}[b]{.24\linewidth}
        \centering
        \includegraphics[scale=0.0935]{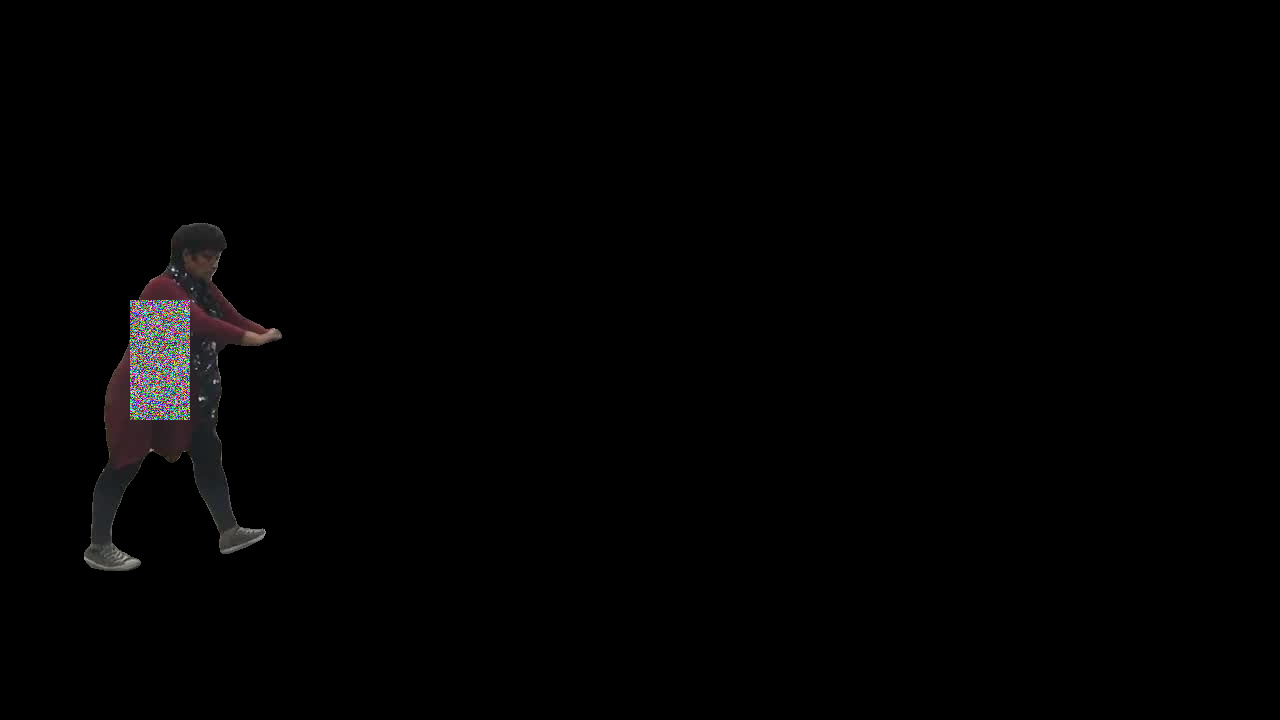}
    \end{minipage}
        \begin{minipage}[b]{.24\linewidth}
        \centering
        \includegraphics[scale=0.0935]{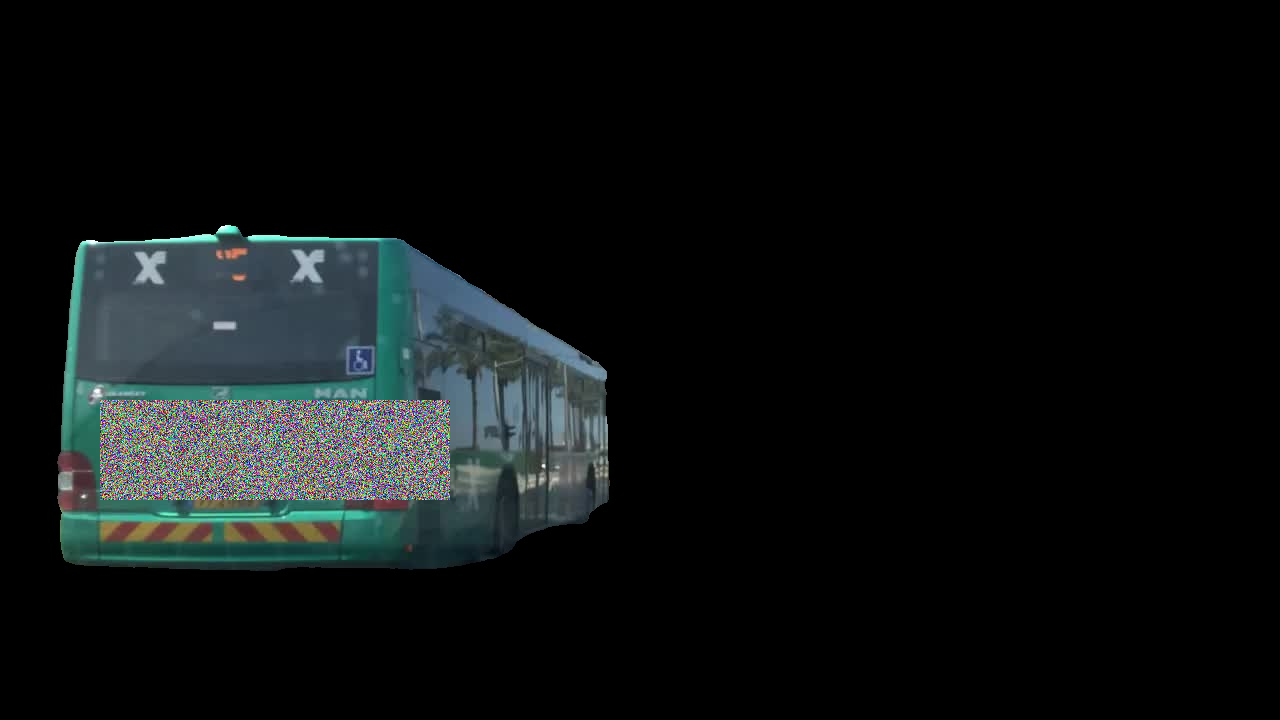}
    \end{minipage}
        \begin{minipage}[b]{.24\linewidth}
        \centering
        \includegraphics[scale=0.0935]{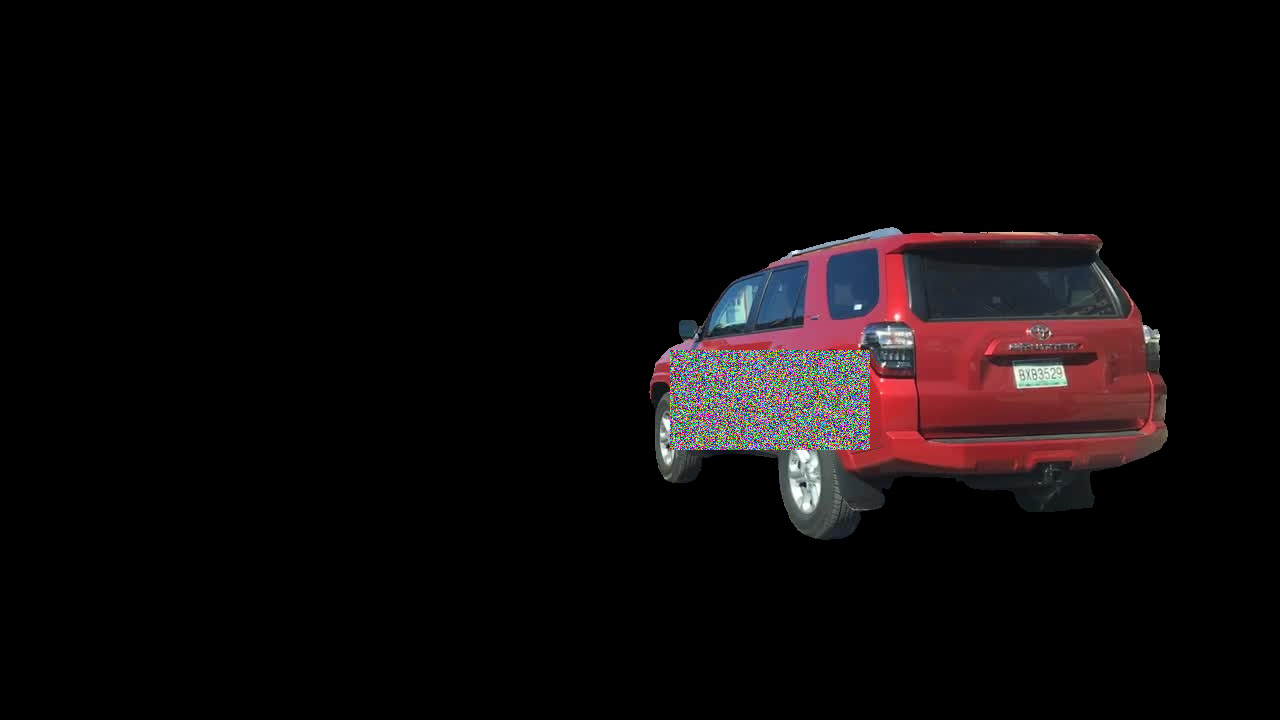}
    \end{minipage}
    \begin{minipage}[b]{.24\linewidth}
        \centering
        \includegraphics[scale=0.0935]{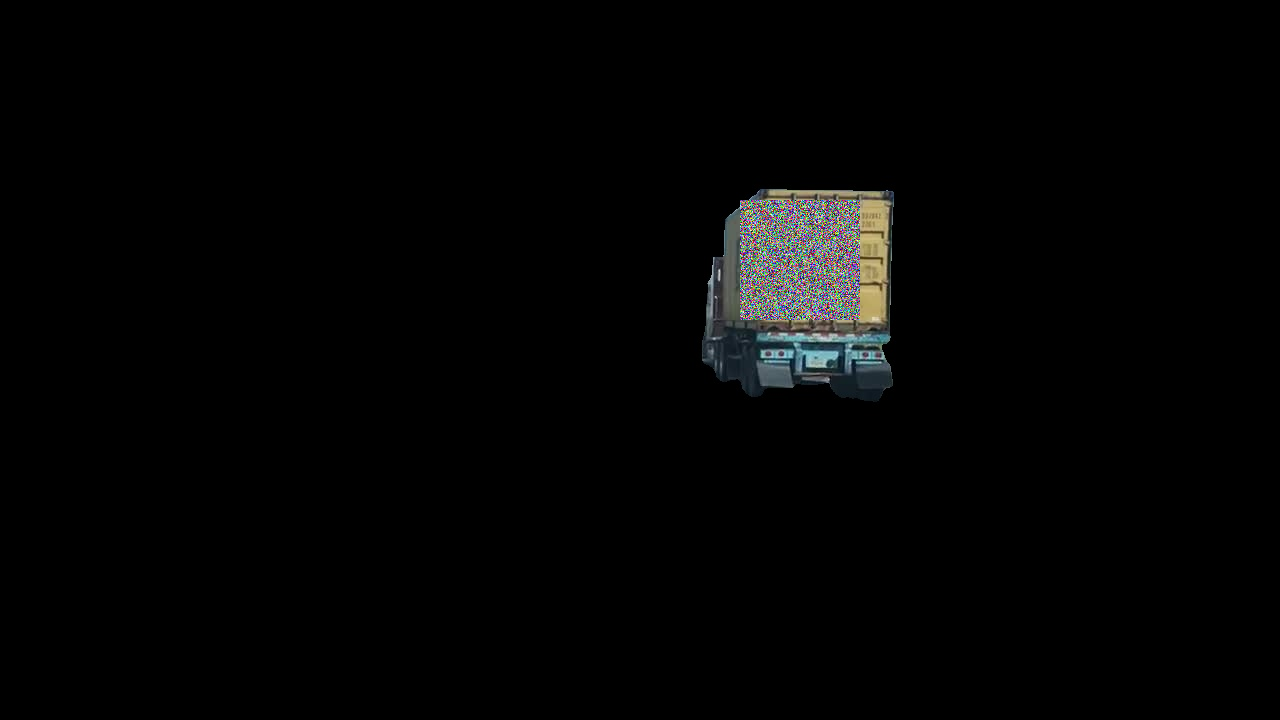}
    \end{minipage}
\label{erase}
}

\subfloat[Grayscale]
{
    \begin{minipage}[b]{.24\linewidth}
        \centering
        \includegraphics[scale=0.0935]{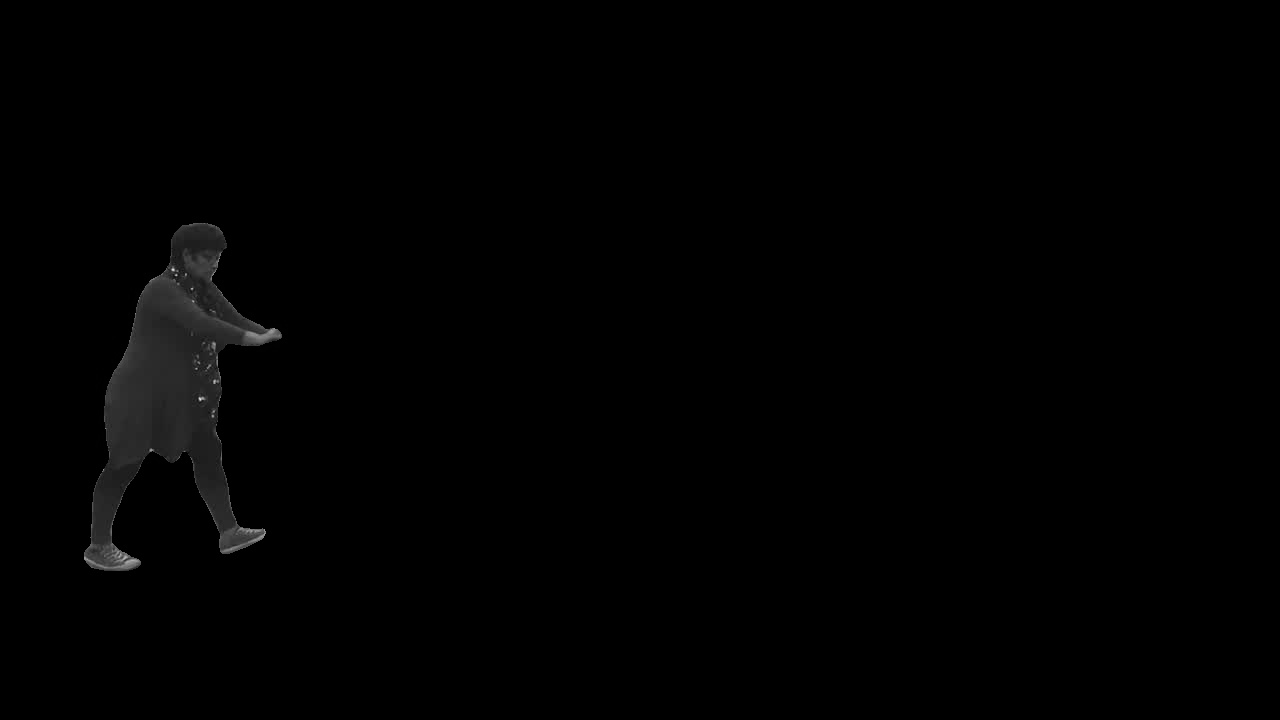}
    \end{minipage}
        \begin{minipage}[b]{.24\linewidth}
        \centering
        \includegraphics[scale=0.0935]{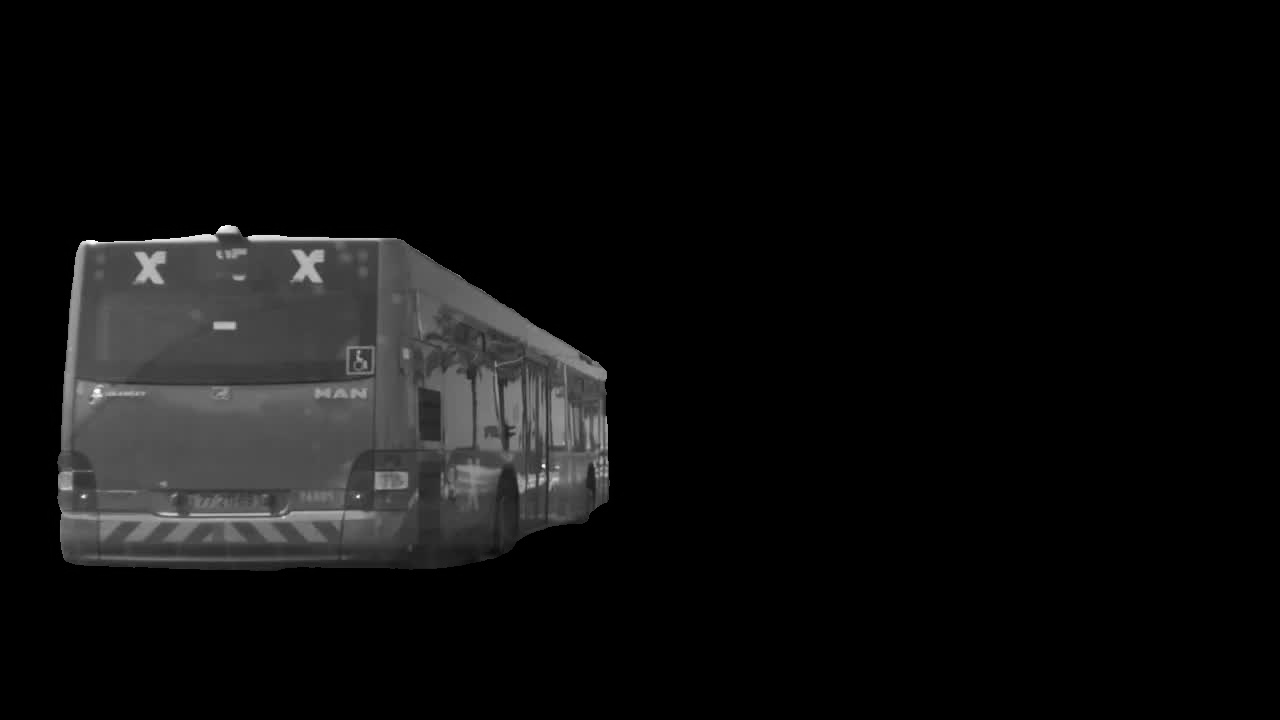}
    \end{minipage}
        \begin{minipage}[b]{.24\linewidth}
        \centering
        \includegraphics[scale=0.0935]{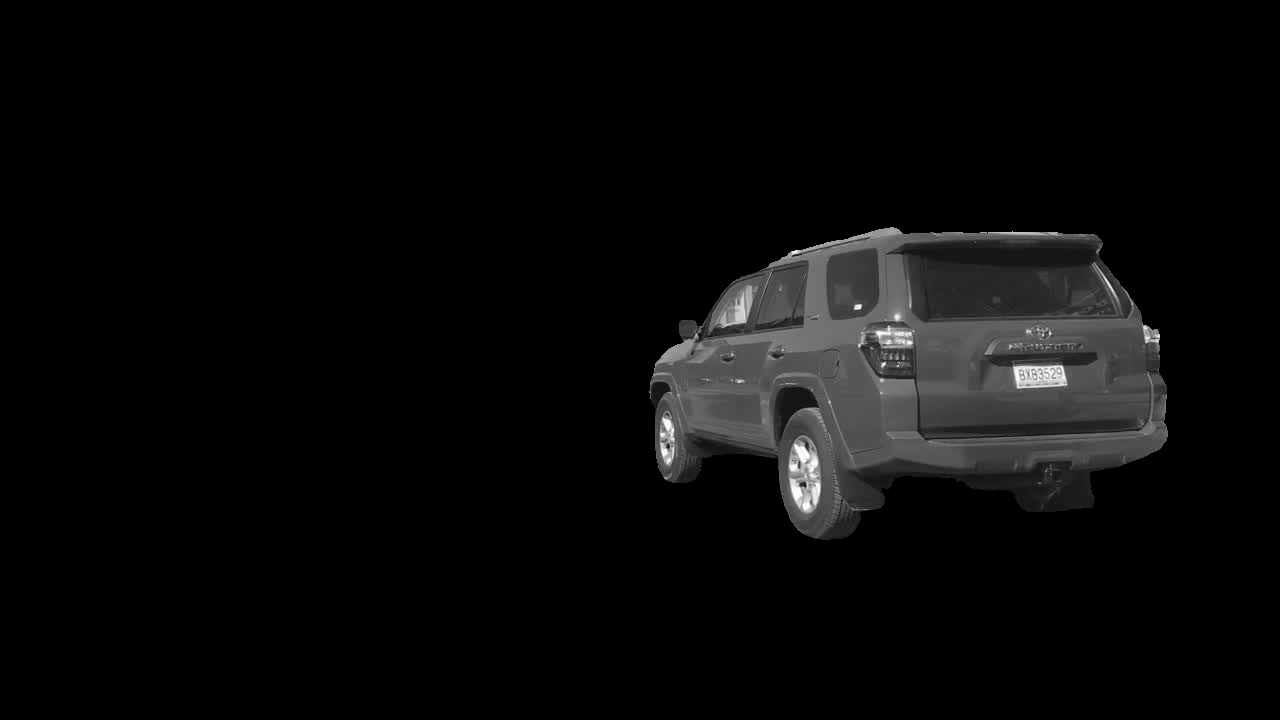}
    \end{minipage}
    \begin{minipage}[b]{.24\linewidth}
        \centering
        \includegraphics[scale=0.0935]{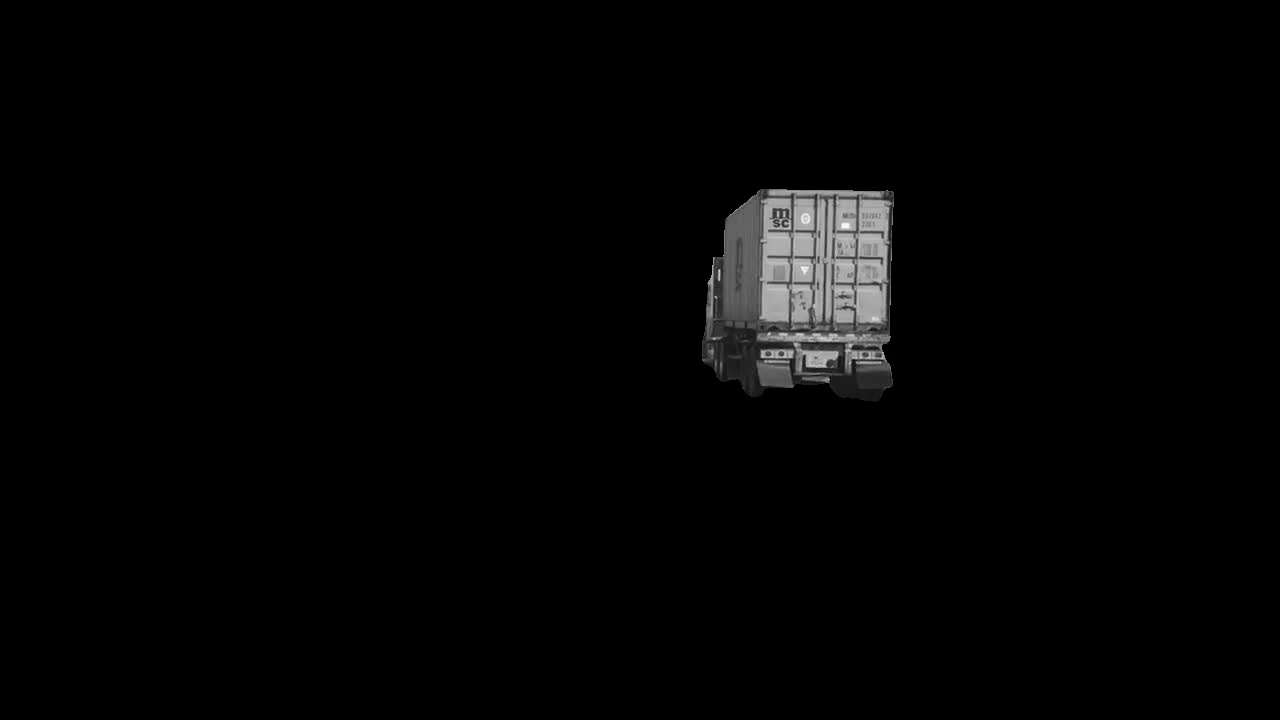}
    \end{minipage}
\label{gray}
}
\caption{Visualization examples of the local transformation strategies.}
\label{augfig}
\end{figure*}

We use Detectron2 \cite{wu2019detectron2} on a 24GB GeForce RTX 3090Ti to implement our method. During training, the temperature $\tau$ in $\mathcal{L}_{imp}$ is set to 0.2. Besides, the details of the local transformation strategies are as follows: 
\begin{itemize}
	\item \textbf{Gaussian Blurring}: We blur the object using a random selected square Gaussian kernel from the size of [23, 27, 29, 31, 33] with standard deviation of 0, as shown in Fig. \ref{gaussian blur}.
	\item \textbf{Color Jittering}: We randomly change the brightness, contrast, saturation and hue of an image by a random uniform offset, as demonstrated in Fig. \ref{color jitter}.
	\item  \textbf{Random Erasing}: We randomly select a rectangle region in an object and erase its pixels with random values, as shown in Fig. \ref{erase}.
  \item \textbf{Grayscale}: We randomly apply grayscale on the object, as shown in Fig. \ref{gray}.
\end{itemize}

\begin{figure*}[htbp]
\centering
\subfloat
{
    \begin{minipage}[b]{.24\linewidth}
        \centering
        \includegraphics[scale=0.13]{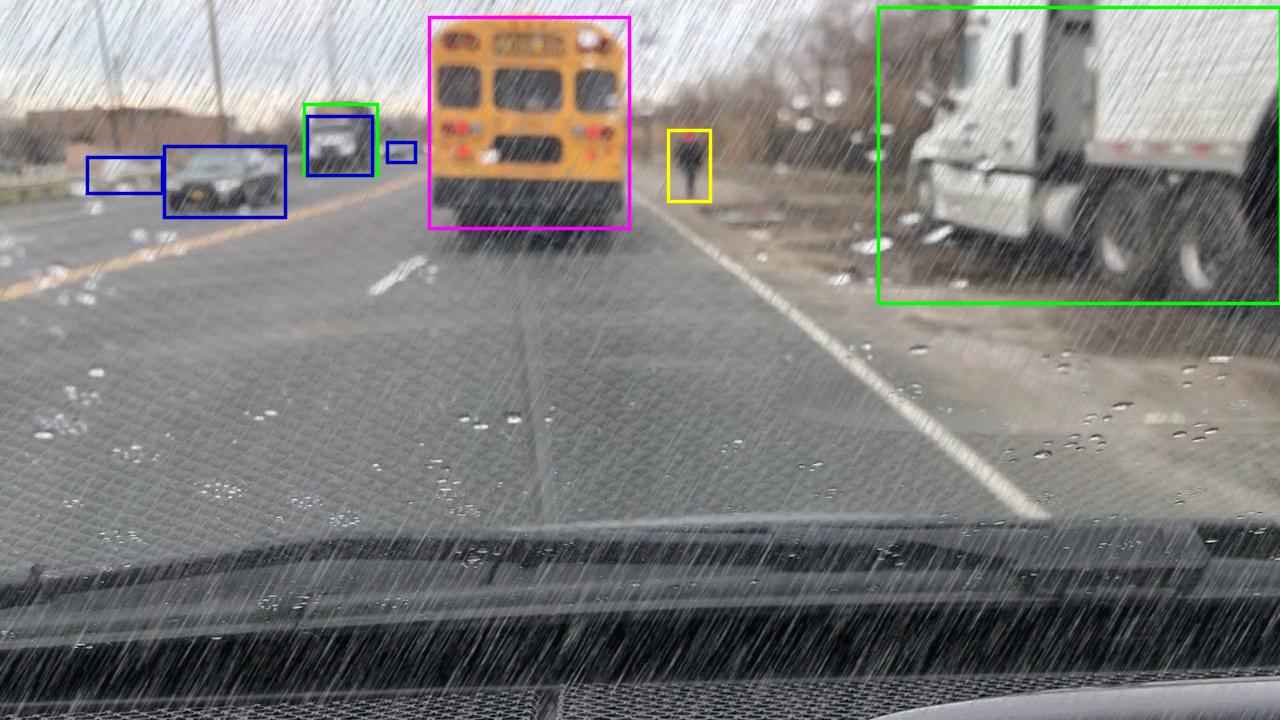}
    \end{minipage}
        \begin{minipage}[b]{.24\linewidth}
        \centering
        \includegraphics[scale=0.33555]{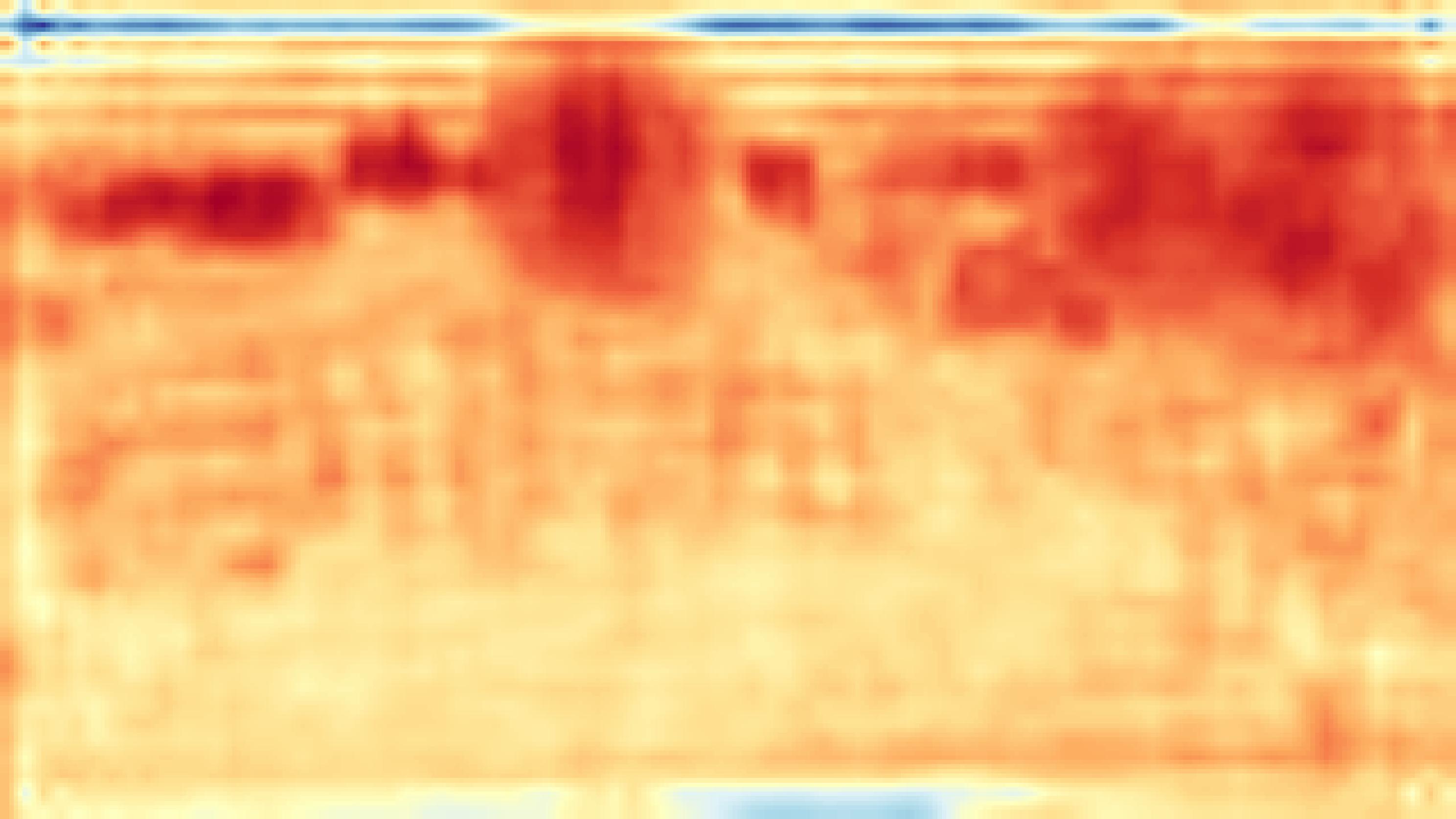}
    \end{minipage}
    \begin{minipage}[b]{.24\linewidth}
        \centering
        \includegraphics[scale=0.13]{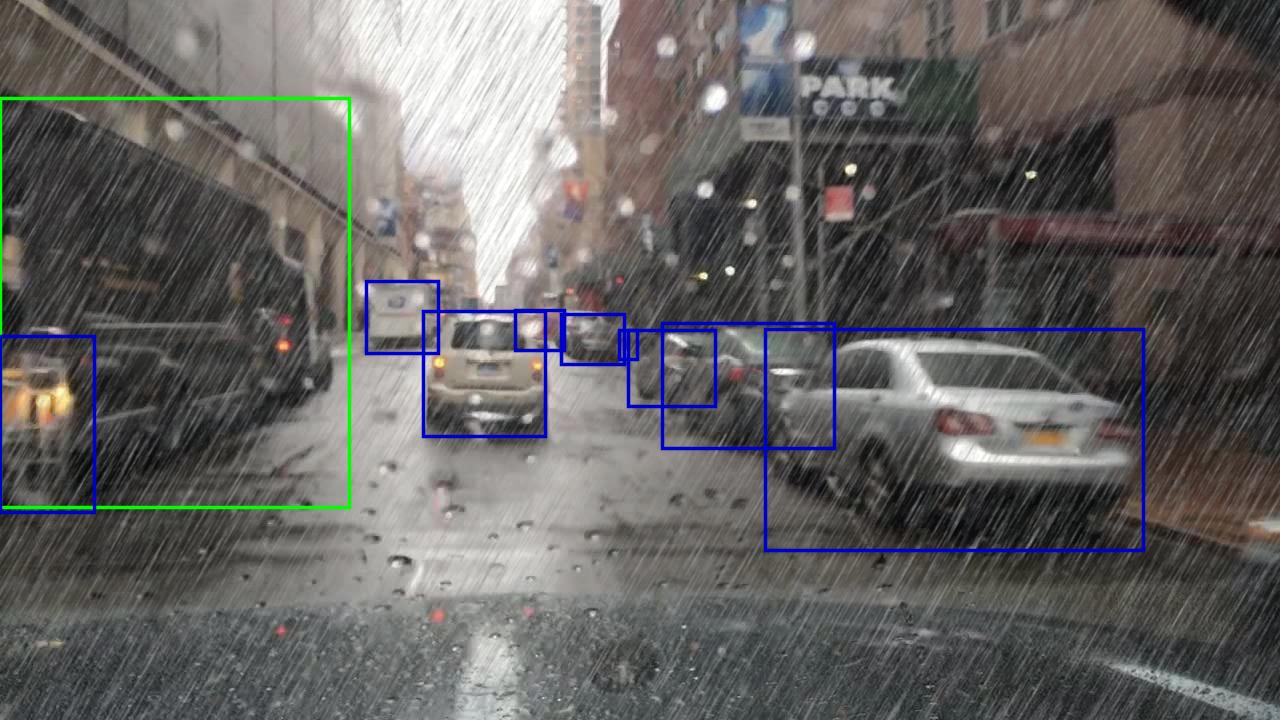}
    \end{minipage}
    \begin{minipage}[b]{.24\linewidth}
        \centering
        \includegraphics[scale=0.3355]{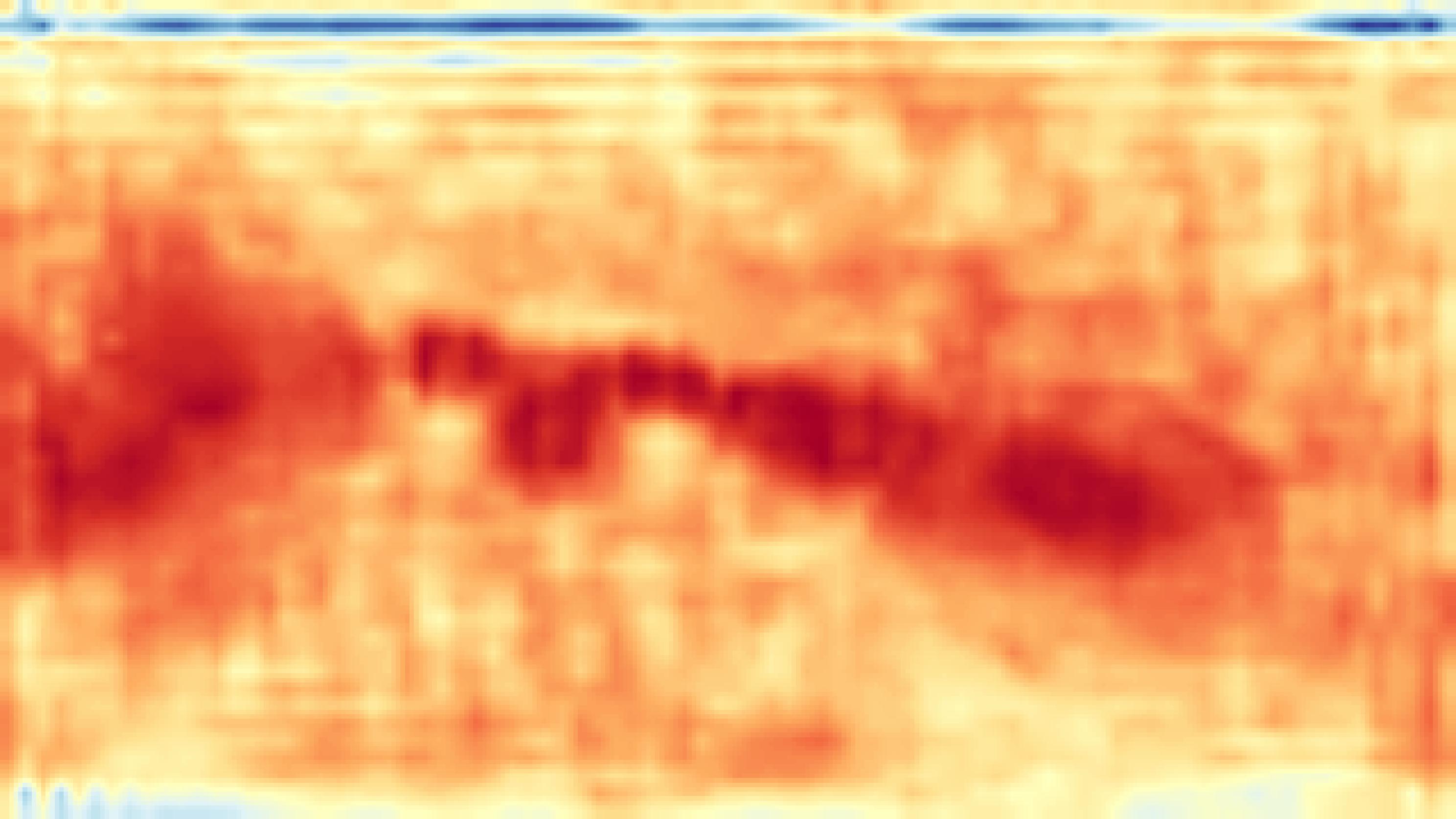}
    \end{minipage}
}
\caption{Visualization of the detection results and attention maps on the Dusk-Rainy scene without $L_{att}$.}
\label{attfig}
\vspace{-0.5em}
\end{figure*}
\subsection{Further analysis of the effectiveness in data debias, attention debias and prototype debias}
For data debias, to validate the effectiveness of the GLT module in bridging the domain gap with unseen target domains, we compare the \textbf{$\bm{L_1}$ distance} of features $F_1$, $F_2$ generated by \textbf{baseline} and \textbf{baseline+GLT} respectively with feature $F_3$ generated by model fine tuned in target domain in Table \ref{tab}. The results indicate that the $F_2$ of baseline+GLT is closer to $F_3$, indicating that the GLT module is effective in data debias. 

For prototype debias, we sample target images and feed into our \textbf{UFR model with $\mathcal{L}_{prot}$} constraint and \textbf{w/o $\mathcal{L}_{prot}$} constraint respectively to produce the prototypes $p_i^c$ of \textbf{category $\bm{c}$} in five domains, and then compute the average prototype across five domains for each category $c$ :
\begin{equation}
    p_{avg}^c = \frac{1}{5}\sum_{i=1}^{5}p_i^c,
\end{equation}
where $i$ is the domain number and $i=1...5$. Then we compute the averaged $\bm{L_1}$ \textbf{distance} $d^c$ for each category between prototype of each domain $p_i^c$ and the average prototype $p_{avg}^c$ to compare the \textbf{concentration degree of prototypes} w/ and w/o $\mathcal{L}_{prot}$ following:
\begin{equation}
d^c=\frac{1}{5}\sum_{i=1}^{5}\lvert p_i^c-p_{avg}^c\rvert.
\end{equation}
The results are shown  in Table \ref{tab2}. The results demonstrate that the model can generate more concentrated prototypes of the same category in different domains with our designed $\mathcal{L}_{prot}$, indicating the prototype debias ability of our model.

The effectiveness of attention debias is reflected from the experiment in paper (Fig. 8). We further demonstrate the attention maps generated by our UFR without $L_{att}$ in Fig.  \ref{attfig} for comparison.\\

\begin{table}
\centering
\setlength{\tabcolsep}{1.5mm}{
    \begin{tabular}{ccccc}
    \hline
         $\tau$ &0.07  &0.2 &0.3 &1.0\\\hline
         mAP (\%) &40.2  &40.8  &40.5  &39.9  \\ \hline
    \end{tabular}}
        \captionof{table}{Results of $\tau$ analysis}
        \label{tab3}
\end{table}  

\subsection{Hyperparameter analysis}
We analyze the impact of temperature $\tau$ in $\mathcal{L}_{imp}$. The contrastive loss is widely used and we follow the common setting to select $\tau$ from [0.07, 0.2, 0.3, 1.0]. We analyze the impact of $\tau$ on Night-Clear scene and the results are demonstrated in Table \ref{tab3}. The results show that when the value of $\tau$ is small, it has a relatively slight effect on the results, and when $\tau$ is increased to 1.0, the results show a significant decrease. We take the value of $\tau$ as 0.2 based on this analysis.

\section{Discussion on the adoption of SAM \cite{kirillov2023segment} in single-domain generalization tasks}
In this task, we leverage the powerful segmentation capabilities of SAM \cite{kirillov2023segment} to produce object masks of training data. However, the use of SAM \cite{kirillov2023segment} may give rise to controversy about whether it violates the single-domain generalization setting. We think that it doesn't violate the single domain setting, for that we don't reach other data or leverage the SAM \cite{kirillov2023segment} for testing results. It is just a tool to obtain object masks. The process can be realized with the help of arbitrary segmentation models or even extracted manually. However, we use a more accurate large model to realize it.  In addition, with the development of foundation models, there has been a trend of how to leverage them for efficiency gains in various tasks. These models are also used in other cross-domain works, such as the CLIP-Gap \cite{vidit2023clip}  in our comparison experiments, which used a CLIP \cite{radford2021learning} model for domain augmentation.

{
    \small
    \bibliographystyle{ieeenat_fullname}
    \bibliography{supplement}
}